%% file: main.tex
\documentclass{article}

\usepackage[english]{babel}

\usepackage[letterpaper,top=2cm,bottom=2cm,left=3cm,right=3cm,marginparwidth=1.75cm]{geometry}

\usepackage{amsmath}
\usepackage{amssymb}
\usepackage{graphicx}
\usepackage{subcaption}
\usepackage{float}
\usepackage{tcolorbox}
\usepackage[colorlinks=true, allcolors=blue]{hyperref}

\usepackage{tikz}
\usetikzlibrary{shapes,arrows,positioning,calc,shadows,decorations.pathreplacing}
\usetikzlibrary{shapes.geometric}
\usetikzlibrary{backgrounds}
\usetikzlibrary{fit} 
\usetikzlibrary{patterns}
\usepackage{adjustbox}

\usepackage{pgfplots}
\pgfplotsset{compat=1.18}
\usetikzlibrary{patterns}
\usepgfplotslibrary{fillbetween}
\title{Exploring Efficiency Frontiers of Thinking Budget in Medical Reasoning: Scaling Laws between Computational Resources and Reasoning Quality}
\makeatletter
\renewcommand\maketitle{
  \begin{center}
    {\LARGE\bfseries\@title\par}
    \vskip 1.5em
    {\normalsize
      \begin{tabular}[c]{c}
        Ziqian Bi\textsuperscript{1*},
        Lu Chen\textsuperscript{2*},
        Junhao Song\textsuperscript{1*},
        Hongying Luo\textsuperscript{2},
        Enze Ge\textsuperscript{1},
        Junmin Huang\textsuperscript{2},\\
        Tianyang Wang\textsuperscript{1},
        Keyu Chen\textsuperscript{1},
        Chia Xin Liang\textsuperscript{1},
        Zihan Wei\textsuperscript{2},
        Huafeng Liu\textsuperscript{2},
        Chunjie Tian\textsuperscript{3},\\
        Jibin Guan\textsuperscript{4},
        Joe Yeong\textsuperscript{5},
        Yongzhi Xu\textsuperscript{2},
        Peng Wang\textsuperscript{2†},
        Xinyuan Song\textsuperscript{1}
        Junfeng Hao\textsuperscript{2†}
      \end{tabular}\par}
    \vskip 0.8em
    {\footnotesize
      \textsuperscript{1}AI Agent Lab, Vokram Group, London, United Kingdom \\
      \textsuperscript{2}Department of Nephrology, Affiliated Hospital of Guangdong Medical University, Zhanjiang, China \\
      \textsuperscript{3}Department of Otorhinolaryngology, Affiliated Hospital of Guangdong Medical University, Zhanjiang, China \\
      \textsuperscript{4}Masonic Cancer Center, University of Minnesota, Minneapolis, United States \\
      \textsuperscript{5}Division of Pathology, Singapore General Hospital, Singapore \\
      \textsuperscript{*}Equal contribution, 
      \textsuperscript{†}Corresponding authors: wangpeng@gdmu.edu.cn, ygzhjf85@gmail.com
    }
  \end{center}
  \vskip 1.5em
}
\makeatother

\begin{document}
\maketitle

\begin{abstract}
This study presents the first comprehensive evaluation of thinking budget mechanisms in medical reasoning tasks, revealing fundamental scaling laws between computational resources and reasoning quality. We systematically evaluated two major model families, Qwen3 (1.7B to 235B parameters) and DeepSeek-R1 (1.5B to 70B parameters), across 15 medical datasets spanning diverse specialties and difficulty levels. Through controlled experiments with thinking budgets ranging from zero to unlimited tokens, we establish logarithmic scaling relationships where accuracy improvements follow a predictable pattern with both thinking budget and model size. Our findings identify three distinct efficiency regimes: high-efficiency (0 to 256 tokens) suitable for real-time applications, balanced (256 to 512 tokens) offering optimal cost-performance tradeoffs for routine clinical support, and high-accuracy (above 512 tokens) justified only for critical diagnostic tasks. Notably, smaller models demonstrate disproportionately larger benefits from extended thinking, with 15 to 20 \% improvements compared to 5 to 10 \% for larger models, suggesting a complementary relationship where thinking budget provides greater relative benefits for capacity-constrained models. Domain-specific patterns emerge clearly, with neurology and gastroenterology requiring significantly deeper reasoning processes than cardiovascular or respiratory medicine. The consistency between Qwen3 native thinking budget API and our proposed truncation method for DeepSeek-R1 validates the generalizability of thinking budget concepts across architectures. These results establish thinking budget control as a critical mechanism for optimizing medical AI systems, enabling dynamic resource allocation aligned with clinical needs while maintaining the transparency essential for healthcare deployment.
\end{abstract}


\input{content/introduction}

\input{content/related}

\input{content/experiment}

\input{content/result}

\input{content/limitation}

\input{content/conclusion}

\newpage
\appendix
\renewcommand{\thesection}{Appendix \Alph{section}}
\input{content/appendix}

\bibliographystyle{ieeetr}
\bibliography{reference}

\end{document}

%% file: content/introduction.tex
\section{Introduction}

Large language models (LLMs) have achieved remarkable progress in complex reasoning tasks, from the GPT series \cite{brown2020gpt3,openai2023gpt4} to Claude models \cite{alain2025evaluating} and the Qwen series \cite{bai2023qwen}. However, traditional models employ fixed computational strategies regardless of problem complexity, limiting their effectiveness in specialized domains.

Qwen3 introduces revolutionary Hybrid Thinking Modes, enabling seamless switching between Thinking Mode and Non-Thinking Mode \cite{yang2025qwen3}. In thinking mode, the model performs step-by-step reasoning with explicit \texttt{$<$think$>$...$<$/think$>$} blocks, making the reasoning process transparent and observable. This transforms models from ``black box'' systems to interpretable reasoning engines, establishing the foundation for the Thinking Budget mechanism.

The Thinking Budget mechanism allows precise control of computational resources during reasoning, achieving a dynamic balance between depth and efficiency. Formally, we define the thinking budget as $T_b = \min(T_{requested}, T_{max})$, where $T_{requested}$ is the user-specified token limit and $T_{max}$ is the model's maximum capacity \cite{li2025steering}. Simple problems receive concise thinking processes with smaller budgets, while complex problems benefit from deep multi-step reasoning with larger budgets. This adaptive allocation transforms the traditional ``one-size-fits-all'' strategy, optimizing the efficiency function $E(T_b) = \frac{\Delta \text{Accuracy}(T_b)}{T_b}$.

\begin{figure*}[htbp]
    \centering
    \includegraphics[width=\textwidth]{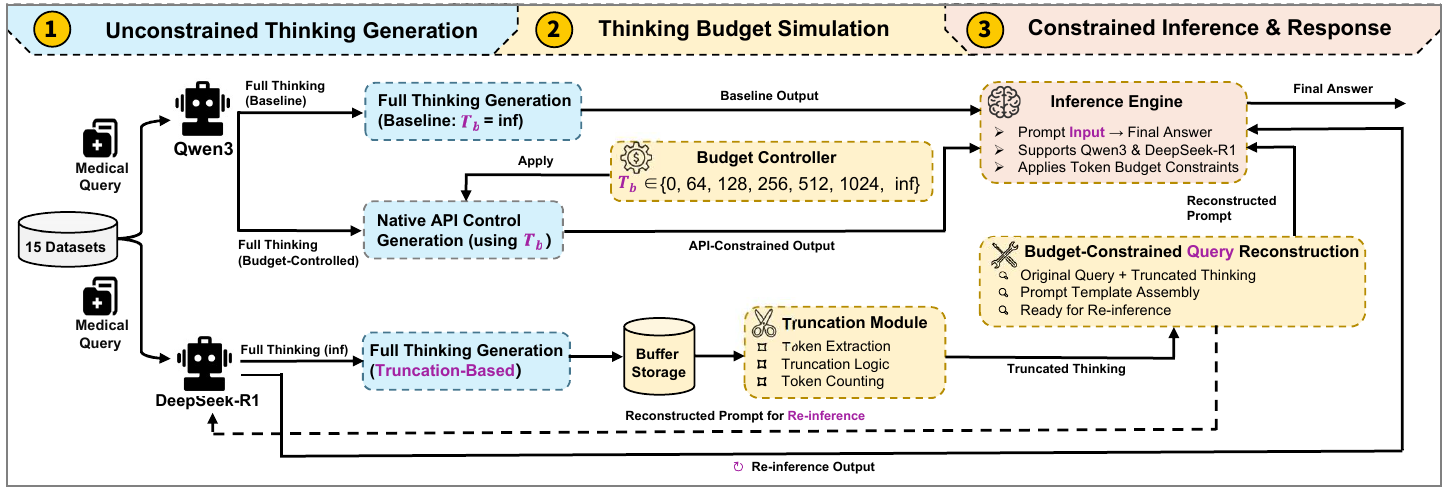}
\caption{
Overview of our three-stage thinking budget evaluation pipeline for medical reasoning tasks.
\textbf{Stage 1 (Unconstrained Thinking Generation):} Both model families generate complete reasoning traces without budget constraints. Qwen3 produces full thinking processes via native API calls, while DeepSeek-R1 generates unrestricted reasoning sequences stored in a buffer for subsequent processing.
\textbf{Stage 2 (Thinking Budget Simulation):} Two distinct control mechanisms are employed. For Qwen3, budget constraints are enforced directly through the native \texttt{thinking\_budget} API parameter during generation. For DeepSeek-R1, which lacks native budget control, we implement a truncation-based approach: (i) extract thinking content between \texttt{<think>} and \texttt{</think>} tags from Stage 1 outputs, (ii) apply token-level truncation at predefined budgets $T_b \in \{0, 64, 128, 256, 512, 1024\}$, and (iii) preserve the \texttt{inf} condition as the original full reasoning trace.
\textbf{Stage 3 (Constrained Inference \& Response):} Budget-constrained prompts are reconstructed by combining the original medical query with truncated thinking content. These reconstructed prompts are then fed to the inference engine for final answer generation.}
    \label{fig:pipeline}
\end{figure*}

\begin{figure}[ht]
  \centering
  \begin{adjustbox}{max height=1\textheight}
    \begin{minipage}{1.0\textwidth}
      \centering
      \includegraphics[width=1\linewidth]{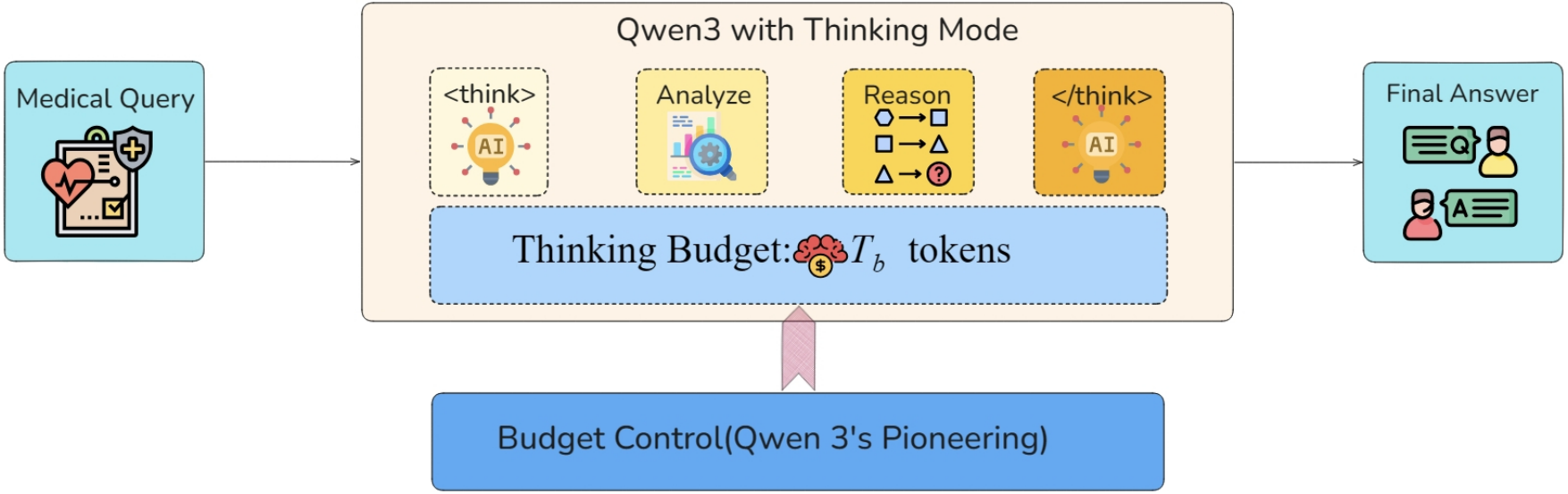}
      \vspace{0.1em}

      \includegraphics[width=1\linewidth]{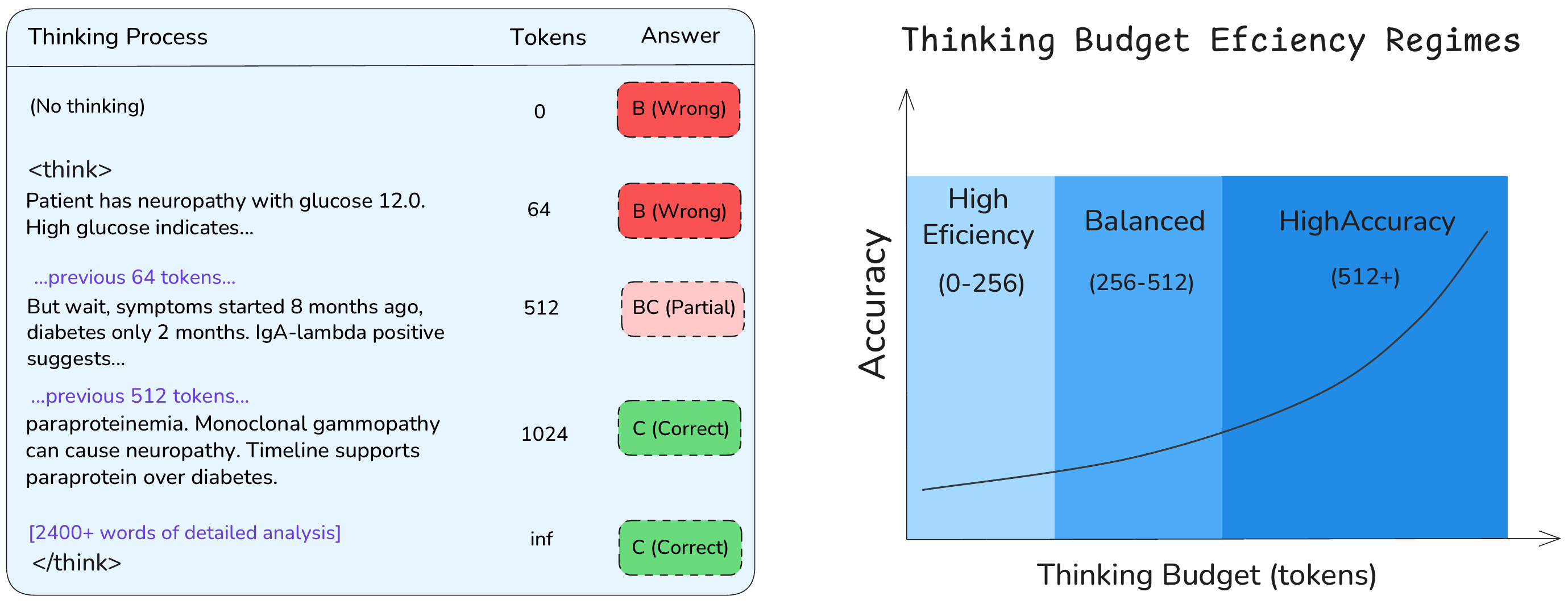}
    \end{minipage}
  \end{adjustbox}
  \caption{Illustration of the Thinking Model and Budget Mechanism. The top panel shows how medical queries are processed through Qwen3's thinking mode with controllable budget allocation. The bottom panel demonstrates how progressive thinking leads to better answers (left table) and visualizes the three efficiency regimes identified in our study (right chart).}
  \label{fig:thinking-budget}
\end{figure}

The medical domain provides an ideal testing environment due to its vast knowledge requirements, multi-level logical processes, and varying task complexity. Medical diagnosis involves a systematic approach that includes a comprehensive assessment of clinical symptoms, integration of patient medical history, and incorporation of relevant diagnostic examinations \cite{salvi2024multi}. This multi-step evaluation process aims to enhance the accuracy and clinical relevance of the diagnostic outcomes. For instance, clinical practice guidelines for disease management or standardized protocols for differential diagnosis serve as essential frameworks in medical decision-making. The explicit reasoning traces enable clinicians to verify diagnostic logic, building essential trust in AI-assisted diagnosis \cite{salybekov2024artificial}. Rather than rendering the final diagnosis difficult to justify based on the available evidence.
Dynamic thinking budgets resolve the contradiction between computational efficiency for simple queries and reasoning depth for complex diagnoses.

Existing research highlights the potential of large language models (LLMs) as diagnostic aids for healthcare professionals, though key challenges remain. First, LLMs may struggle to differentiate between diseases with overlapping symptom profiles or rare conditions \cite{schumacher2025rare}. Second, the diagnostic accuracy of these models is difficult to translate into high-stakes medical domains where precision is paramount \cite{su2025large}. Third, their capacity to interpret natural language remains limited, particularly in recognizing non-standardized or colloquial medical terminology \cite{soroush2024large}. Furthermore, disease progression is inherently dynamic, often presenting with variable clinical manifestations across different stages. Despite these limitations, recent innovative advancements in LLMs are reshaping the boundaries of what is clinically feasible.

While scaling laws \cite{kaplan2020scaling} have guided training-time optimization, inference-time resource allocation remains underexplored. We apply efficiency frontier theory \cite{markowitz1952portfolio,markowitz1959portfolio} to identify optimal thinking budget configurations that maximize reasoning quality under computational constraints. This work systematically explores thinking budget efficiency frontiers in medical reasoning across 15 datasets spanning multiple specialties and difficulty levels.

This work addresses several key research questions. We investigate whether thinking mode consistently improves medical reasoning performance and explore whether diminishing returns exist with increased thinking budgets. We seek to identify optimal budget configurations for different task complexities and examine whether thinking depth and reasoning quality follow predictable scaling laws. Additionally, we analyze how structural characteristics of reasoning processes influence final accuracy.

Our research makes several significant contributions. We present the first systematic evaluation of thinking budget mechanisms in medical reasoning and develop an efficiency frontier framework for optimal resource allocation. We establish quantitative mappings between task complexity and optimal budgets, providing empirical validation that demonstrates 5-20\% performance improvements with appropriate thinking allocation. Furthermore, our structural analysis reveals important connections between thinking quality and reasoning accuracy, offering insights for future medical AI system design.

The paper is organized as follows: Section~\ref{sec:related} reviews related work; Section~\ref{sec:experiment} describes experimental design; Section~\ref{sec:results} presents results and analysis; Section~\ref{sec:limitations} discusses limitations; and the final section concludes with future directions.

%% file: content/related.tex
\section{Related Work}
\label{sec:related}

Our work builds upon several interconnected research areas spanning thinking mechanisms in language models, medical AI systems, scaling laws, and adaptive computation. The concept of explicit thinking processes in language models has evolved from early chain-of-thought prompting \cite{wei2022chain} to native thinking capabilities in recent model architectures. While Wei et al. demonstrated that encouraging models to show intermediate reasoning steps improves complex task performance, subsequent work has explored various prompting strategies including zero-shot reasoning \cite{kojima2022large}, self-consistency \cite{zhang2025igniting}, complexity-based prompting \cite{fu2023complexitybased}, tree-of-thoughts \cite{yao2023tree}, and self-refinement \cite{madaan2023selfrefine}. The performance gain from the chain-of-thought can be formalized as:
$$\Delta P_{\text{CoT}} = P(\text{answer}|\text{prompt}, \text{reasoning}) - P(\text{answer}|\text{prompt}),$$
where the inclusion of explicit reasoning steps significantly increases accuracy on complex tasks. Recent advances in inference-time computation \cite{gschwind2025ai} and process supervision \cite{lightman2023lets} have further validated the importance of thinking processes, with optimal test-time compute scaling following:
$$\text{Performance} \propto \log(1 + \alpha \cdot T_{\text{inference}}),$$
where $\alpha$ represents task-dependent scaling efficiency. Modern thinking-enabled models like OpenAI's o1 series and Qwen3 have internalized these capabilities, with Qwen3's innovation lying in transparent, controllable thinking processes through budget mechanisms. Medical AI has similarly evolved from early rule-based expert systems like MYCIN \cite{shortliffe2012computer} to modern deep learning approaches. Recent breakthroughs include Med-PaLM \cite{singhal2025toward} achieving expert-level performance, GatorTron \cite{yang2022large} demonstrating domain-specific pretraining benefits, and subsequent systems like Med-Gemini \cite{saab2024capabilities}, AMIE \cite{tu2025towards}, and diagnostic-focused models \cite{mcduff2023towards}. The rapid adoption of ChatGPT and GPT-4 in medical contexts \cite{kung2023performance,lee2023benefits,eriksen2023use} has highlighted both opportunities and challenges. Specialized medical LLMs have emerged including multimodal systems like LLaVA-Med \cite{li2023llava}, Med-Flamingo \cite{moor2023medflamingo}, and BiomedGPT \cite{luo2024biomedgpt}, domain-specific models like Radiology-GPT \cite{cao2025development}, Clinical Camel \cite{toma2023clinical}, and MedAlpaca \cite{han2023medalpaca}, as well as open-source initiatives including PMC-LLaMA \cite{wu2024pmc}, MEDITRON \cite{chen2023meditron}, BioMistral \cite{labrak2024biomistral}, and HuatuoGPT \cite{chen2024huatuogpt}. However, these systems typically lack explicit thinking mechanisms necessary for verifiable clinical reasoning. The theoretical foundation for our work draws from scaling law research initiated by Kaplan et al. \cite{kaplan2020scaling} and refined by Hoffmann et al. \cite{hoffmann2022chinchilla}, which established relationships between model performance and computational resources during training:
$$L(N) = \left(\frac{N_c}{N}\right)^{\alpha_N},$$
where $L(N)$ is the test loss for a model with $N$ parameters, $N_c$ and $\alpha_N$ are fitted constants. Our contribution extends these principles to inference-time thinking budget allocation, proposing:
$$L(N, T_b) = \left(\frac{N_c}{N}\right)^{\alpha_N} \cdot \left(\frac{T_c}{T_b + T_0}\right)^{\alpha_T},$$
where $T_b$ is the thinking budget, $T_0$ is a baseline thinking requirement, and $\alpha_T$ captures thinking efficiency. Traditional efficiency optimization techniques including quantization \cite{mathur2024mind}, pruning \cite{han2015pruning}, and knowledge distillation \cite{khan2025optimizing} focus on uniform computational reduction, whereas our thinking budget approach enables dynamic resource allocation based on task complexity. The medical domain provides rich evaluation opportunities through benchmarks like MedQA \cite{jin2021medqa}, PubMedQA \cite{jin2019pubmedqa}, and MultiMedQA \cite{singhal2023large}, though our stratified evaluation across 15 specialized datasets offers more granular insights into reasoning requirements across medical specialties and complexity levels. Adaptive computation concepts from Graves' ACT \cite{graves2016adaptive} and recent Mixture of Experts models \cite{zhang2024mpmoe} provide architectural precedents, but thinking budgets offer user-controllable inference-level adaptation specifically for reasoning tasks. The critical importance of interpretability in medical AI has been extensively documented \cite{abacha2024review,loh2023application,alsaleh2024enhancing,chen2024explainable,farah2024promise,pfeifer2024how}. While traditional interpretability methods including attention visualization \cite{vaswani2017attention}, gradient-based explanations \cite{sundararajan2017integrated}, and concept activation vectors \cite{kim2018tcav} provide post-hoc insights, thinking mechanisms fundamentally align with medical diagnostic processes by making reasoning transparent and observable, enabling calibration to match expected clinical reasoning depth across different scenarios. Recent systematic reviews have highlighted the growing need for explainable AI in healthcare \cite{muhammad2024unveiling,rosenbacke2024explainable}, with particular emphasis on building trust between clinicians and AI systems. The challenge of hallucination in medical contexts \cite{ji2023survey} further underscores the importance of verifiable reasoning processes. Retrieval-augmented generation approaches \cite{lewis2020retrieval,borgeaud2022improving} have shown promise in grounding medical responses, with systems like ALMANAC \cite{zakka2024almanac} and recent benchmarking efforts \cite{xiong2024benchmarking} demonstrating improved factuality. Chain-of-verification \cite{dhuliawala2023chain} and similar techniques aim to reduce hallucination through explicit verification steps, aligning with our thinking budget approach.

Recent work further consolidates the view that inference-time reasoning control, rather than purely architectural scaling or static prompting, is becoming a central axis for improving reliability and safety in medical language models. Several studies emphasize that next-generation medical LLMs must support explicit, adjustable reasoning depth to accommodate heterogeneous clinical scenarios, ranging from routine factual recall to complex differential diagnosis, while maintaining predictable computational cost and interpretability \cite{lai2025provable,park2025medalignplus}. In parallel, advances in test-time scaling and controllable inference have highlighted the limitations of fixed decoding strategies, motivating adaptive computation mechanisms that dynamically allocate reasoning resources based on task complexity and uncertainty \cite{gschwind2025ai}. These developments collectively suggest a shift from post-hoc explainability and prompt-based reasoning toward native, budget-aware thinking processes as a first-class design principle for medical AI systems. Our work aligns with this emerging direction by formalizing thinking budget allocation as an inference-time control mechanism and empirically evaluating its impact across diverse medical specialties and reasoning demands.

%% file: content/experiment.tex
\section{Experimental Design}
\label{sec:experiment}
 
We conducted comprehensive experiments using two prominent model families that support thinking mechanisms. The Qwen3 series \cite{bai2023qwen} encompasses models ranging from 1.7B to 235B parameters, featuring native thinking budget support that allows direct control over reasoning depth through the \texttt{thinking\_budget} parameter. The DeepSeek-R1 series \cite{liu2025deepseek} comprises models from 1.5B to 70B parameters. Unlike Qwen3, DeepSeek-R1 models lack native thinking budget control, necessitating a novel methodological approach to ensure fair comparison.

To address this challenge, we developed a truncation-based approach for DeepSeek-R1 models that maintains experimental rigor while enabling controlled comparison. Our method begins by generating complete thinking processes using each model's native capabilities, then systematically truncates the thinking content to match desired budget constraints. The process involves extracting thinking content between \texttt{<think>} and \texttt{</think>} tags and truncating to specified token limits (64, 128, 256, 512, 1024). For the ``None'' budget condition, all thinking content is removed entirely, while the ``inf'' budget retains the complete original thinking process. This methodology ensures that all thinking content originates from the same model being evaluated, maintaining consistency across experimental conditions.

The thinking budget parameters we employed span a wide range to capture the full spectrum of reasoning behaviors. The ``None'' condition provides baseline performance without any thinking process, while token budgets of 64, 128, 256, 512, and 1024 allow us to observe how performance scales with increasing computational investment. The ``inf'' (unlimited) budget reveals each model's maximum reasoning capability when unconstrained by computational limits.

Our evaluation encompasses 15 carefully curated medical datasets spanning diverse specialties and difficulty levels, building upon established medical AI benchmarks \cite{jin2021medqa,zhang2024medqa,chen2023medprompt,brin2023comparing}. The datasets are stratified into two primary difficulty tiers reflecting real-world medical practice hierarchies. Chief physician level datasets represent the top 200 most challenging questions in each specialty, encompassing advanced cardiovascular disease diagnosis and treatment, complex gastrointestinal disorders, blood disorders and malignancies, complicated infectious disease cases, advanced kidney disease management, complex neurological conditions, and advanced pulmonary diseases. These datasets test the limits of medical reasoning with cases that would challenge experienced specialists.

Attending physician-level datasets comprise the top 200 difficulty questions, focusing on standard clinical scenarios including routine cardiovascular cases, common gastrointestinal disorders, basic hematological conditions, standard infectious diseases, common kidney disorders, basic neurological conditions, and common respiratory diseases. This tier represents the breadth of cases encountered in typical clinical practice. Additionally, we include the American Journal of Kidney Diseases (AJKD) dataset, which features research-oriented nephrology questions derived from cutting-edge medical literature, providing a unique perspective on how models handle emerging medical knowledge.

Each dataset contains 200 carefully validated multiple-choice questions designed to test different aspects of medical reasoning, from pattern recognition and differential diagnosis to treatment selection and prognostic assessment. This comprehensive coverage ensures our findings generalize across the full spectrum of medical decision-making scenarios.

Our primary evaluation metric is accuracy, defined as the percentage of correctly answered questions:
$$\text{Accuracy} = \frac{\text{Number of Correct Answers}}{\text{Total Number of Questions}} \times 100\%.$$
To quantify performance improvements, we define:
$$\Delta P(T_b) = P(T_b) - P(T_0),$$
where $P(T_b)$ represents performance at thinking budget $T_b$ and $P(T_0)$ is baseline performance without thinking.

We also analyze the thinking token ratio:
$$R_{\text{thinking}} = \frac{T_{\text{output}}}{T_{\text{input}}},$$
which indicates how models scale their reasoning depth relative to question complexity.

This straightforward metric provides clear insights into model performance while enabling direct comparison across different thinking budget configurations. Beyond raw accuracy, we conduct extensive analyses of thinking token distributions to understand how models allocate computational resources, performance scaling patterns to identify optimal budget ranges through the optimization problem:
$$\max_{T_b} \frac{\text{Accuracy}(T_b)}{\text{Cost}(T_b)} \text{ subject to } T_b \leq T_{\text{budget}},$$
and saturation points where additional thinking yields diminishing returns, characterized by $\frac{\partial \text{Accuracy}}{\partial T_b} < \epsilon$ for small $\epsilon > 0$.

The experimental protocol ensures rigorous and reproducible evaluation across all conditions. For each model-dataset-budget combination, we sequentially process all 200 questions, applying the appropriate thinking budget control mechanism (native API for Qwen3 or truncation for DeepSeek-R1). Model responses are generated with controlled thinking depth, and answers are extracted and validated against ground truth. Throughout this process, we meticulously record both accuracy metrics and thinking token statistics for subsequent analysis.

Our statistical analyses encompass multiple dimensions of model behavior. We examine token length distributions for both questions and thinking processes to understand the relationship between input complexity and reasoning depth. Performance comparisons across model sizes within each family reveal scaling behaviors, while cross-family analyses at fixed thinking budgets illuminate architectural differences. We also identify efficiency frontiers for each dataset, determining the optimal trade-offs between computational investment and performance gains.

Our comprehensive evaluation encompassed 630 distinct experimental conditions, systematically varying across model families, model sizes, datasets, and thinking budget levels. This extensive experimental design enables robust conclusions about thinking budget mechanisms in medical reasoning contexts.

%% file: content/result.tex
\section{Results}
\label{sec:results}

\begin{figure}[htbp]
\centering
\includegraphics[width=0.7\textwidth]{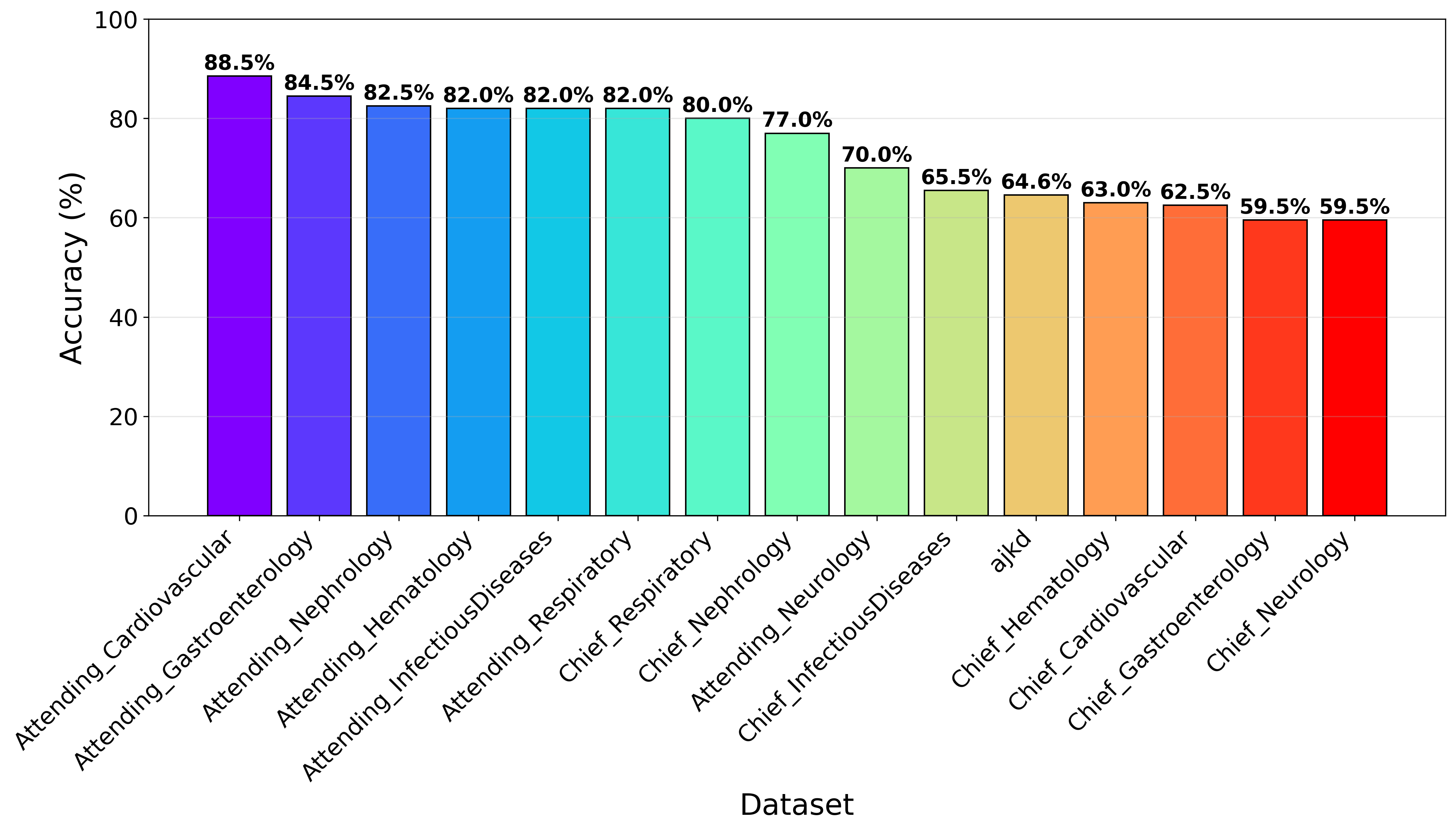}
\caption{Dataset difficulty ranking based on Qwen3:235B performance with unlimited thinking budget. Accuracy ranges from 88.5\% (Attending Cardiovascular) to 59.5\% (Chief Neurology), illustrating the varying complexity of medical reasoning tasks across specialties.}
\label{fig:dataset_difficulty}
\end{figure}

We evaluated model performance across 15 medical datasets spanning multiple specialties and difficulty levels. Figure \ref{fig:dataset_difficulty} presents the overall difficulty landscape, revealing that attending physician level datasets consistently achieve higher accuracies (70-88.5\%) compared to chief physician level datasets (59.5-80\%). This pattern validates our dataset stratification approach and highlights the substantial performance gap between routine clinical questions and complex diagnostic scenarios.

Both Qwen3 and DeepSeek-R1 families demonstrate consistent performance improvements with increased thinking budgets. Larger models outperform smaller ones (e.g., Qwen3:235B achieves 62.5\% vs 45\% for 1.7B on chief cardiovascular), with most models showing rapid improvement up to 256 tokens and diminishing returns beyond 512 tokens.

Without thinking, Qwen3 models generally outperform DeepSeek-R1 counterparts, but both families show 15-20\% improvements for smaller models and 5-10\% for larger ones when thinking is enabled.

Token distribution analysis shows the chief physician questions average 400-600 tokens versus 250-350 for the attending level. Models adapt their thinking depth with ratios from 2.3x to 6.2x of question length, generating reasoning traces from 168 to over 15,000 tokens when unconstrained.

The empirical data reveal consistent scaling relationships across all datasets, which we model as:
$$\text{Accuracy}(T_b, M_s) = \alpha \log(T_b + 1) + \beta \log(M_s) + \gamma + \epsilon,$$
where $T_b$ is the thinking budget, $M_s$ is the model size in billions of parameters, $\alpha \approx 0.08$, $\beta \approx 0.12$, $\gamma$ varies by dataset difficulty, and $\epsilon \sim \mathcal{N}(0, \sigma^2)$ represents model-specific variance with $\sigma \approx 0.02$.

The marginal utility of thinking tokens follows:
$$\frac{\partial \text{Accuracy}}{\partial T_b} = \frac{\alpha}{T_b + 1}.$$
This derivative reveals why performance improvements diminish with larger budgets. At $T_b = 255$, the marginal improvement is approximately $\frac{0.08}{256} \approx 0.0003$ per token, while at $T_b = 31$ it is $\frac{0.08}{32} \approx 0.0025$ per token—an 8× difference in efficiency.

Three distinct efficiency regimes emerge from our analysis: a high-efficiency regime (0-256 tokens) characterized by rapid accuracy improvements with efficiency $E > 0.0003$ suitable for real-time applications, a balanced regime (256-512 tokens) offering optimal performance-cost tradeoffs with $0.0001 < E < 0.0003$ for routine clinical decision support, and a high-accuracy regime (512+ tokens) where marginal improvements with $E < 0.0001$ may justify additional computational costs only for critical diagnostic tasks.

The efficiency frontier can be formally characterized by the Pareto-optimal set:
$$\mathcal{F}^* = \{(T_b, A(T_b)) : \nexists T'_b < T_b \text{ such that } A(T'_b) \geq A(T_b)\},$$
where $A(T_b)$ represents accuracy at thinking budget $T_b$. The optimal thinking budget for a given cost constraint $C_{\max}$ is:
$$T^*_b = \arg\max_{T_b} \left\{A(T_b) : C(T_b) \leq C_{\max}\right\},$$
where the cost function $C(T_b) = c_0 + c_1 T_b$ includes both fixed overhead $c_0$ and per-token cost $c_1$.

\begin{figure}[htbp]
\centering
\includegraphics[width=1\linewidth]{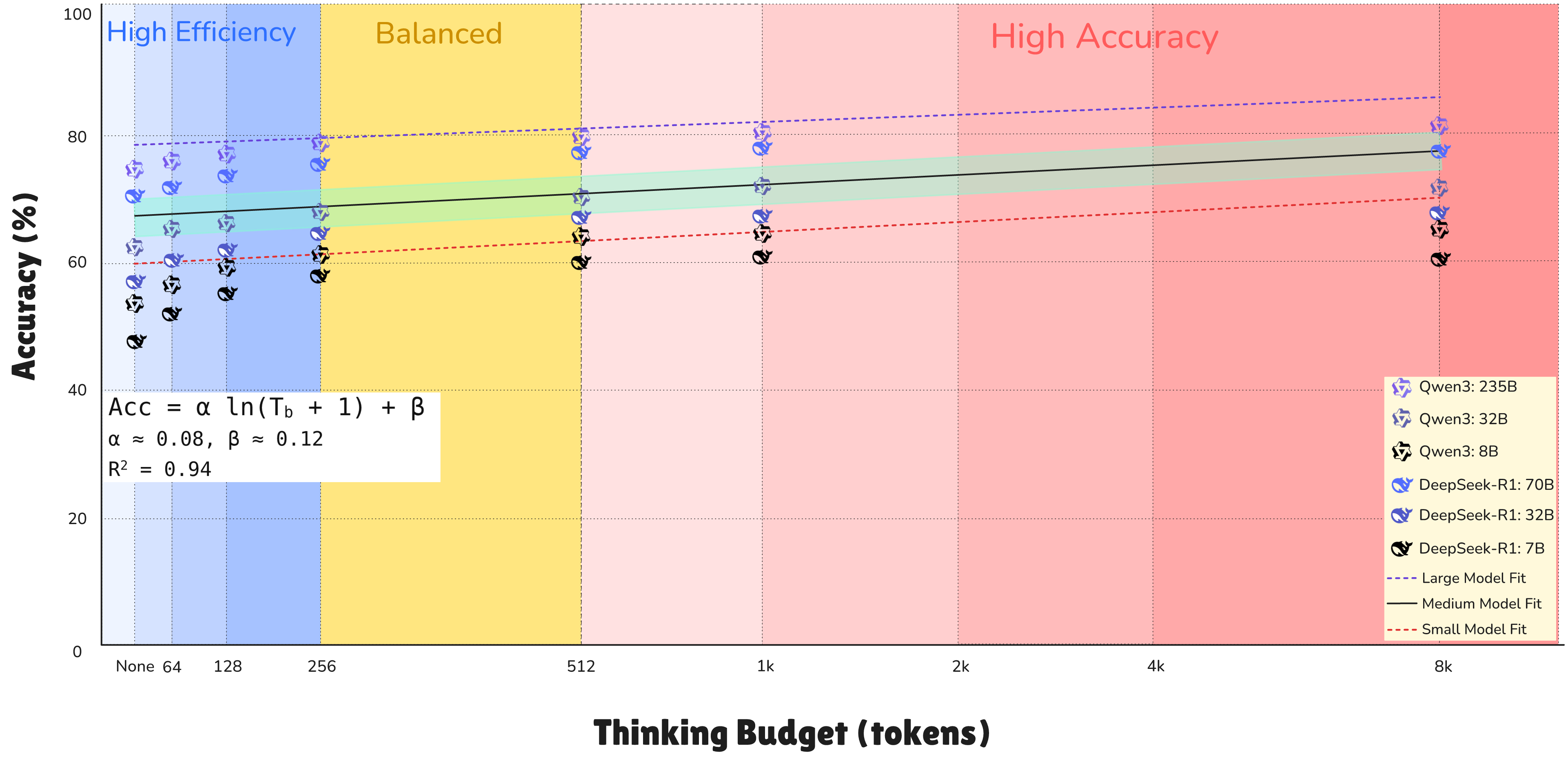}
\caption{Logarithmic scaling of accuracy with thinking budget across model families. Scatter points show empirical results for six representative models (3 from each family), with fitted regression lines demonstrating the scaling law $\text{Accuracy} = \alpha \ln(T_b + 1) + \beta \ln(M_s) + \gamma$. Background shading indicates efficiency regimes: high efficiency (blue, 0-256 tokens), balanced (yellow, 256-512 tokens), and high accuracy (red, 512+ tokens). The 95\% confidence interval (gray dotted lines) shows the consistency of the scaling relationship. Smaller models exhibit steeper slopes ($\alpha \approx 0.095$) compared to larger models ($\alpha \approx 0.08$), confirming that thinking budget provides greater relative benefits for capacity-constrained models.}
\label{fig:thinking_budget_scaling}
\end{figure}

Figure \ref{fig:thinking_budget_scaling} visualizes the logarithmic scaling relationship between thinking budget and accuracy across different model sizes. The scatter plot clearly demonstrates three key findings: (1) the logarithmic nature of performance improvements, with rapid gains in the 0-256 token range followed by diminishing returns; (2) the consistent scaling pattern across both model families, validating our unified scaling law; and (3) the inverse relationship between model size and thinking budget benefit, where smaller models show steeper improvement curves.

\begin{figure}[htbp]
\centering
\begin{subfigure}[b]{0.24\textwidth}
\centering
\includegraphics[width=\textwidth]{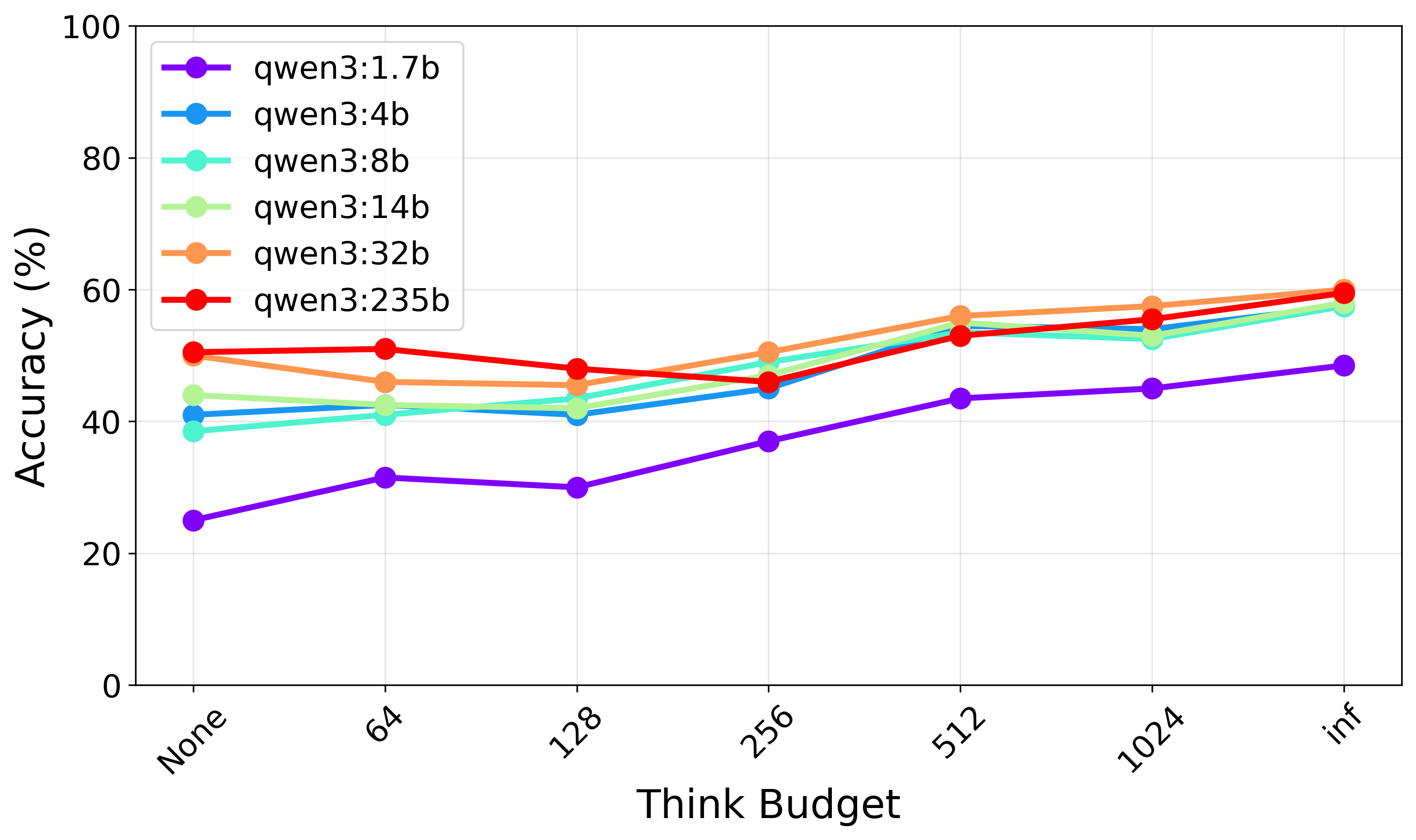}
\caption{Chief Neurology (Qwen3)}
\end{subfigure}
\hfill
\begin{subfigure}[b]{0.24\textwidth}
\centering
\includegraphics[width=\textwidth]{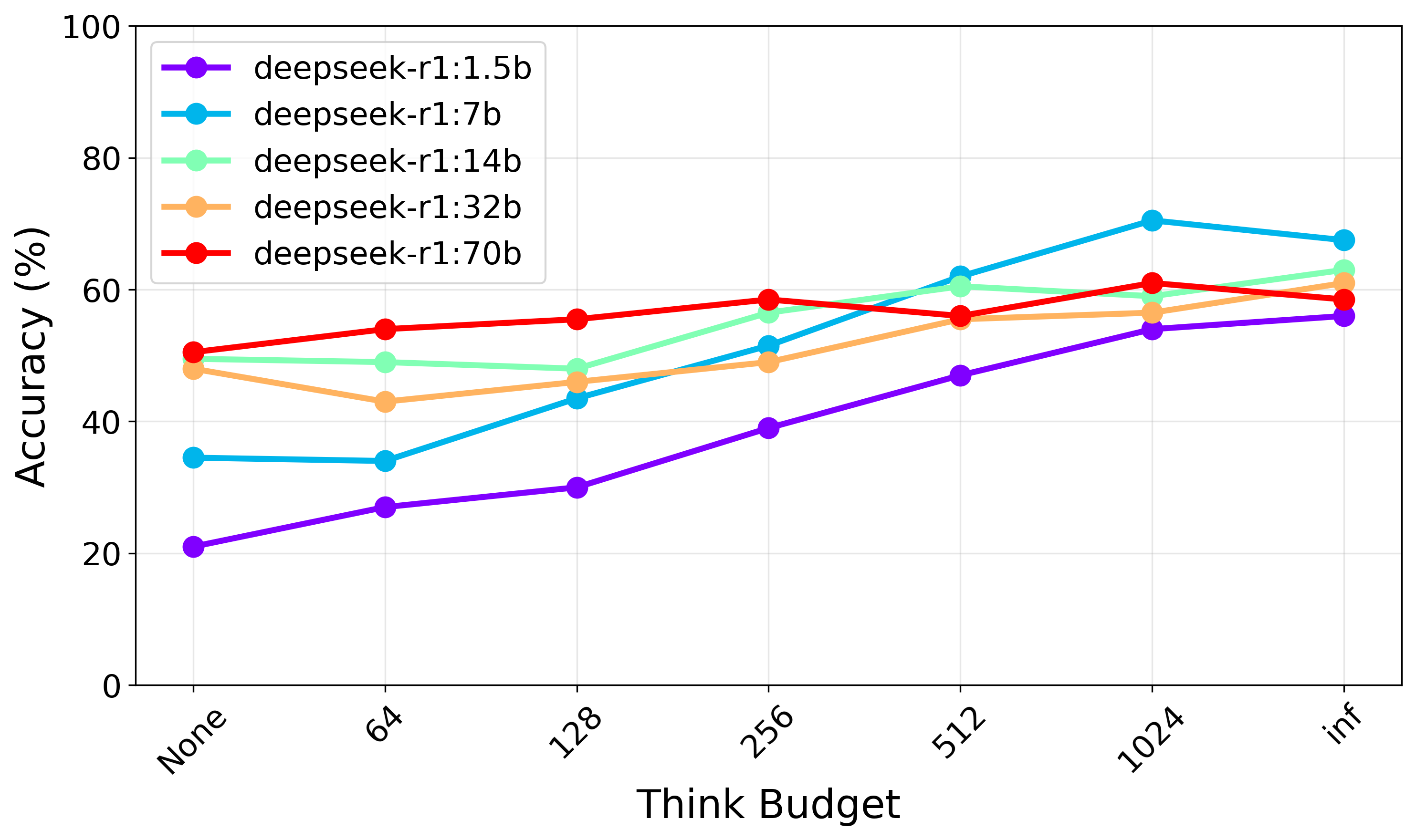}
\caption{Chief Neurology (DeepSeek-R1)}
\end{subfigure}
\hfill
\begin{subfigure}[b]{0.24\textwidth}
\centering
\includegraphics[width=\textwidth]{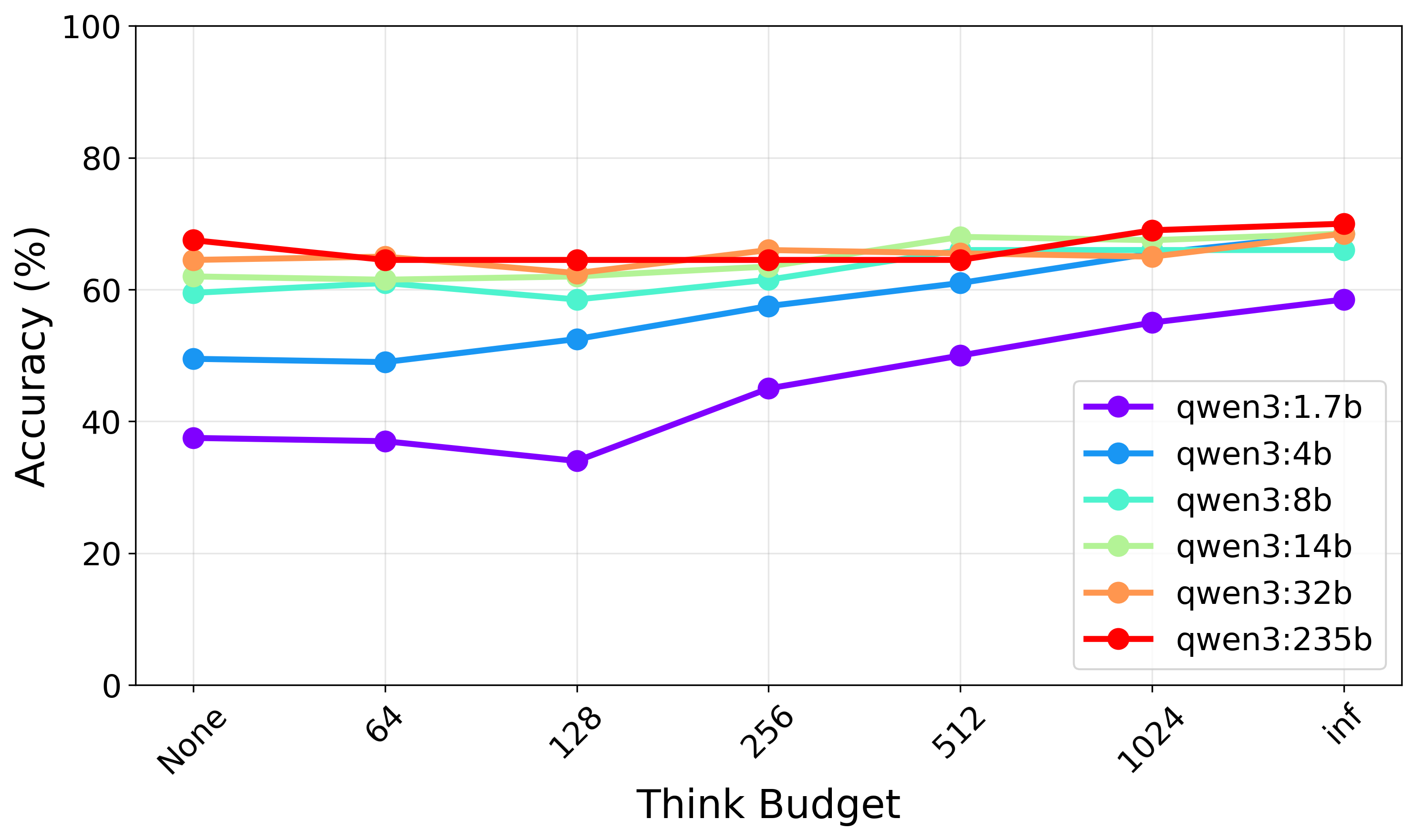}
\caption{Attending Neurology (Qwen3)}
\end{subfigure}
\hfill
\begin{subfigure}[b]{0.24\textwidth}
\centering
\includegraphics[width=\textwidth]{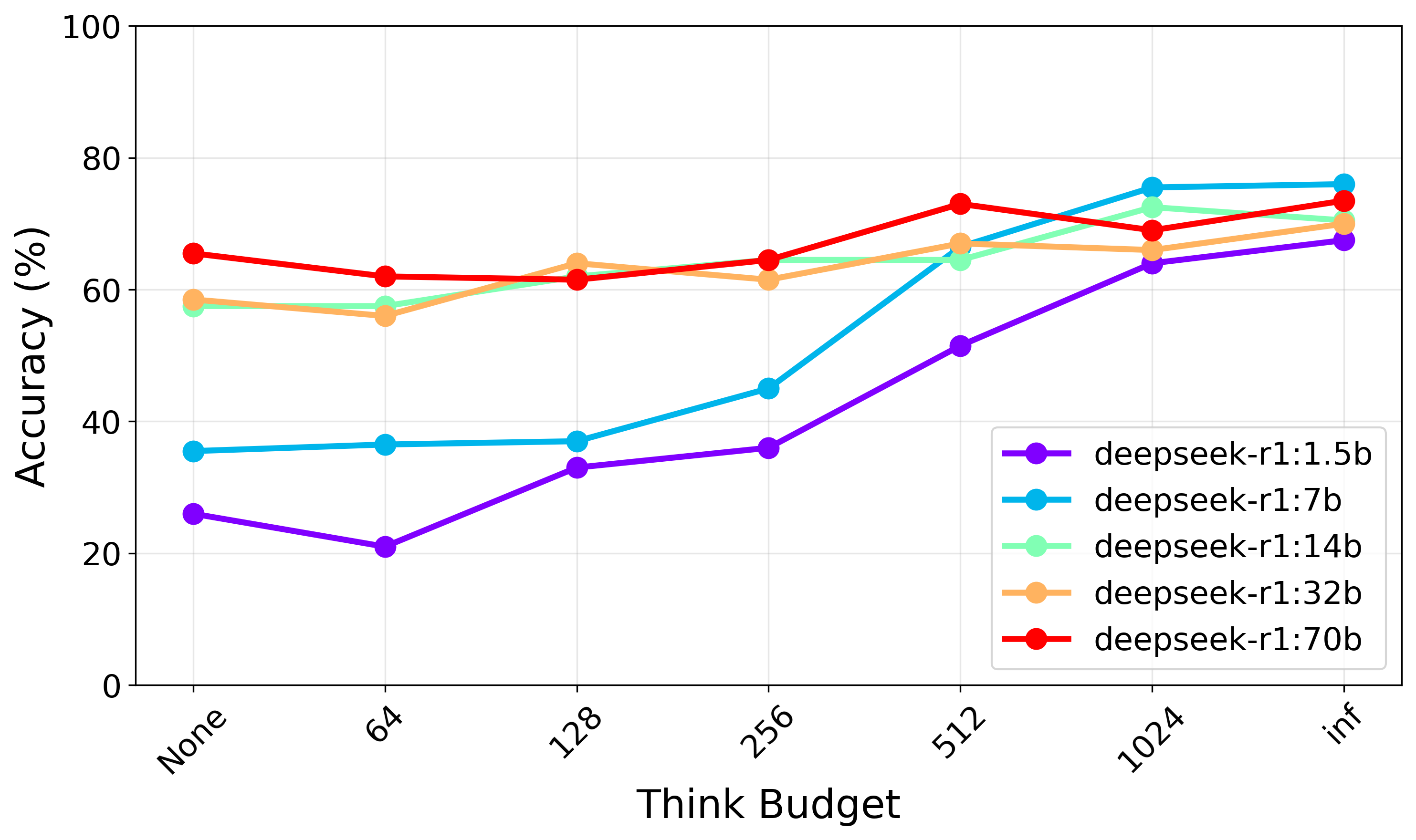}
\caption{Attending Neurology (DeepSeek-R1)}
\end{subfigure}

\vspace{0.5em}

\begin{subfigure}[b]{0.24\textwidth}
\centering
\includegraphics[width=\textwidth]{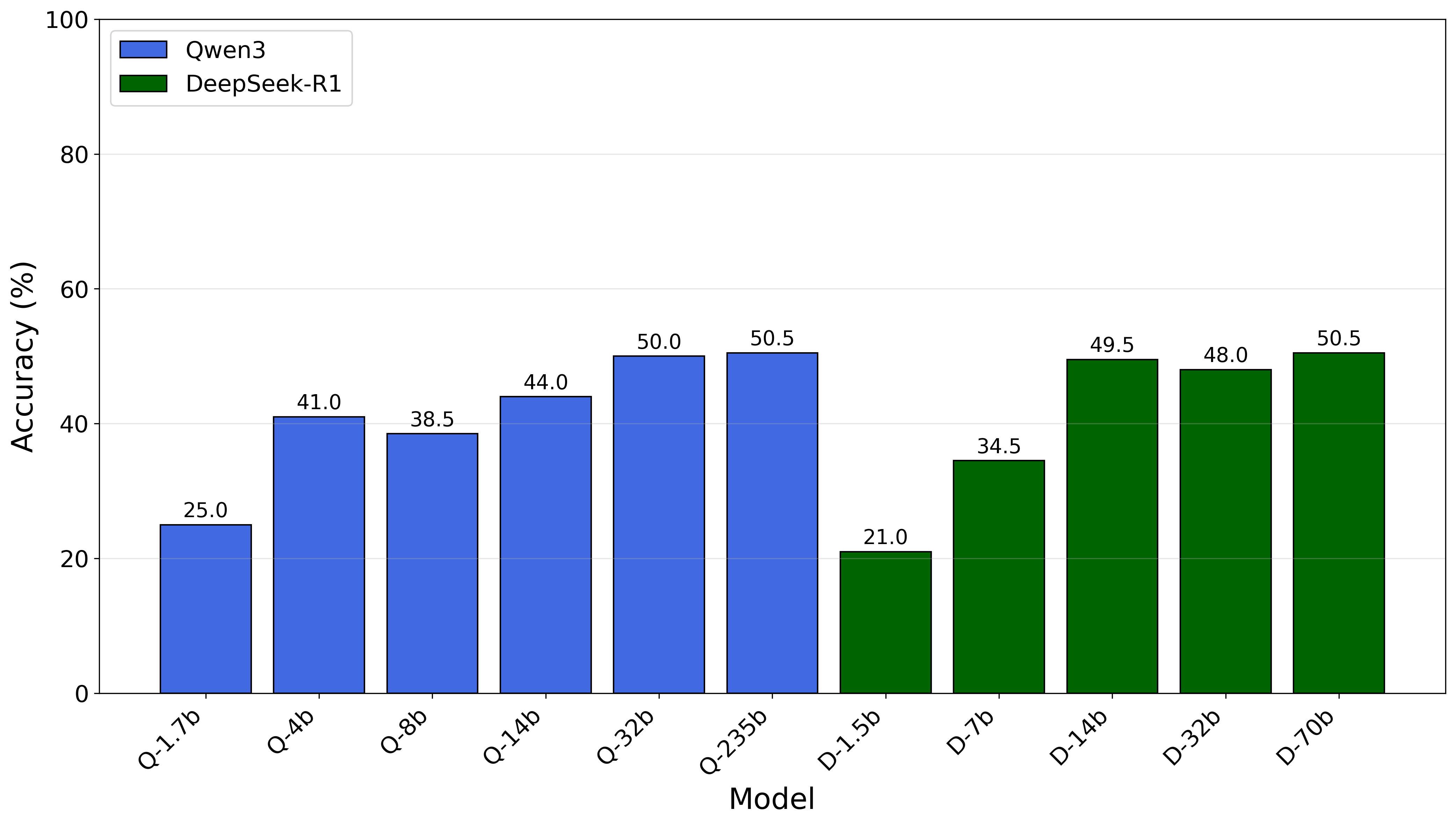}
\caption{Chief Neurology (No thinking)}
\end{subfigure}
\hfill
\begin{subfigure}[b]{0.24\textwidth}
\centering
\includegraphics[width=\textwidth]{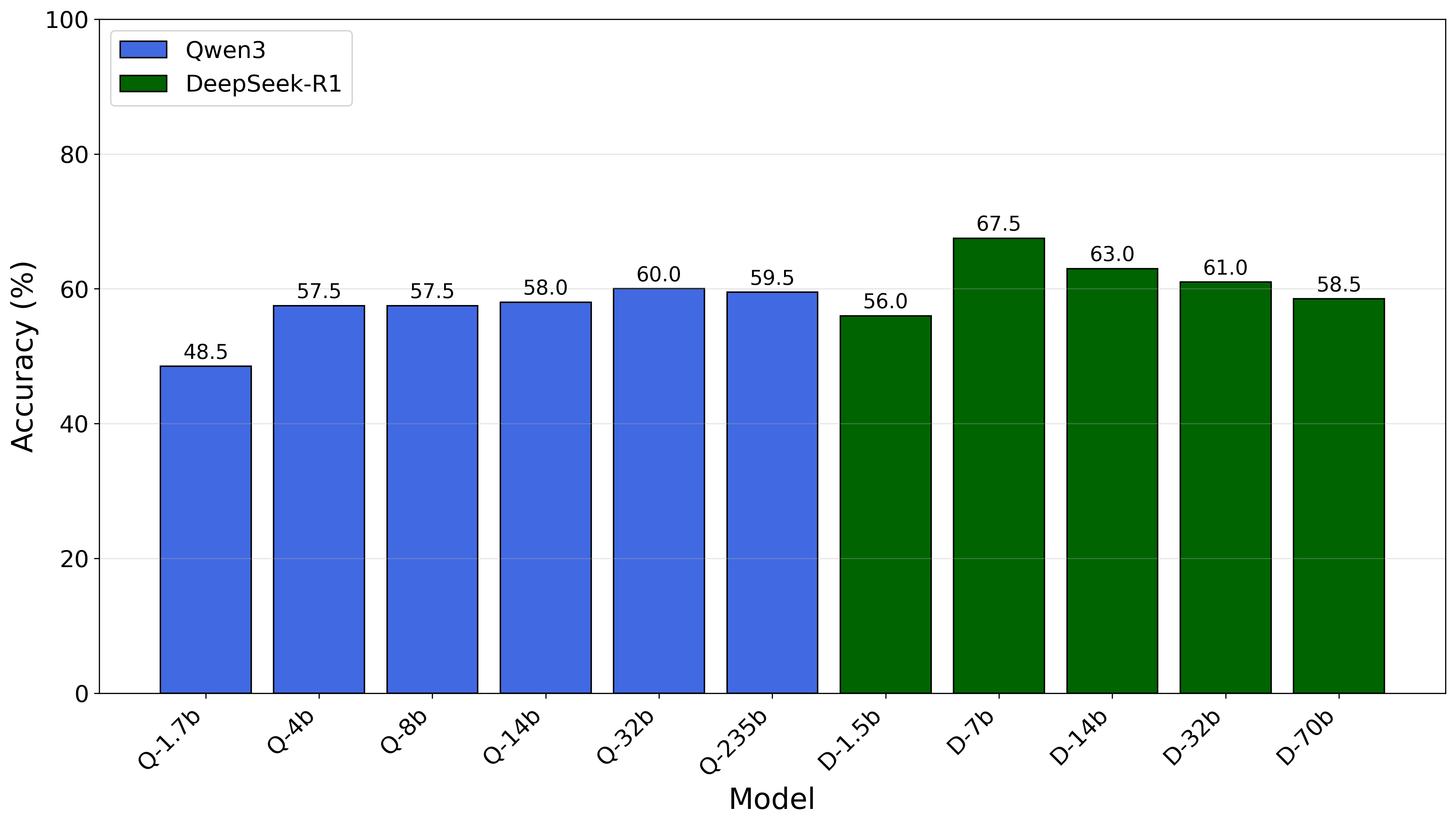}
\caption{Chief Neurology (Unlimited)}
\end{subfigure}
\hfill
\begin{subfigure}[b]{0.24\textwidth}
\centering
\includegraphics[width=\textwidth]{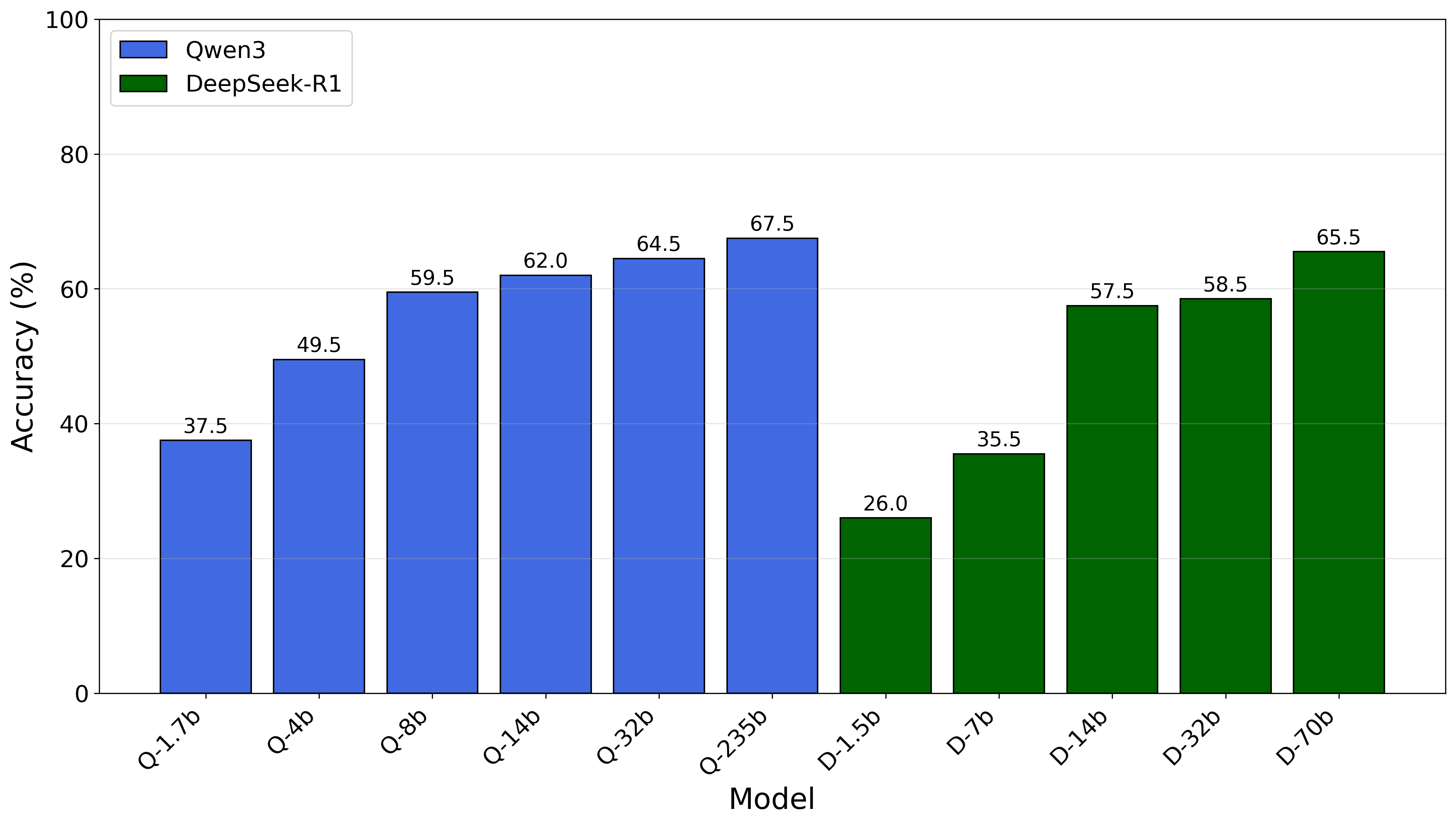}
\caption{Attending Neurology (No thinking)}
\end{subfigure}
\hfill
\begin{subfigure}[b]{0.24\textwidth}
\centering
\includegraphics[width=\textwidth]{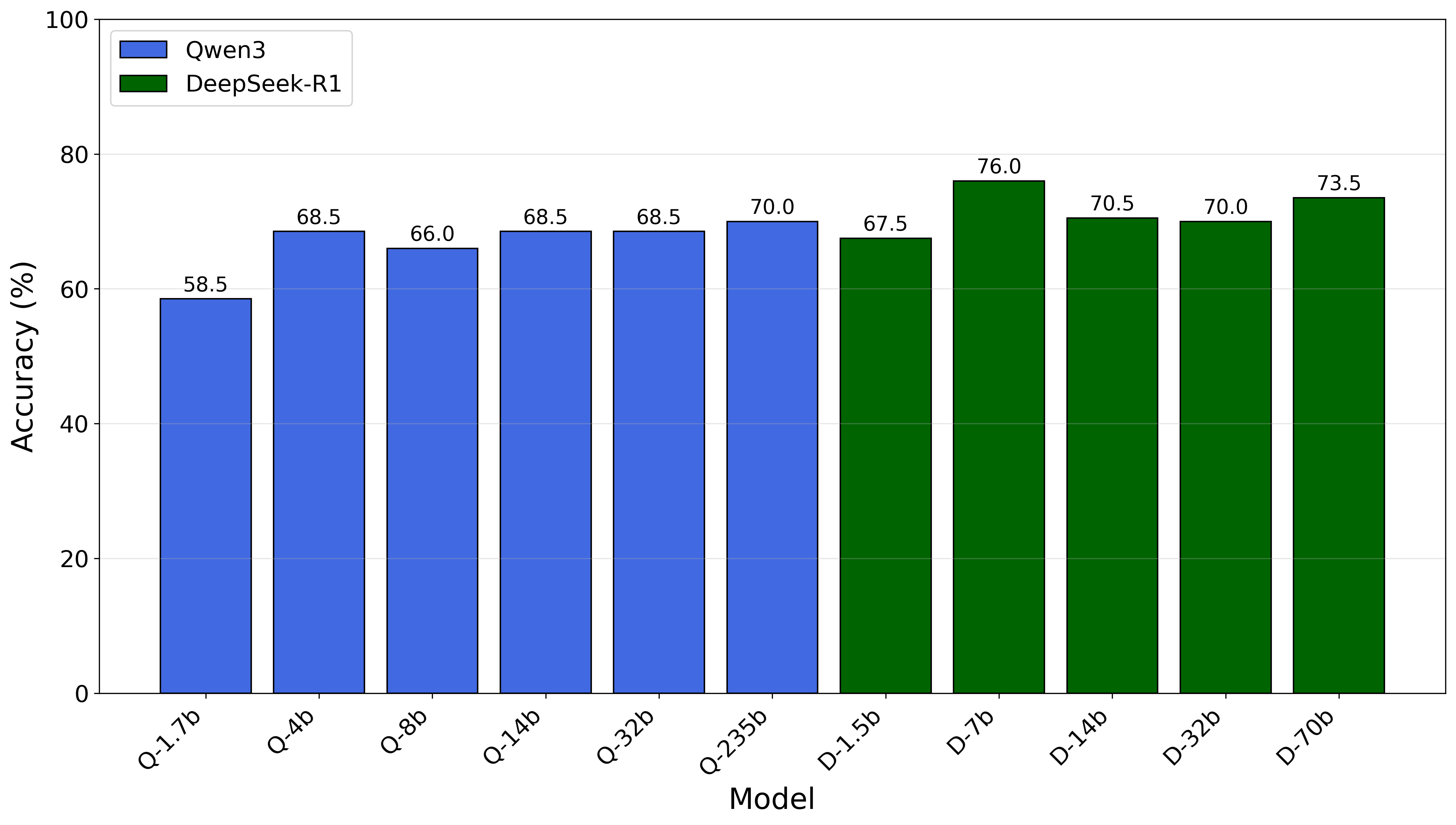}
\caption{Attending Neurology (Unlimited)}
\end{subfigure}

\caption{Comprehensive results for Neurology datasets (most challenging). The top row shows performance curves across thinking budgets for both model families at the chief and attending levels. Bottom row compares model performance without thinking (None) versus unlimited thinking (inf), revealing substantial improvements from thinking processes, especially for smaller models.}
\label{fig:neurology_results}
\end{figure}

\begin{figure}[htbp]
\centering
\begin{subfigure}[b]{0.24\textwidth}
\centering
\includegraphics[width=\textwidth]{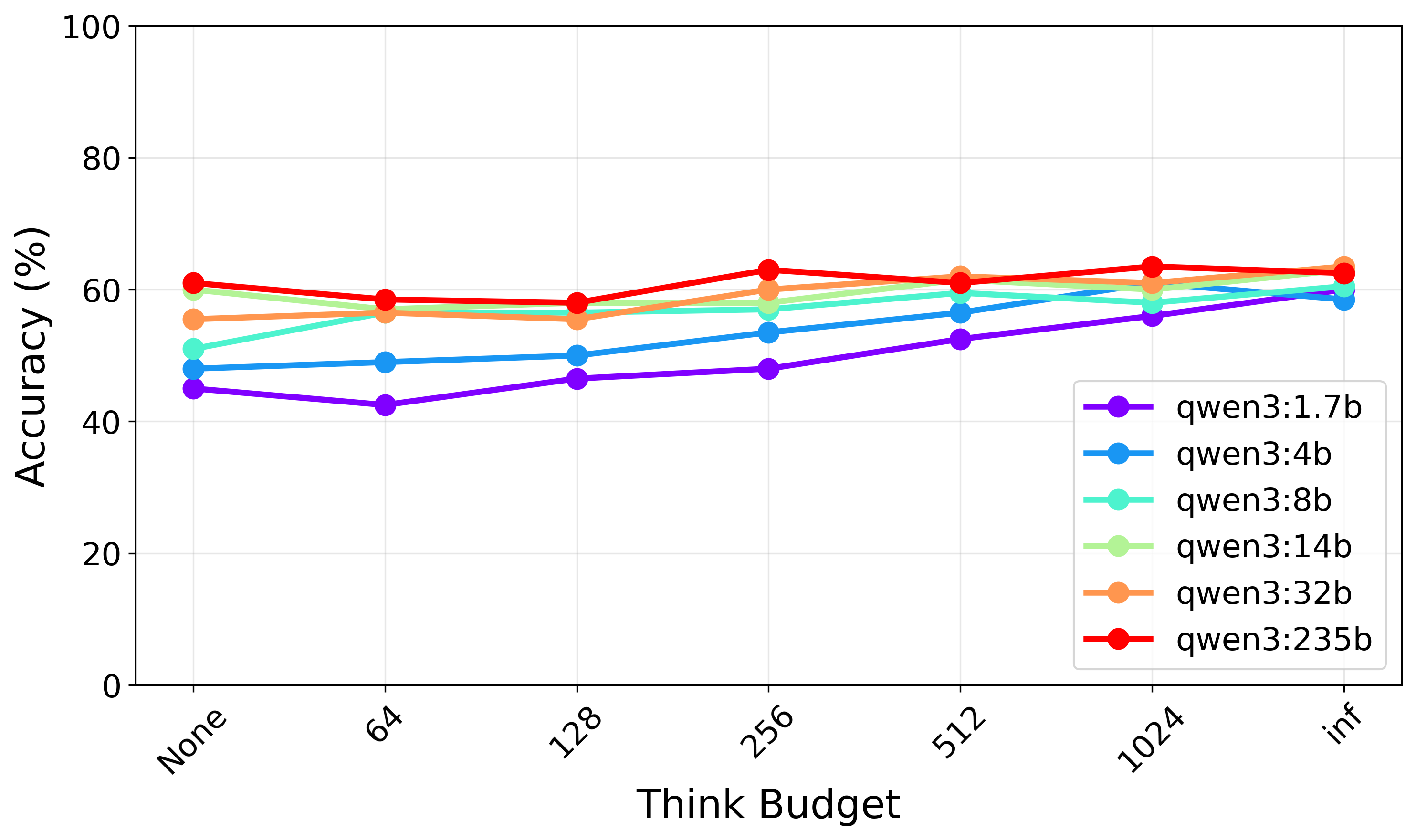}
\caption{Chief Cardiovascular (Qwen3)}
\end{subfigure}
\hfill
\begin{subfigure}[b]{0.24\textwidth}
\centering
\includegraphics[width=\textwidth]{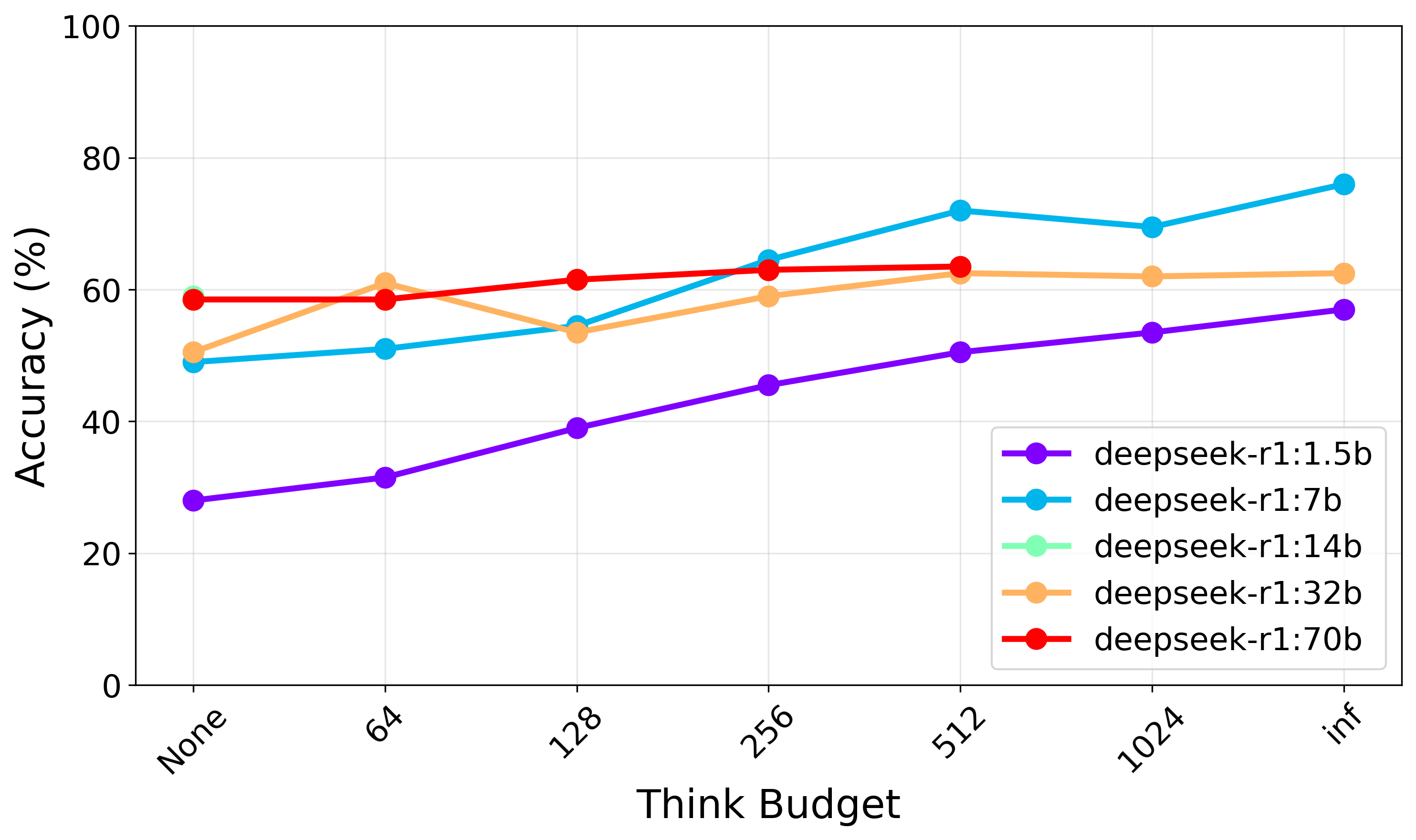}
\caption{Chief Cardiovascular (DeepSeek-R1)}
\end{subfigure}
\hfill
\begin{subfigure}[b]{0.24\textwidth}
\centering
\includegraphics[width=\textwidth]{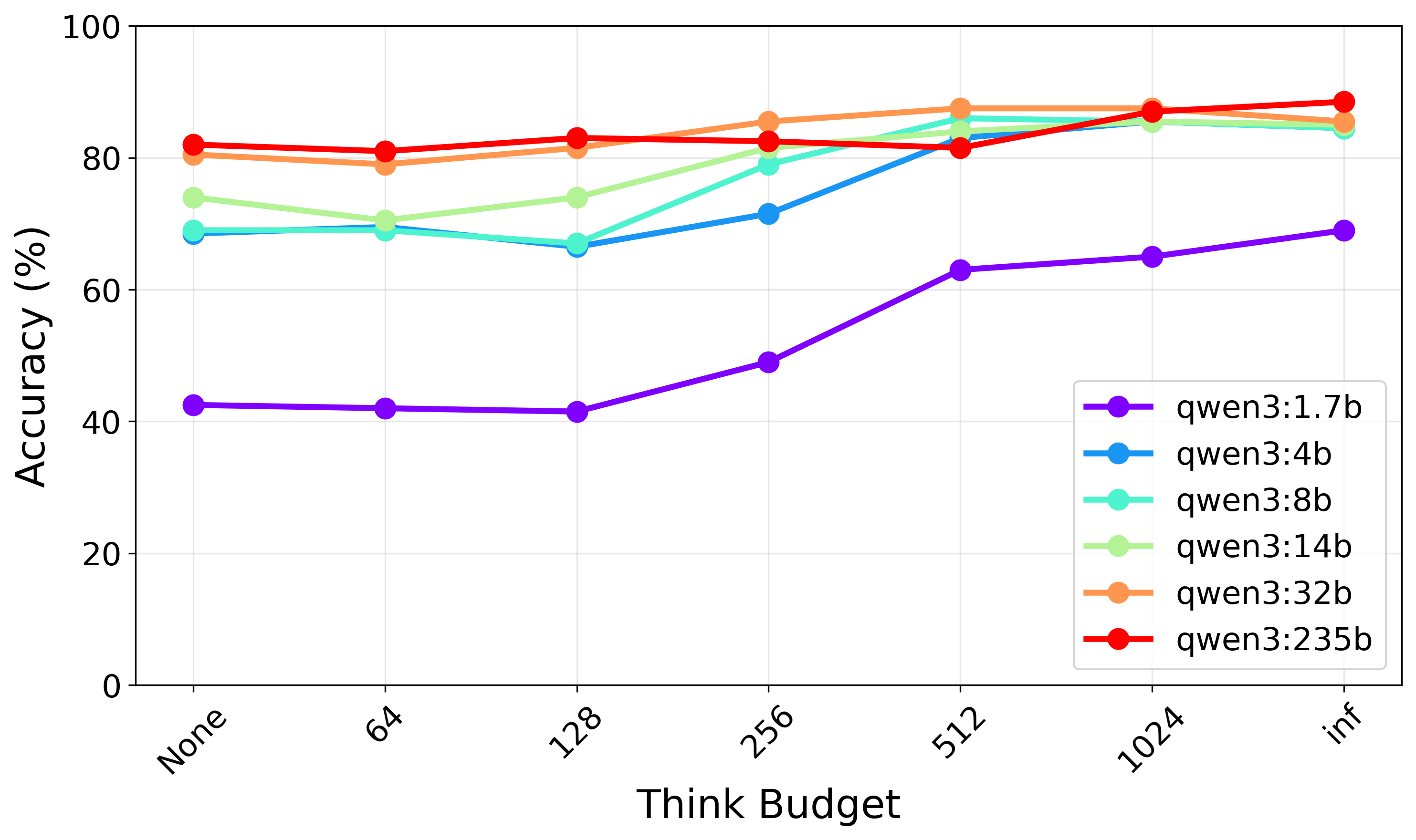}
\caption{Attending Cardiovascular (Qwen3)}
\end{subfigure}
\hfill
\begin{subfigure}[b]{0.24\textwidth}
\centering
\includegraphics[width=\textwidth]{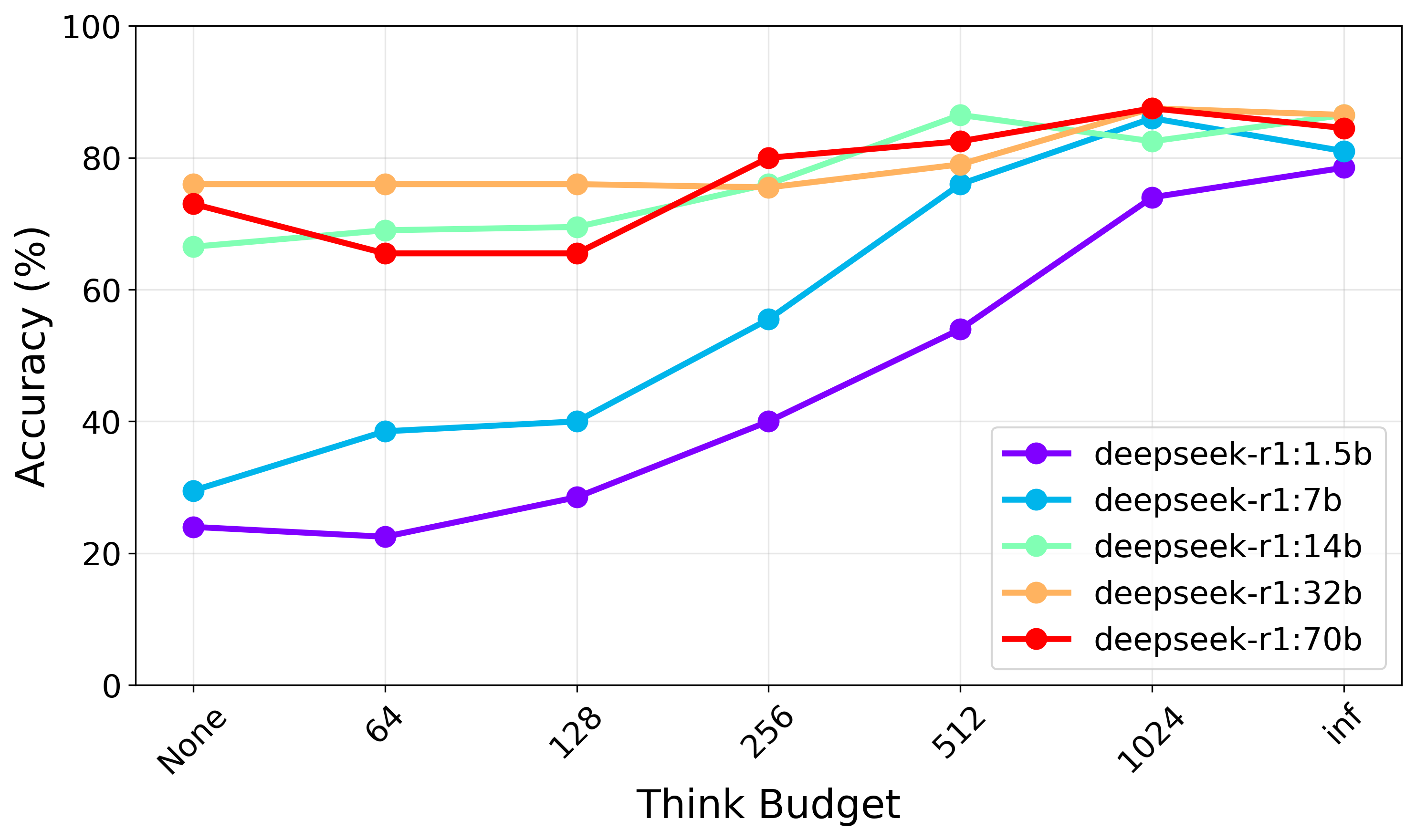}
\caption{Attending Cardiovascular (DeepSeek-R1)}
\end{subfigure}

\vspace{0.5em}

\begin{subfigure}[b]{0.24\textwidth}
\centering
\includegraphics[width=\textwidth]{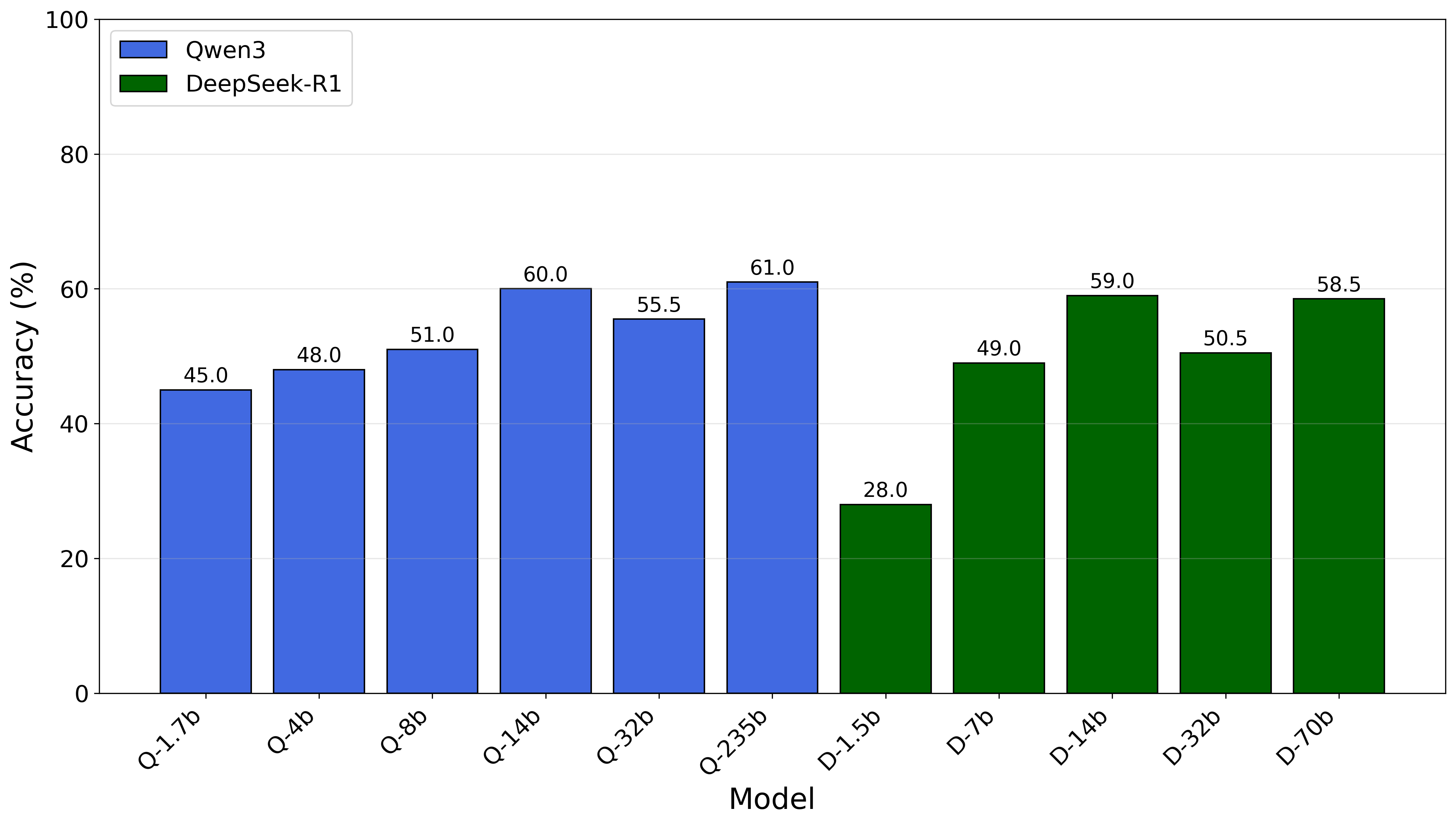}
\caption{Chief Cardiovascular (No thinking)}
\end{subfigure}
\hfill
\begin{subfigure}[b]{0.24\textwidth}
\centering
\includegraphics[width=\textwidth]{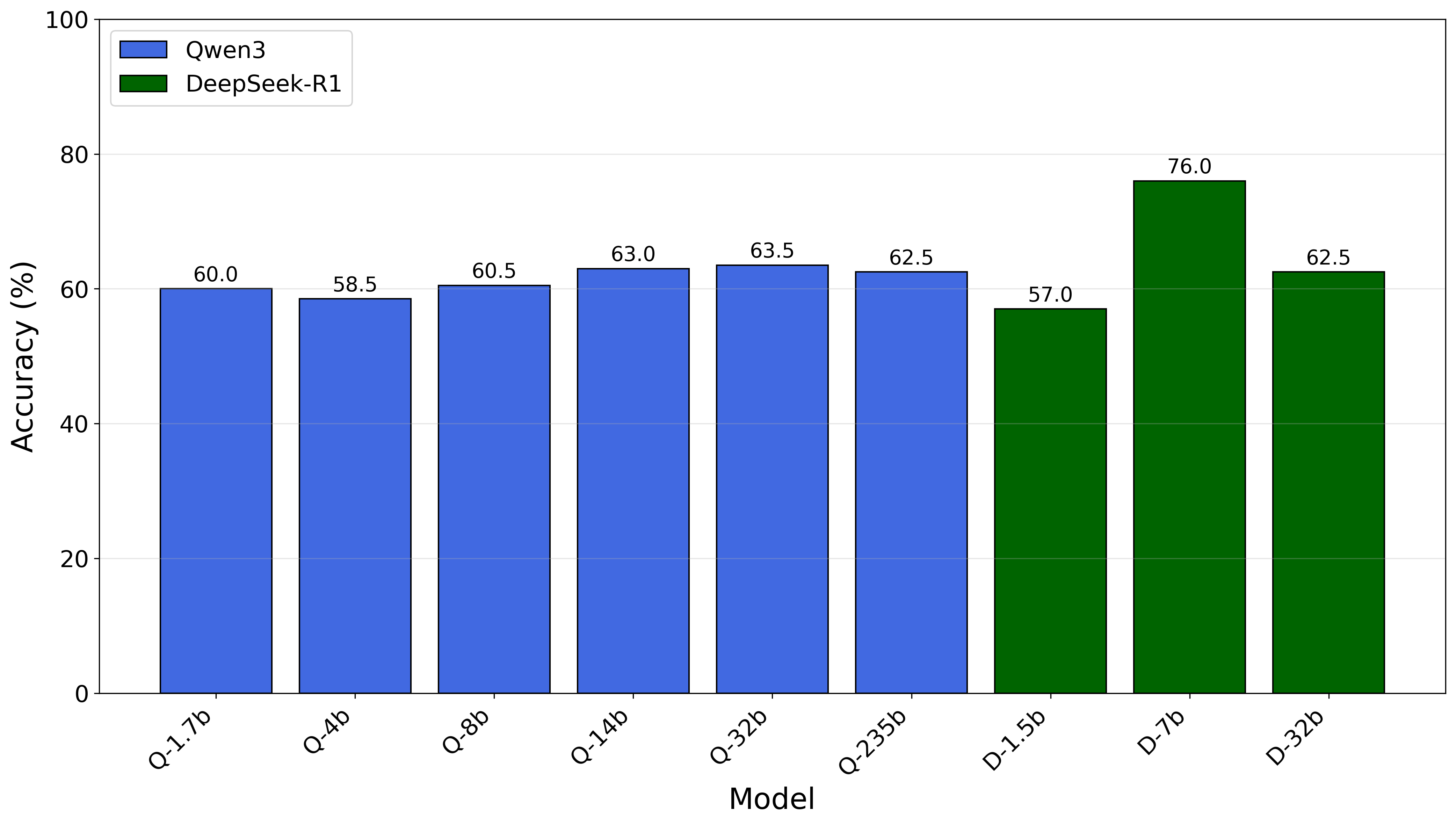}
\caption{Chief Cardiovascular (Unlimited)}
\end{subfigure}
\hfill
\begin{subfigure}[b]{0.24\textwidth}
\centering
\includegraphics[width=\textwidth]{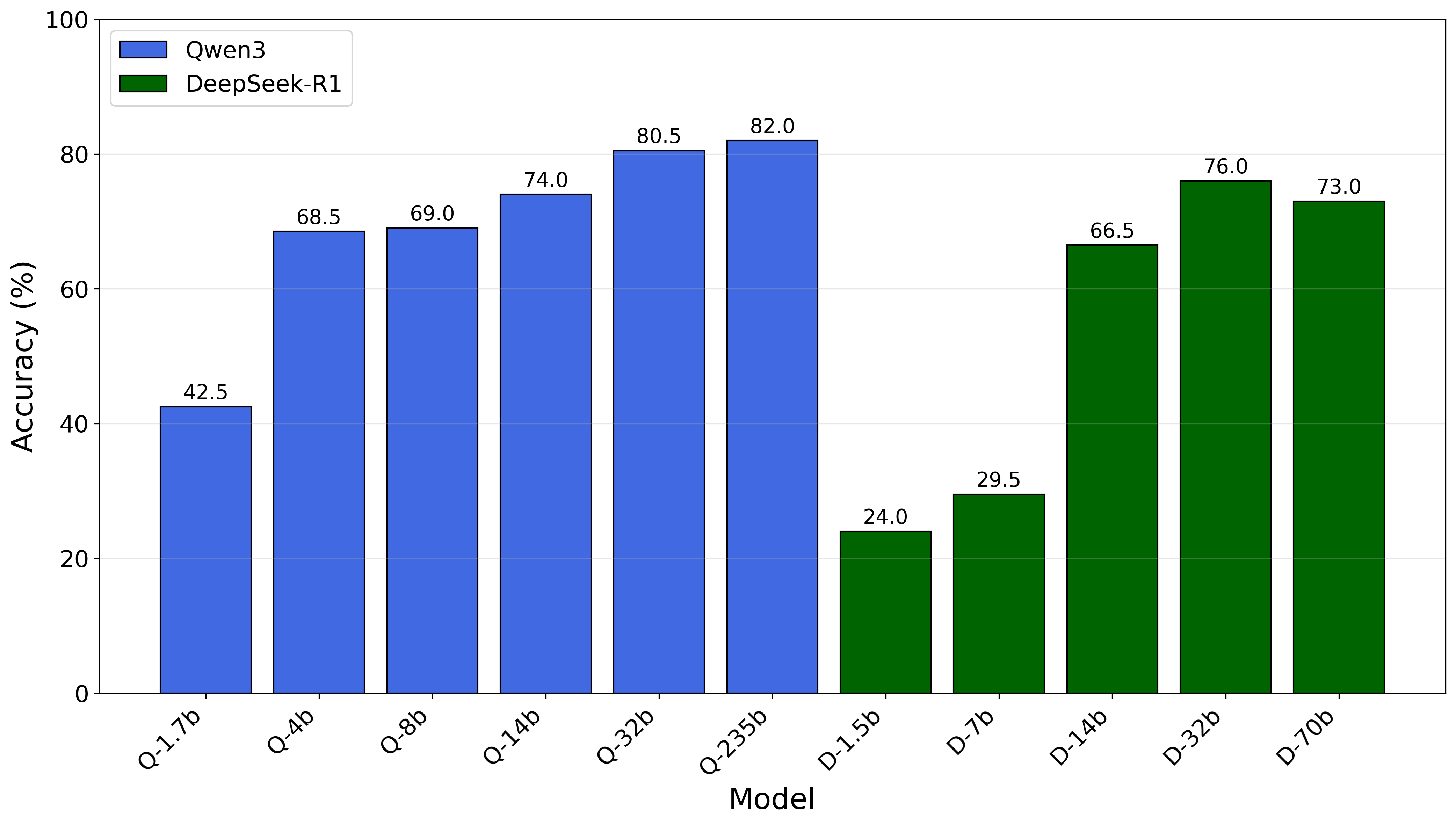}
\caption{Attending Cardiovascular (No thinking)}
\end{subfigure}
\hfill
\begin{subfigure}[b]{0.24\textwidth}
\centering
\includegraphics[width=\textwidth]{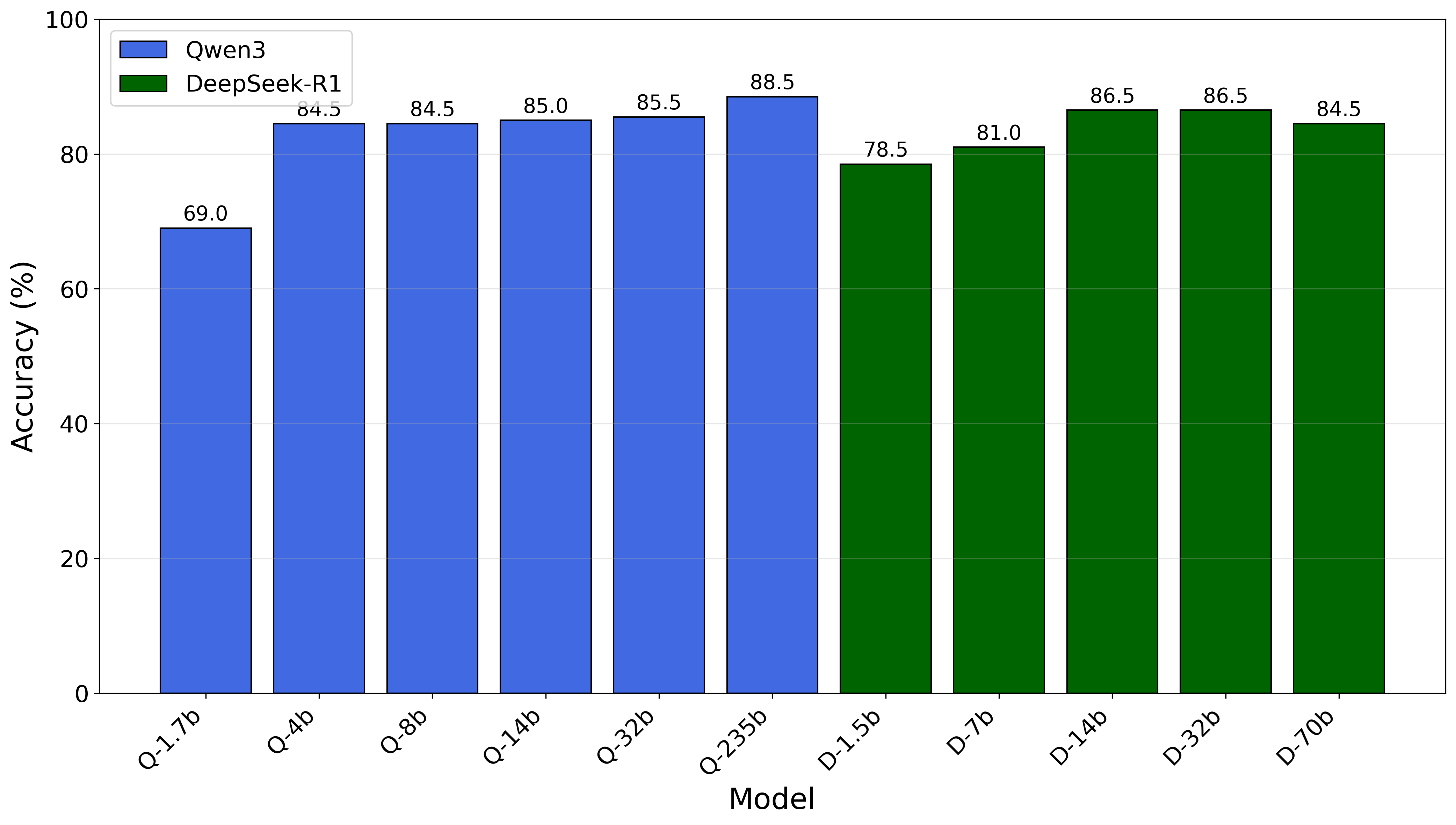}
\caption{Attending Cardiovascular (Unlimited)}
\end{subfigure}

\caption{Comprehensive results for Cardiovascular datasets (easiest). Performance curves show rapid saturation at lower thinking budgets compared to Neurology. The bottom row demonstrates that even without thinking, models achieve relatively high accuracy on these simpler tasks, with thinking providing more modest improvements.}
\label{fig:cardiovascular_results}
\end{figure}

Figures \ref{fig:neurology_results} and \ref{fig:cardiovascular_results} show results for the most challenging (Neurology) and easiest (Cardiovascular) datasets. Complete results are in \ref{sec:complete_results}.

Our comprehensive evaluation reveals that thinking universally improves performance by 5-20\% across all models, with most tasks achieving optimal performance at 256-512 tokens. Notably, smaller models benefit more from extended thinking, demonstrating the compensatory effect of reasoning depth for limited model capacity. The consistency between Qwen3's native API and our DeepSeek-R1 truncation method validates the generalizability of thinking budget concepts across different model architectures. Specialty-specific patterns emerge clearly, with neurology and gastroenterology requiring substantially deeper thinking processes compared to cardiovascular medicine, reflecting the varying complexity of medical reasoning across domains. These findings collectively establish thinking budget control as an essential mechanism for optimizing medical AI systems.

%% file: content/limitation.tex
\section{Discussion}\label{sec:limitations}

This study presents a comprehensive analysis of thinking budget mechanisms in medical reasoning tasks, establishing explicit scaling laws between inference-time computational resources and reasoning quality. Through extensive experiments across 15 medical datasets and two major model families, we demonstrate that controllable reasoning depth constitutes an effective paradigm for optimizing AI performance in clinical applications.

Several fundamental insights emerge from our findings. First, we empirically observe a logarithmic scaling relationship between allocated thinking budget and model accuracy, with clear efficiency frontiers emerging around 256–512 tokens for most medical tasks. This result provides actionable guidance for healthcare institutions seeking to balance computational cost against diagnostic accuracy. The consistency of scaling behavior across Qwen3’s native thinking budget interface and our truncation-based strategy for DeepSeek-R1 further validates the generality of thinking budget principles across model architectures, including those without native thinking support.

Our analysis reveals pronounced domain-specific differences in computational demand. Neurology and gastroenterology consistently require deeper reasoning processes, whereas cardiovascular and respiratory medicine often achieve strong performance under moderate thinking budgets. These patterns can be attributed to several factors: (1) the lack of one-to-one correspondence between clinical symptoms and pathological lesions; (2) delayed therapeutic responses that limit the diagnostic value of short-term clinical trials; and (3) the predominance of training data centered on common disease presentations, resulting in insufficient standardized reasoning protocols for rare or complex conditions. Consequently, complex diagnostic tasks necessitate more reasoning steps and greater integration of heterogeneous clinical information.

Importantly, this study does not evaluate differential diagnosis or downstream diagnostic planning. Prior work has reported suboptimal LLM performance in differential diagnosis settings, often due to incomplete dataset annotations, missing negative symptoms, and insufficient modeling of disease interrelationships. Moreover, LLMs tend to adopt conservative treatment recommendations that may diverge from real-world clinical decision-making, underscoring the necessity of clinician oversight for aggressive therapeutic strategies.

We further observe a complementary relationship between model scale and reasoning depth. Smaller models benefit disproportionately from extended thinking budgets, while larger models can sustain strong performance even under constrained budgets. This finding suggests that optimal deployment strategies should jointly consider model selection and reasoning budget configuration rather than relying on scale alone. From a practical standpoint, while prior studies have demonstrated the diagnostic accuracy of LLMs in high-pressure and time-sensitive clinical scenarios, significant limitations remain, particularly in disease classification and severity stratification. In addition, emerging evidence indicates that the inclusion of nondeterministic sociodemographic variables—such as race, gender, income level, and educational background—may introduce bias or harmful outputs. Frameworks such as EquityGuard have been proposed to mitigate the influence of sensitive attributes, and existing literature highlights the importance of secondary evaluation layers and active reasoning strategies to enhance reliability in clinical decision support systems.

Based on our empirical findings, we propose a dynamic thinking budget allocation strategy:
$$
T_b =
\begin{cases}
64\text{–}128 & \text{for routine screening (complexity score } < 0.3\text{)} \\
256\text{–}512 & \text{for standard diagnosis (complexity score } 0.3\text{–}0.7\text{)} \\
1024+ & \text{for complex cases (complexity score } > 0.7\text{)}
\end{cases}
$$
This strategy aligns computational effort with clinical complexity and provides a principled alternative to fixed inference-time budgets.

From an interpretability and accountability perspective, explicit thinking processes address a critical barrier to clinical adoption by enabling physicians to inspect and verify model reasoning \cite{pfeifer2024how,alsaleh2024enhancing}. Transparent reasoning aligns closely with established clinical workflows, particularly in domains such as imaging and pathology, where diagnostic reasoning is grounded in identifiable lesions and structured evidence. Moreover, when reasoning traces constitute explicit evidence, they facilitate diagnosis and treatment stratification, risk assessment, and post hoc auditing. In cases of dispute, such transparency enables attribution of errors to model bias or data limitations. Nevertheless, explanation overload remains an open challenge, highlighting the need for structured and adaptive reasoning representations that balance informativeness with usability.

Despite these insights, several limitations merit discussion. Our truncation-based approach for DeepSeek-R1 models, while ensuring consistency by using each model’s own generated content, may not perfectly replicate native thinking budget control. The truncation process could interrupt reasoning chains at suboptimal points, potentially underestimating model capabilities under constrained budgets. We characterize this effect as:
$$
\text{Loss}_{\text{truncation}} = \int_{T_b}^{T_{\text{full}}} p(t) \cdot v(t) \, dt,
$$
where $p(t)$ denotes the probability that critical reasoning occurs at token position $t$, and $v(t)$ represents the value of that reasoning. Future work should explore truncation strategies that better preserve reasoning coherence by minimizing $\text{Loss}_{\text{truncation}}$.

The scope of our evaluation is limited to multiple-choice questions. Real clinical reasoning often involves open-ended problem solving, differential diagnosis generation, and treatment planning, which cannot be fully captured by MCQ formats. Furthermore, our focus on medical reasoning constrains generalizability; thinking budget requirements may differ substantially in other domains such as mathematics, programming, or general reasoning, motivating cross-domain validation.

From a technical standpoint, we do not explicitly model computational cost in terms of wall-clock latency or energy consumption. The relationship between thinking tokens and actual inference cost is non-linear and hardware-dependent, influenced by batching strategies and system-level optimizations. Moreover, although our results suggest domain-specific optimal thinking budgets, we currently lack a principled method for automatically determining these thresholds. Developing adaptive mechanisms that infer appropriate reasoning depth from task characteristics remains an important direction for future research.

Looking ahead, several promising avenues emerge. Adaptive thinking mechanisms that dynamically adjust reasoning depth based on complexity indicators could further improve efficiency. Additionally, replacing monolithic reasoning traces with structured, multi-stage reasoning pipelines incorporating intermediate validation points may enhance both accuracy and usability, providing natural checkpoints for clinical review.

Overall, our findings establish thinking budget mechanisms as a viable and effective approach for controlling reasoning depth in medical AI systems. The consistent scaling laws observed across diverse medical specialties suggest fundamental properties governing how language models allocate computational resources for complex reasoning. As reasoning-enabled models continue to advance, intelligent inference-time resource allocation will be critical for responsible deployment in high-stakes clinical settings.

%% file: content/conclusion.tex
\section{Conclusion}

In conclusion, this work demonstrates that thinking budget mechanisms provide a principled and practical approach to controlling reasoning depth in medical language models. By establishing explicit relationships between computational investment and reasoning quality, we show that inference-time resource allocation can be systematically optimized to meet diverse clinical demands while preserving interpretability and efficiency. Rather than relying solely on larger models or increased training data, future medical AI systems can benefit from dynamic, budget-aware reasoning strategies that align computational effort with clinical complexity. As reasoning-enabled language models continue to advance, thinking budget control will play a central role in bridging the gap between benchmark performance and reliable real-world clinical deployment.
\section{Acknowledgment}

\subsection{Author Contributions}
Z. Bi, L. Chen, J. Song—conceptualization, coding, methodology, and original draft; T. Wang, C. Liang, K. Chen—software, formal analysis, visualization; J. Huang, H. Luo, E. Ge, Z. Wei, H. Liu, C. Tian, J. Guan, J. Yeong, X. Song, Y. Xu—resources, data, and review; P. Wang, J. Hao—supervision, funding acquisition, and final approval. All authors read and approved the manuscript.

\subsection{Statements and Declarations}

Ethical approval: No human subjects were involved in this study; ethical approval was not sought.

Competing interests: The authors declared no potential conflicts of interest concerning the research, authorship, and/or publication of this article.

Funding: This research was supported by The  High-level Talents in the Affiliated Hospital of Guangdong Medical University [GCC20220013, GCC2022046].

Data availability statement: The data that support the findings of this study are available from the corresponding author upon reasonable request.

%% file: content/appendix.tex
\section{Sample Medical Reasoning Cases}
\label{sec:medical_cases}

This appendix presents representative examples from our evaluation datasets to illustrate the complexity of medical reasoning tasks and demonstrate how different models approach these problems under varying thinking budgets. The cases selected exemplify the challenging diagnostic scenarios that distinguish expert-level medical reasoning from routine clinical assessment.

\subsection{Case Study 1: Complex Peripheral Neuropathy Diagnosis}

\textbf{Clinical Presentation:} A 39-year-old female patient presented with an 8-month history of progressive peripheral neuropathy. Initial symptoms included bilateral plantar numbness with cotton-like sensation and needle-like pain, accompanied by pitting edema below both knees. The condition progressively worsened, with weakness extending from lower to upper extremities, eventually preventing her from holding utensils or performing daily activities. Physical examination revealed distal muscle atrophy with reduced muscle strength (proximal grade 4-5, distal grade 2-3) and absent deep tendon reflexes. Electrophysiological studies demonstrated decreased nerve conduction velocity and amplitude in both sensory and motor nerves, with amplitude reduction being the predominant feature.

\textbf{Diagnostic Findings:} Laboratory investigations revealed several key abnormalities:
\begin{itemize}
\item Glucose tolerance test: Fasting glucose 6.3 mmol/L, 2-hour glucose 12.0 mmol/L (indicating diabetes)
\item Blood immunofixation electrophoresis: IgA-$\lambda$ positive
\item Cerebrospinal fluid: Protein 2170 mg/L (normal: 0-350 mg/L)
\item Toxicology screening: Negative
\item Comprehensive autoimmune panels, tumor markers, and infectious disease screening: Normal
\end{itemize}

\textbf{Critical Timeline:} The patient developed peripheral neuropathy symptoms 8 months ago, but high blood sugar was only discovered 2 months ago—a crucial detail indicating that the neuropathy preceded the diabetes diagnosis by 6 months.

\textbf{Question:} Based on the above information, the most likely cause of the patient's peripheral neuropathy is:
\begin{itemize}
\item A. Chronic inflammatory demyelinating polyneuropathy
\item B. Diabetic peripheral neuropathy
\item C. Paraproteinemia
\item D. Toxic peripheral neuropathy
\item E. Paraneoplastic syndrome
\item F. Hereditary sensory motor neuropathy
\end{itemize}

\begin{tcolorbox}[title=Model Thinking Process (Qwen3:235B with unlimited budget)]
\emph{The patient is a 39-year-old female with peripheral neuropathy. Symptoms started with numbness and pain in feet, progressing to weakness. EMG showed decreased conduction velocity and amplitude, more so in amplitude, affecting both sensory and motor nerves. Dexamethasone helped with edema but not other symptoms. Negative neostigmine test rules out myasthenia gravis.}

\emph{Looking at lab results: glucose tolerance test shows elevated levels, especially fasting (6.3) and 2-hour (12.0), indicating diabetes. IgA-$\lambda$ is positive, but Bence Jones is negative. The IgA-$\lambda$ might indicate a monoclonal gammopathy, but no evidence of multiple myeloma here.}

\emph{Diabetic neuropathy is a common cause, especially with her glucose levels. However, the EMG findings show more axonal loss (amplitude decrease). CIDP usually presents with demyelination (slowed conduction), which this patient has, but the amplitude is also low. Paraproteinemia (like IgA-$\lambda$) can cause neuropathy... (approximately 2400 words omitted)}

\emph{But wait, the glucose levels are high, but she was diagnosed 2 months ago, and the neuropathy started 8 months ago. Wait, the neuropathy started 8 months ago, but high blood sugar was found 2 months ago. That would mean that her diabetes might have developed after the neuropathy started. Therefore, diabetic neuropathy might not be the primary cause. That's a critical point.}

\emph{If the neuropathy started 8 months ago, and diabetes was diagnosed 2 months ago, then the diabetes is unlikely to be the cause of the neuropathy. That changes things. Because the timeline doesn't support diabetes causing the neuropathy. So that rules out B. Then, the next possible cause is C. Paraproteinemia. Since IgA-$\lambda$ is present, which could be a monoclonal gammopathy. So option C.}

\emph{Therefore, the answer is C.}
\end{tcolorbox}

\subsubsection{Model Performance Analysis}

The performance of different models on this challenging case provides insights into how thinking budget affects diagnostic accuracy. When examined with unlimited thinking budget, the Qwen3:235B model demonstrated sophisticated clinical reasoning that mirrors expert physician analysis. Notably, the model's thinking process exhibits repetitive cycles of reasoning—repeatedly weighing diabetic neuropathy against paraproteinemia, reconsidering the same evidence multiple times, and circling back to previously examined points. This repetitive pattern is characteristic of the model's deliberative process when given unlimited thinking tokens, allowing it to thoroughly explore different diagnostic possibilities even at the cost of redundancy. Despite this repetition, the model ultimately arrives at the correct conclusion through careful temporal analysis, recognizing the critical detail that neuropathy onset preceded diabetes diagnosis by 6 months—a key factor that definitively argues against diabetic neuropathy as the primary etiology.

Table \ref{tab:case_answers} shows how model responses evolved with different thinking budgets:

\begin{table}[htbp]
\centering
\small
\caption{Model responses across thinking budgets for peripheral neuropathy case}
\label{tab:case_answers}
\begin{tabular}{lcccccccc}
\hline
Model & None & 64 & 128 & 256 & 512 & 1024 & inf & Correct \\
\hline
Qwen3:1.7B & B,D,F & B,D,F & B,D,F & B,C,F & B,C,F & B,C & C,B & C \\
Qwen3:4B & C,D,E & D,C & D,E & B,C,D & B,C & B,C & C & C \\
Qwen3:8B & C,F & C,F & C,F & A,B,C & B,C & B,C & C & C \\
Qwen3:14B & B,C,D & B,C,D,E & B,C,D & B,C & B,C & B,C & B,C & C \\
Qwen3:32B & B,C,E & C,E & C,E & A,B,C & B,C & B,C & C & C \\
Qwen3:235B & C & C & C & B,C,E,F & B,C,E & C & C & C \\
\hline
DeepSeek-R1:1.5B & B* & C & A & B & B & B & C & C \\
DeepSeek-R1:7B & A & B & A & B & B & C & C & C \\
DeepSeek-R1:14B & C & C & A,C & A,B,C,E,F & B & B,C & B,C & C \\
DeepSeek-R1:32B & A,C & C & A,C,E & A & B,C & B,C & C & C \\
DeepSeek-R1:70B & D & A,B,D,E,F & A,B,C,D,E,F & A,B,C,D & B & C,B & - & C \\
\hline
\end{tabular}
\end{table}

The analysis reveals critical patterns in diagnostic reasoning. Most models eventually converged on paraproteinemia (C) when given sufficient thinking budget, though smaller models initially favored diabetic neuropathy (B) despite the incompatible timeline. The key insight—that neuropathy preceded diabetes by 6 months—required careful temporal analysis that only emerged with adequate thinking. Larger models like Qwen3:235B maintained diagnostic accuracy even with minimal thinking, suggesting that model capacity provides a foundation that thinking processes can enhance but not replace.

*Note: DeepSeek-R1:1.5B without thinking budget incorrectly identified diabetic peripheral neuropathy, stating "\textit{it is most likely due to diabetic peripheral neuropathy... This condition often presents with chronic inflammation in the feet and hands, leading to nerve damage, especially when blood sugar levels are abnormal.}" This response missed the critical temporal relationship between symptom onset and diabetes diagnosis.

\subsection{Case Study 2: Midbrain Hemorrhage and Gaze Abnormality}

\textbf{Clinical Presentation:} An 84-year-old male with a 10-year history of hypertension suddenly developed left hemiplegia, aphasia, and vomiting while playing mahjong. He had been taking compound antihypertensive tablets with blood pressure usually controlled between 120-140/70-90 mmHg.

\textbf{Physical Examination:}
\begin{itemize}
\item Blood pressure: 220/120 mmHg
\item Consciousness: Drowsy
\item Pupils: Equal size and round, light reflex present
\item Eye position: Both eyes staring at the tip of the nose
\item Motor: No spontaneous movement of left limbs, spontaneous movement of right limbs
\item Reflexes: Bilateral positive pathological signs in lower limbs
\end{itemize}

\textbf{Imaging:} Head CT showed right thalamic hemorrhage breaking into the ventricular system (lateral ventricle, third ventricle, fourth ventricle).

\textbf{Question:} The patient's eyes stare at the tip of the nose because:
\begin{itemize}
\item A. The cortical lateral vision center is damaged
\item B. The subcortical lateral vision center is damaged
\item C. The medial longitudinal fasciculus is damaged
\item D. The midbrain upward vision center is damaged
\item E. The midbrain downward vision center is damaged
\item F. The ventricular system is damaged
\end{itemize}

\begin{tcolorbox}[title=Model Thinking Process (Qwen3:235B with unlimited budget)]
\emph{The patient is an 84-year-old male with hypertension who developed sudden left hemiplegia, aphasia, and vomiting. Key finding: both eyes staring at the tip of the nose, indicating fixed downward gaze (vertical gaze palsy). CT shows right thalamic hemorrhage breaking into ventricles.}

\emph{Analyzing each option:
A/B - Cortical/subcortical lateral vision centers control horizontal, not vertical gaze.
C - MLF damage causes internuclear ophthalmoplegia, not fixed downward gaze.
D - Midbrain upward vision center (riMLF) damage prevents upward gaze, causing eyes to remain downward.
E - Downward center damage would prevent looking down, not cause fixed downward gaze.
F - Ventricular damage alone doesn't cause specific eye movement abnormalities... (approximately 800 words omitted here)}

\emph{The thalamic hemorrhage extending into the third ventricle can damage midbrain structures. When the upward gaze center is damaged, unopposed downward center action causes fixed downward deviation. This is classic for dorsal midbrain syndrome (Parinaud's syndrome).}

\emph{Therefore, the answer is D - the midbrain upward vision center is damaged.}
\end{tcolorbox}

\begin{table}[htbp]
\centering
\caption{Model responses for midbrain hemorrhage case across thinking budgets}
\label{tab:case2_answers}
\begin{tabular}{lcccccccc}
\hline
Model & None & 64 & 128 & 256 & 512 & 1024 & inf & Correct \\
\hline
Qwen3:1.7B & E & E & E & E & D & D & D & D \\
Qwen3:4B & E & E & E & D & D & D & D & D \\
Qwen3:8B & E & E & E & D & D & D & D & D \\
Qwen3:14B & F & F & F & E & D & D & D & D \\
Qwen3:32B & F & E & E & D & D & D & D & D \\
Qwen3:235B & F & F & F & D & D & D & D & D \\
\hline
DeepSeek-R1:1.5B & E & D & D & D & D & D & D & D \\
DeepSeek-R1:7B & D & D & D & D & D & D & D & D \\
DeepSeek-R1:14B & D & E & D & E & E & D & D & D \\
DeepSeek-R1:32B & D & E & E & E & D & D & D & D \\
DeepSeek-R1:70B & D & E & E & D,E & D & D & - & D \\
\hline
\end{tabular}
\end{table}

\subsubsection{Model Performance Analysis}

The performance analysis reveals a clear pattern of improvement with thinking budget. Without thinking, most models incorrectly identified either the midbrain downward vision center (E) or ventricular system damage (F) as the cause. This reflects a common misconception—assuming that downward gaze must result from damage to downward gaze centers, when in fact it results from damage to upward gaze centers causing unopposed downward deviation. With sufficient thinking budget (typically 256-512 tokens), models correctly identified damage to the midbrain upward vision center (D) as the cause. DeepSeek-R1 models showed superior performance, achieving correct diagnosis even without thinking budget in the case of the 7B model, suggesting better baseline understanding of neuroanatomy.

\subsection{Case Study 3: TIA Risk Stratification Using ABCD Score}

\textbf{Clinical Presentation:} A 68-year-old male presented with intermittent right lower limb shaking for 4 months. The shaking had a frequency of 3-4Hz, lasted 2-3 minutes per episode, occurred every 2-3 weeks, mostly when standing from a sitting position. The last attack was 2 days before admission. The shaking was accompanied by numbness and weakness of the right limbs, without disturbance of consciousness, tongue bite, or incontinence. Between attacks, activity was normal. Past medical history included hypertension for 10 years and diabetes for 3 years.

\textbf{Physical Examination:}
\begin{itemize}
\item Blood pressure: 135/85 mmHg
\item Clear consciousness, fluent speech
\item Limb muscle strength grade 5, normal muscle tone
\item Normal coordination and sensation
\item Limb tendon reflexes (++)
\item No pathological reflexes or meningeal signs
\item No vascular murmurs in neck
\end{itemize}

\textbf{Investigations:}
\begin{itemize}
\item Brain CT: No obvious abnormalities
\item Video EEG: No abnormalities during symptom attacks
\end{itemize}

\textbf{Question:} The ABCD score of this patient is:
\begin{itemize}
\item A. 2 points
\item B. 3 points
\item C. 4 points
\item D. 5 points
\item E. 6 points
\item F. 7 points
\end{itemize}

\begin{tcolorbox}[title=Model Thinking Process (Qwen3:235B with unlimited budget)]
\emph{The patient is a 68-year-old male with intermittent right lower limb shaking for 4 months. These episodes last 2-3 minutes, occur every 2-3 weeks, mostly when standing from sitting, and are accompanied by right limb numbness and weakness. He has a 10-year history of hypertension and 3-year history of diabetes. Physical exam shows BP 135/85 mmHg, normal neurological findings between attacks.}

\emph{The question asks for the ABCD score, which is used to predict stroke risk after TIA. Let me calculate each component:}

\emph{A - Age: The patient is 68 years old. In the ABCD score, age $\geq$60 years gets 1 point. So A = 1 point.}

\emph{B - Blood pressure: The patient's BP is 135/85 mmHg. For the ABCD score, BP $\geq$140/90 mmHg gets 1 point. Since 135/85 is less than 140/90, B = 0 points.}

\emph{C - Clinical features: The patient has unilateral weakness (right limbs) during attacks. In the ABCD score:
- Unilateral weakness = 2 points
- Speech disturbance without weakness = 1 point
- Other symptoms = 0 points
The patient has unilateral weakness, so C = 2 points. Note that speech is normal, so no additional points.}

\emph{D - Duration of symptoms: Each attack lasts about 2-3 minutes. In the ABCD score:
- $\geq$60 minutes = 2 points
- 10-59 minutes = 1 point
- $<$10 minutes = 0 points
Since 2-3 minutes is less than 10 minutes, D = 0 points.}

\emph{D - Diabetes: The patient has a 3-year history of diabetes. Presence of diabetes gets 1 point. So the second D = 1 point.}

\emph{Total ABCD score = A + B + C + D + D = 1 + 0 + 2 + 0 + 1 = 4 points.}

\emph{Looking at the options:
A. 2 points
B. 3 points
C. 4 points
D. 5 points
E. 6 points
F. 7 points}

\emph{The answer is C. 4 points.}
\end{tcolorbox}

\begin{table}[htbp]
\centering
\caption{Model responses for ABCD score calculation across thinking budgets}
\label{tab:case3_answers}
\begin{tabular}{lcccccccc}
\hline
Model & None & 64 & 128 & 256 & 512 & 1024 & inf & Correct \\
\hline
Qwen3:1.7B & A,B,C,D,E,F & B,C,D & A,B,C,D & B,C,D & C & C & C & C \\
Qwen3:4B & C & B & B & B,C & C & C & C & C \\
Qwen3:8B & B,C,D,E,F & B & B & C & C & C & C & C \\
Qwen3:14B & B,C,D,E,F & B,C,D,E,F & B,C,D,E,F & E & C & C & C & C \\
Qwen3:32B & B,C,D,E,F & D & B,F & D & C & C & C & C \\
Qwen3:235B & E & E & C & C & C & C & C & C \\
\hline
DeepSeek-R1:1.5B & B & B & B & C & C & C & C & C \\
DeepSeek-R1:7B & B & C & C & C & C & C & C & C \\
DeepSeek-R1:14B & B & B & B & B & B & B,C & C & C \\
DeepSeek-R1:32B & B & B & C & C & C & C & C & C \\
DeepSeek-R1:70B & B & B & B & B & C & C & - & C \\
\hline
\end{tabular}
\end{table}

\subsubsection{Model Performance Analysis}

This case demonstrates how thinking budget dramatically improves clinical risk assessment accuracy. The ABCD score calculation requires careful evaluation of multiple clinical parameters: age, blood pressure, clinical features, symptom duration, and diabetes status. Without thinking budget, most models struggled with accurate scoring—Qwen3:235B incorrectly calculated 6 points (E), while smaller models often selected multiple scores simultaneously, indicating uncertainty. The improvement with thinking budget is striking: Qwen3:235B achieved correct scoring with just 128 tokens, while smaller models required 512 tokens or more. DeepSeek-R1:7B showed exceptional performance, correctly calculating the score even with minimal thinking budget, suggesting superior baseline understanding of clinical scoring systems. The case highlights how structured thinking enables models to systematically evaluate each ABCD component, avoiding common errors like misreading blood pressure values or incorrectly weighting clinical features. This systematic approach to risk stratification is crucial for appropriate triage and treatment decisions in stroke prevention.

\subsection{Implications for Medical AI Systems}

These three cases collectively demonstrate several critical insights for medical AI deployment. First, the complexity of medical reasoning extends far beyond simple pattern matching—successful diagnosis and treatment planning require temporal analysis, comprehensive differential consideration, and nuanced understanding of therapeutic options. Second, thinking budget provides a controllable mechanism for ensuring thorough analysis in complex cases while maintaining efficiency for routine diagnoses. Third, the transparent reasoning processes enable clinicians to verify not just the final answer but the logic path taken, building trust through interpretability. Understanding these patterns of improvement—from incomplete differentials to comprehensive assessments, from missed treatments to full therapeutic planning—can guide both system design and clinical integration strategies. Most importantly, these examples show that with appropriate thinking budget allocation, AI systems can achieve expert-level performance on challenging medical cases while maintaining the transparency required for clinical acceptance.

\newpage
\section{Token Distribution Analysis}

\subsection{Question Token Distribution}

\begin{figure}[H]
\centering
\includegraphics[width=0.9\textwidth]{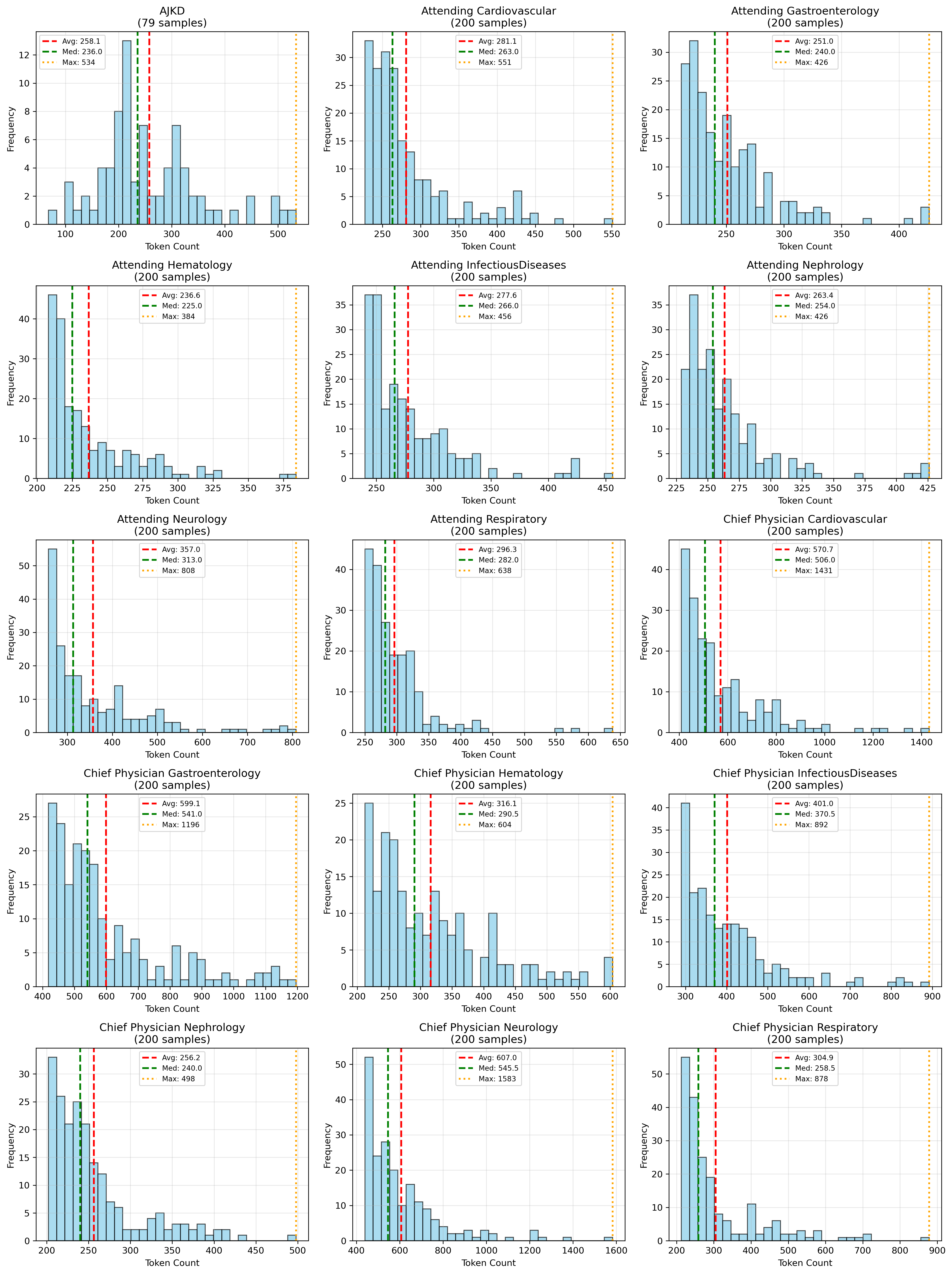}
\caption{Question token length distributions across all 15 medical datasets. Each subplot shows the frequency distribution of question lengths for a specific dataset, with vertical lines indicating mean (red), median (green), and maximum (orange) values. The distributions reveal significant variation in question complexity across specialties, with chief physician level questions (bottom two rows) generally showing longer and more variable lengths compared to attending physician level questions (top two rows). Neurology datasets consistently show the highest complexity with the widest distributions.}
\label{fig:question_token_dist_full}
\end{figure}

\newpage
\subsection{Thinking Token Distribution}

\begin{figure}[H]
\centering
\includegraphics[width=0.9\textwidth]{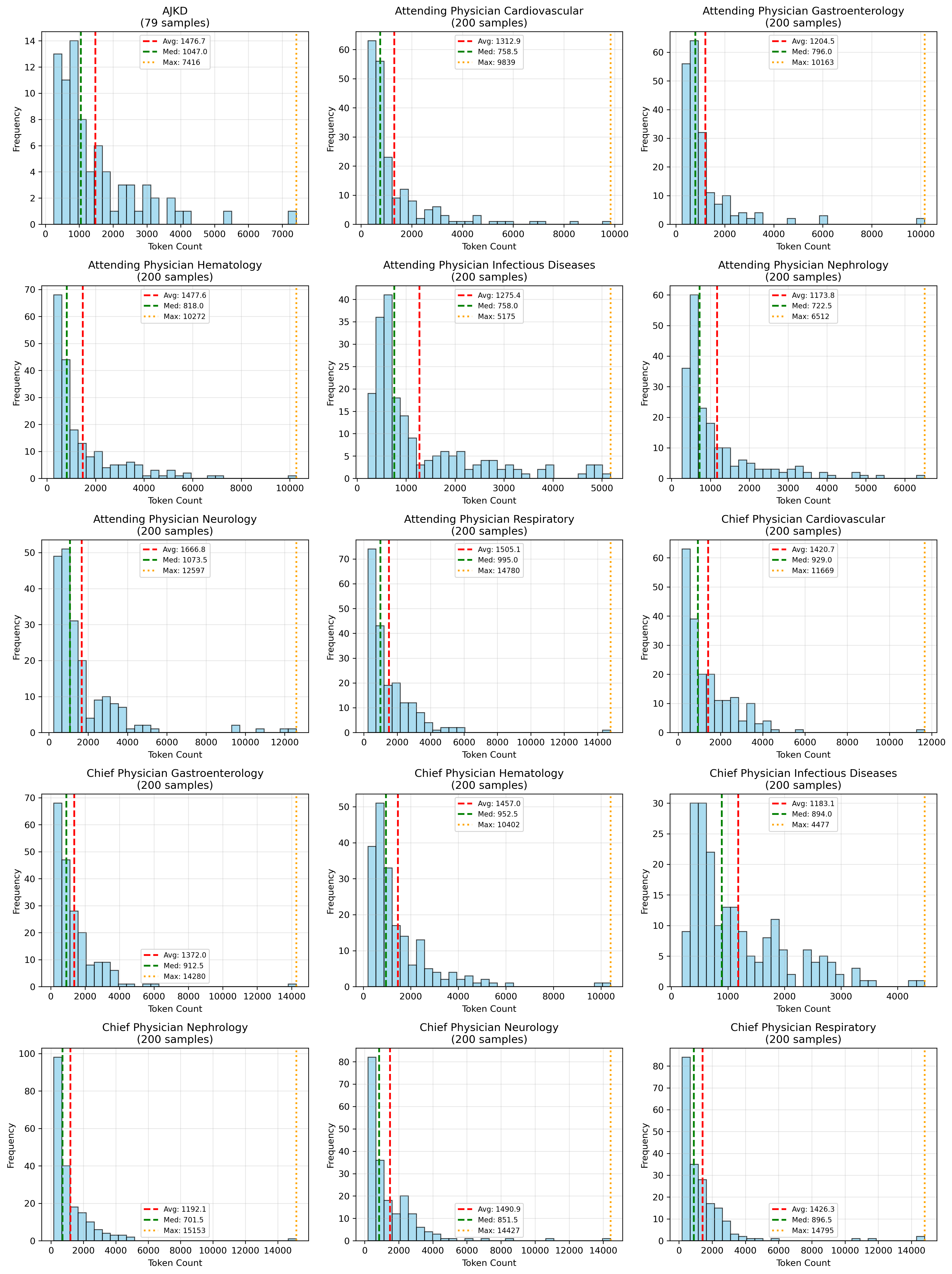}
\caption{Thinking token length distributions for Qwen3:235B model with unlimited thinking budget across all 15 medical datasets. Each subplot displays the frequency distribution of thinking process lengths, with vertical lines marking mean (red), median (green), and maximum (orange) values. The distributions exhibit pronounced right skew, indicating that while most questions require moderate thinking depths (1000-2000 tokens), a subset of complex cases trigger extensive reasoning processes exceeding 10,000 tokens. The variation in thinking depth both within and across datasets underscores the importance of adaptive thinking budget mechanisms.}
\label{fig:thinking_token_dist_full}
\end{figure}

\newpage
\section{Complete Results: Model Performance Across All Datasets}
\label{sec:complete_results}

This section presents comprehensive performance results organized by medical specialty, facilitating direct comparison between difficulty levels, model families, and thinking budget conditions.

\subsection{Cardiovascular}

\begin{figure}[H]
\centering
\begin{subfigure}[b]{0.24\textwidth}
\centering
\includegraphics[width=\textwidth]{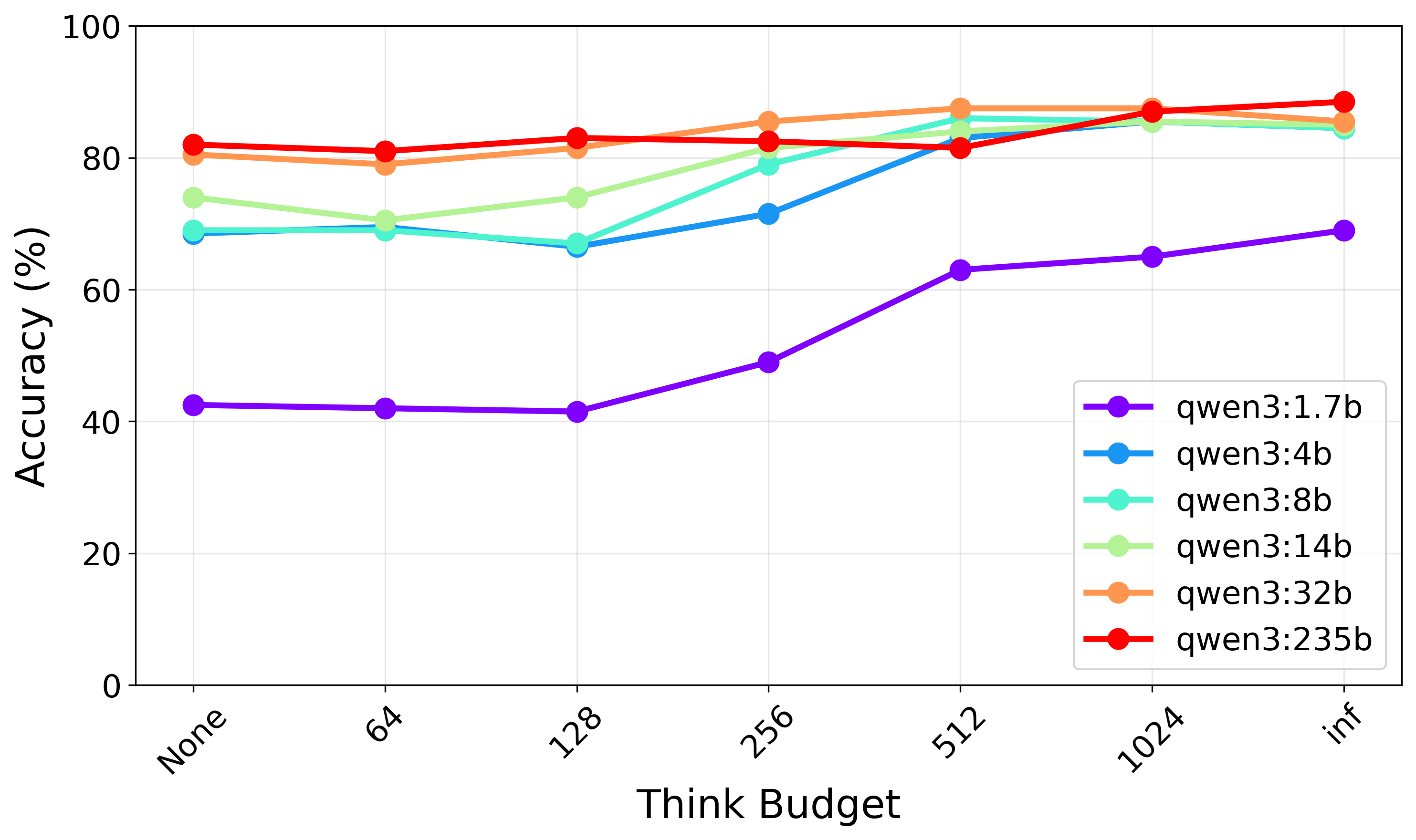}
\caption{Attending (Qwen3)}
\end{subfigure}
\hfill
\begin{subfigure}[b]{0.24\textwidth}
\centering
\includegraphics[width=\textwidth]{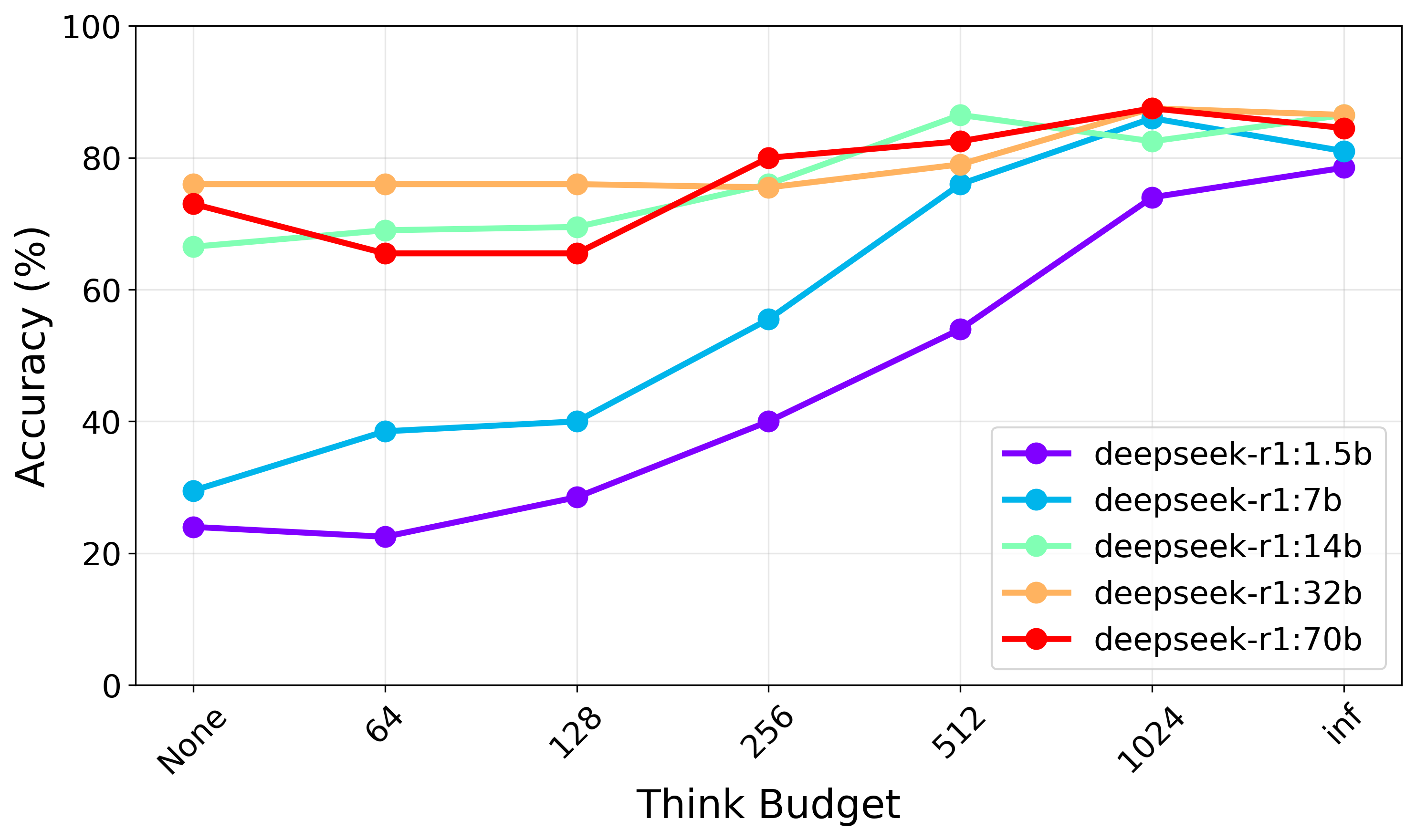}
\caption{Attending (DeepSeek-R1)}
\end{subfigure}
\hfill
\begin{subfigure}[b]{0.24\textwidth}
\centering
\includegraphics[width=\textwidth]{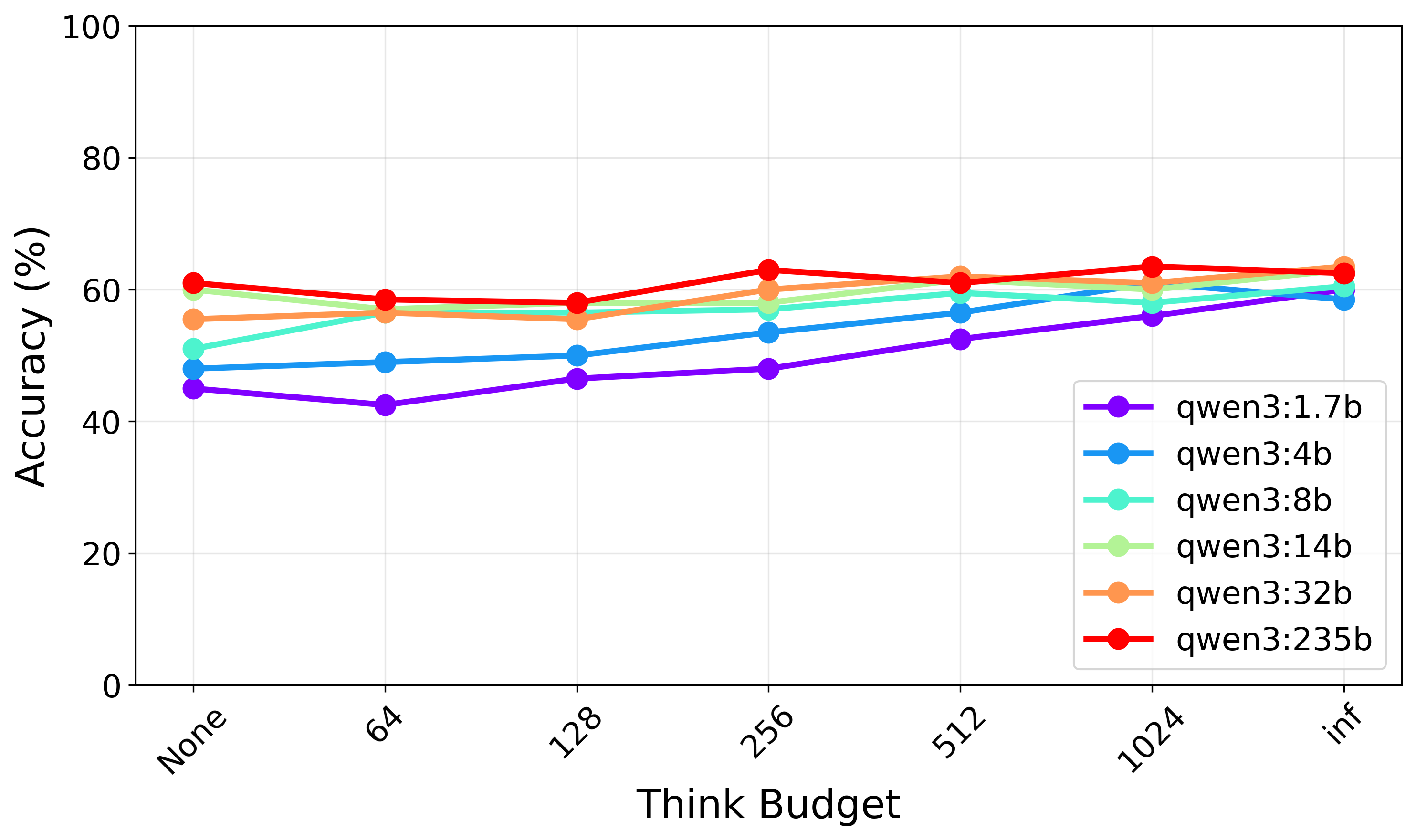}
\caption{Chief (Qwen3)}
\end{subfigure}
\hfill
\begin{subfigure}[b]{0.24\textwidth}
\centering
\includegraphics[width=\textwidth]{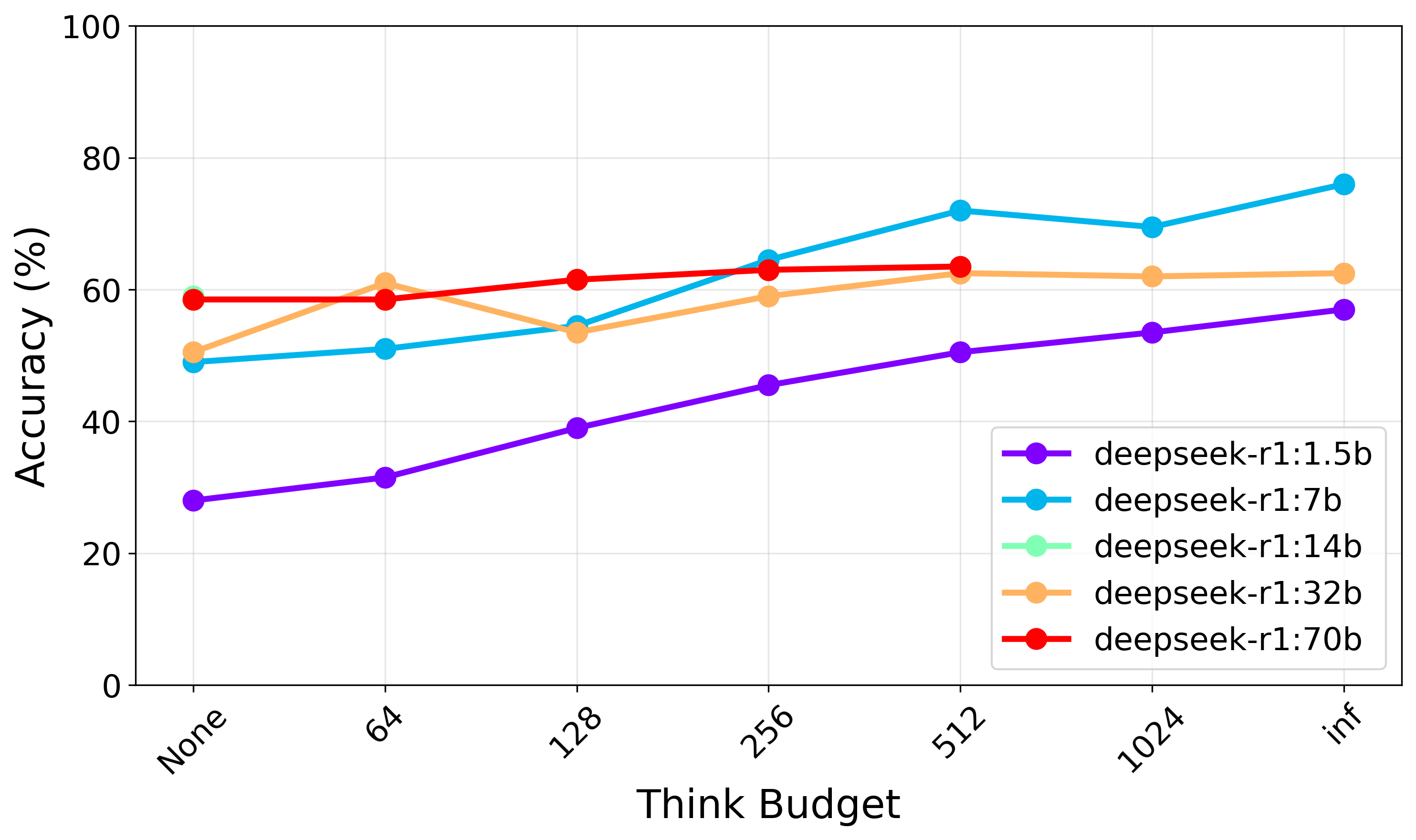}
\caption{Chief (DeepSeek-R1)}
\end{subfigure}

\vspace{0.5em}

\begin{subfigure}[b]{0.24\textwidth}
\centering
\includegraphics[width=\textwidth]{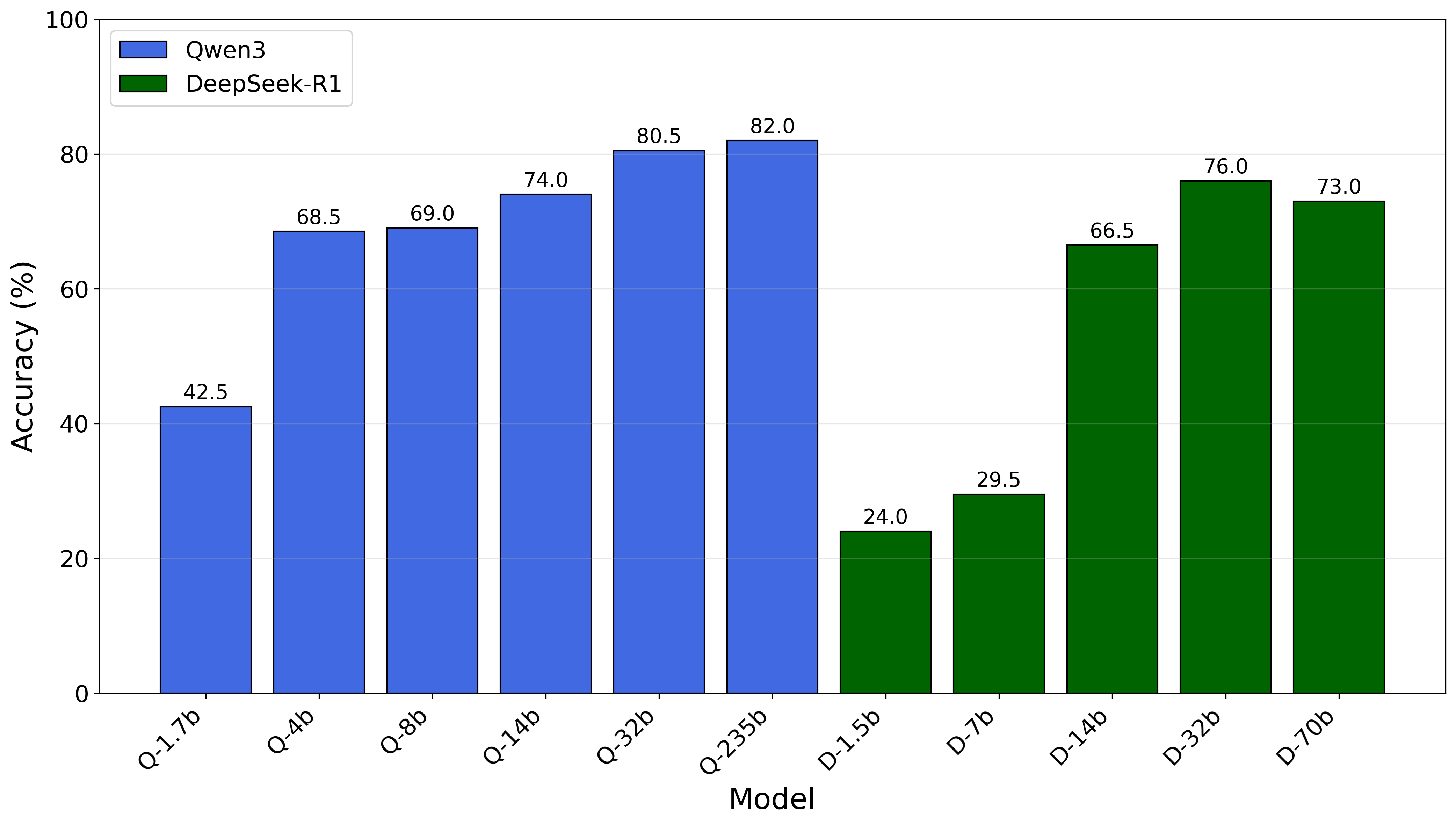}
\caption{Attending (None)}
\end{subfigure}
\hfill
\begin{subfigure}[b]{0.24\textwidth}
\centering
\includegraphics[width=\textwidth]{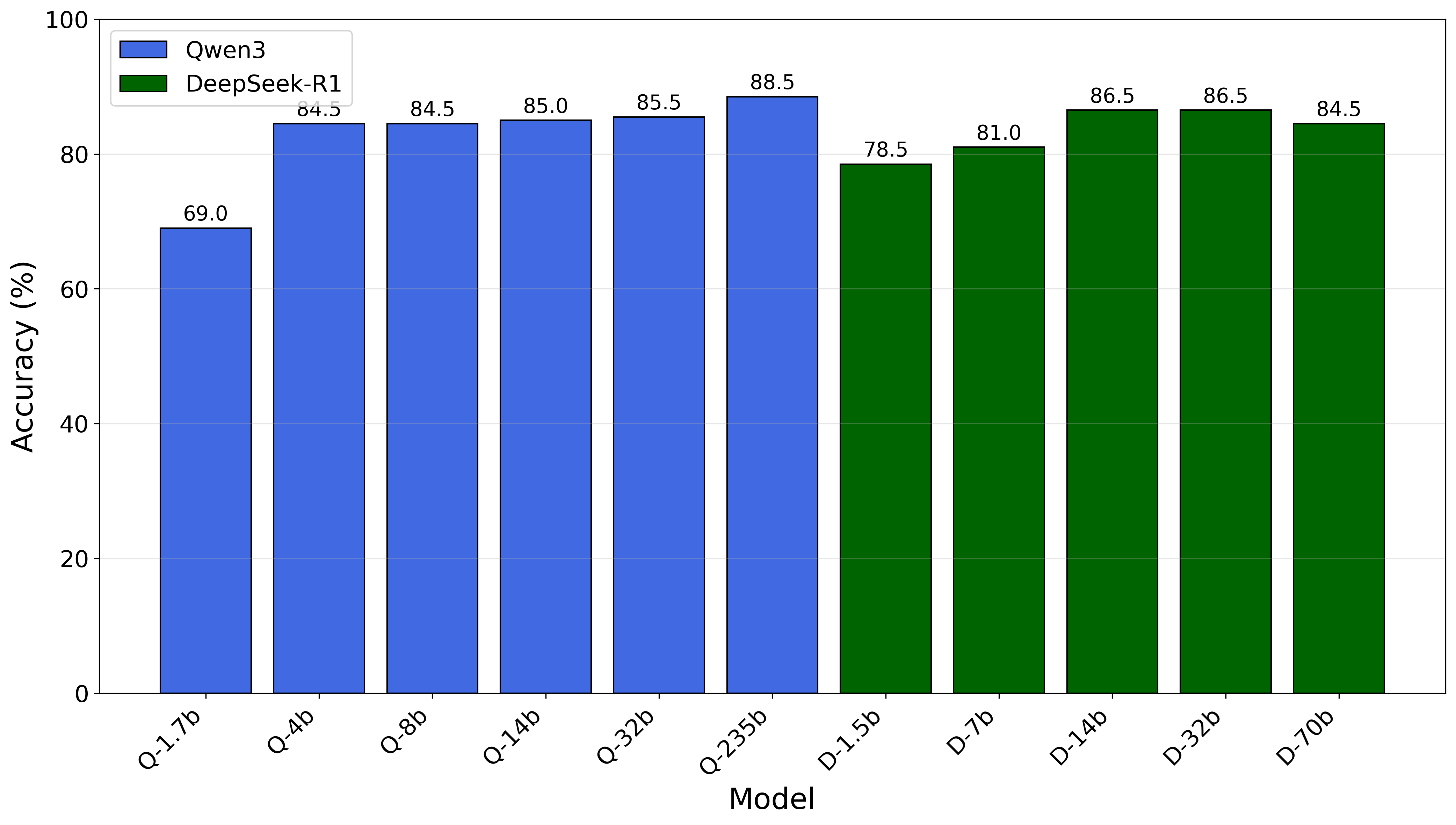}
\caption{Attending (inf)}
\end{subfigure}
\hfill
\begin{subfigure}[b]{0.24\textwidth}
\centering
\includegraphics[width=\textwidth]{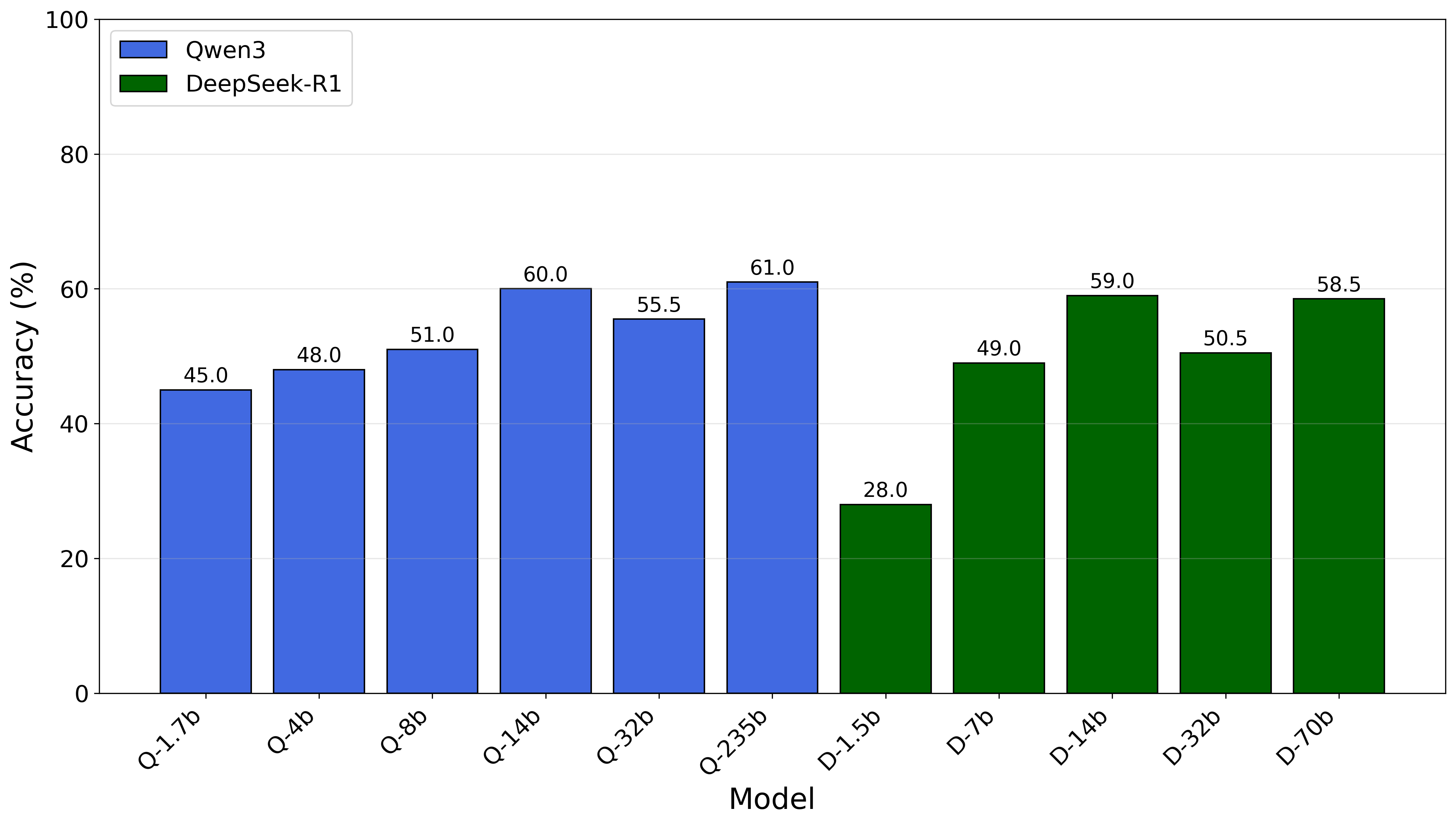}
\caption{Chief (None)}
\end{subfigure}
\hfill
\begin{subfigure}[b]{0.24\textwidth}
\centering
\includegraphics[width=\textwidth]{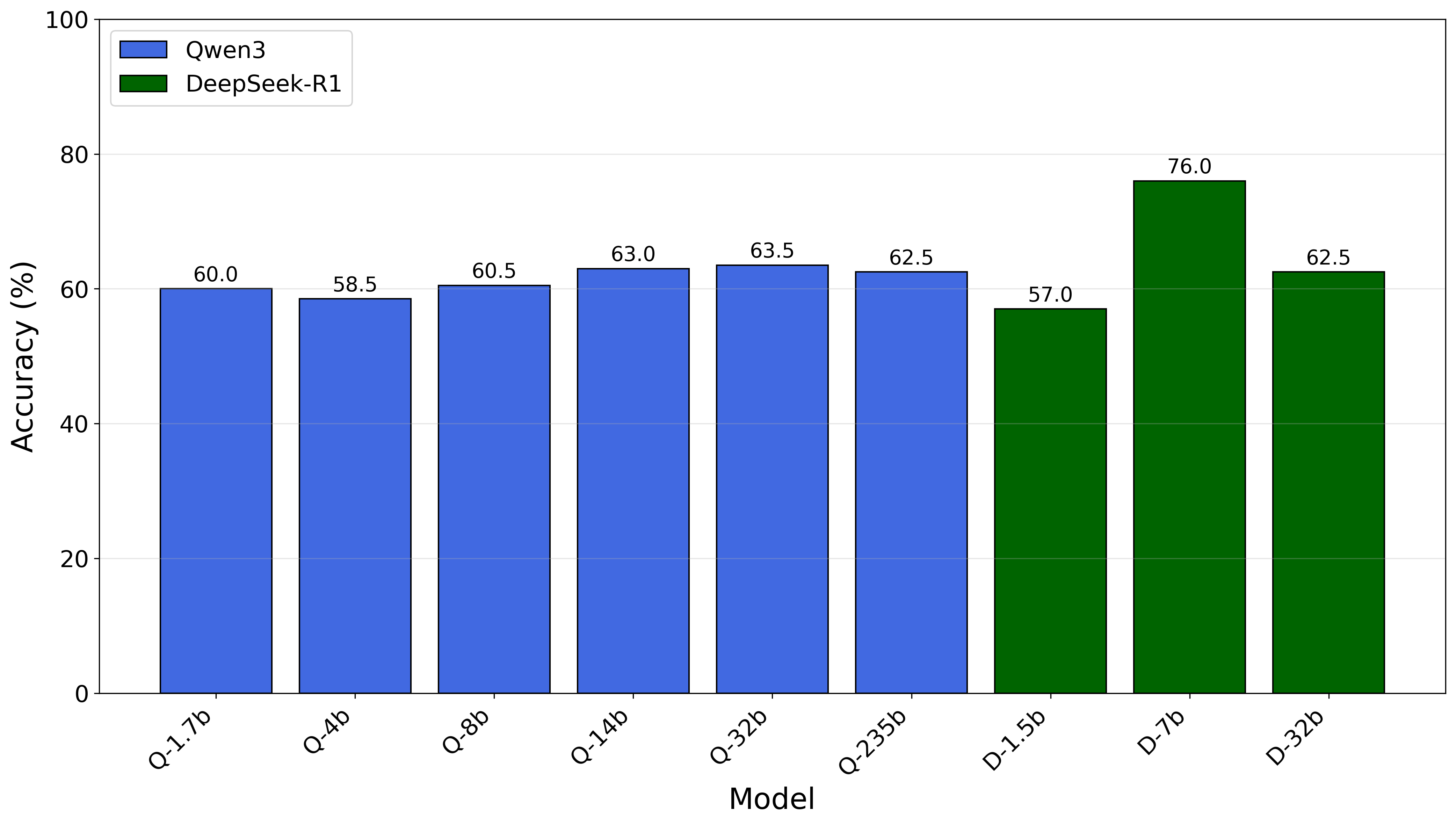}
\caption{Chief (inf)}
\end{subfigure}
\caption{Cardiovascular medicine performance across thinking budgets}
\label{fig:cardiovascular_complete}
\end{figure}

\subsection{Gastroenterology}

\begin{figure}[H]
\centering
\begin{subfigure}[b]{0.24\textwidth}
\centering
\includegraphics[width=\textwidth]{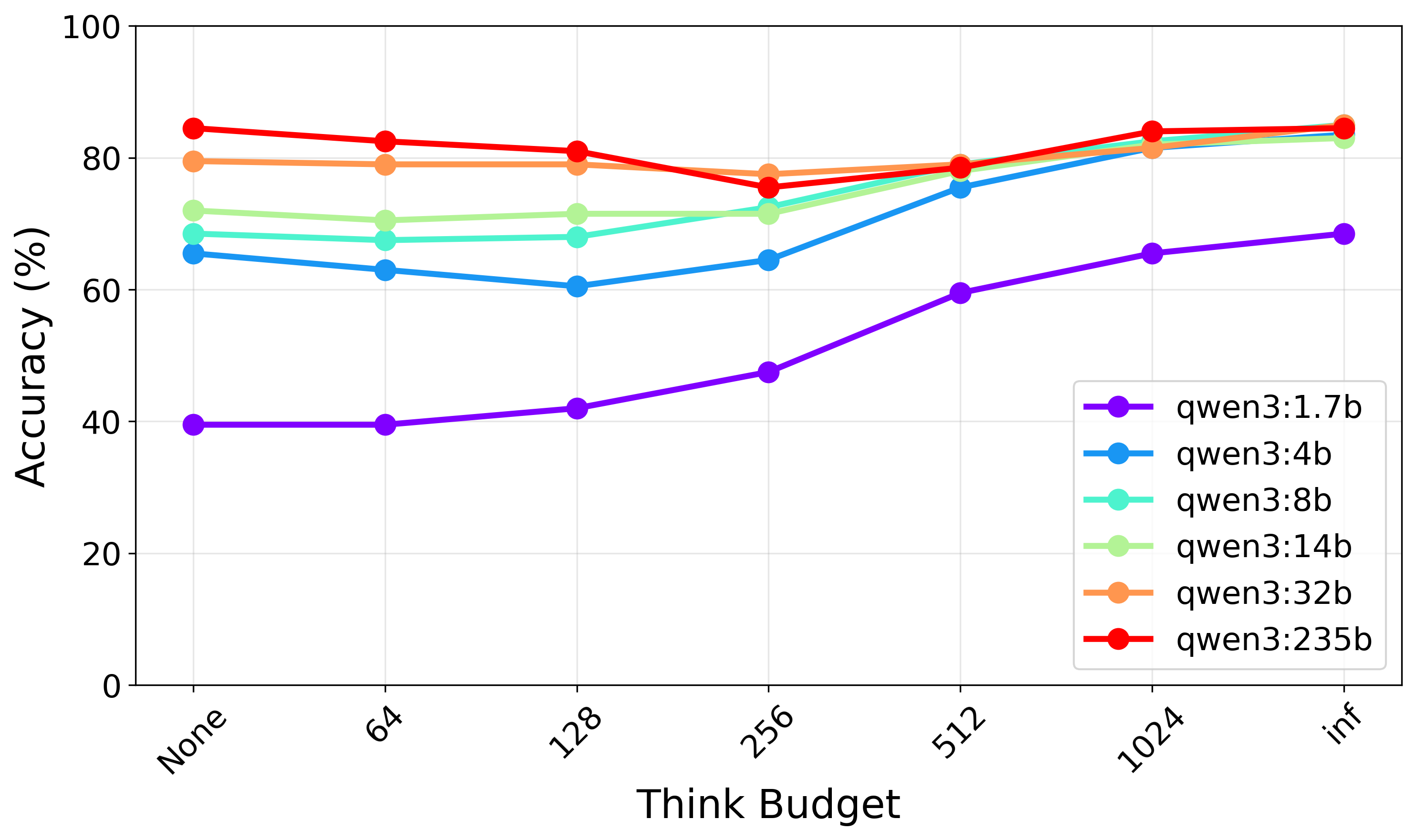}
\caption{Attending (Qwen3)}
\end{subfigure}
\hfill
\begin{subfigure}[b]{0.24\textwidth}
\centering
\includegraphics[width=\textwidth]{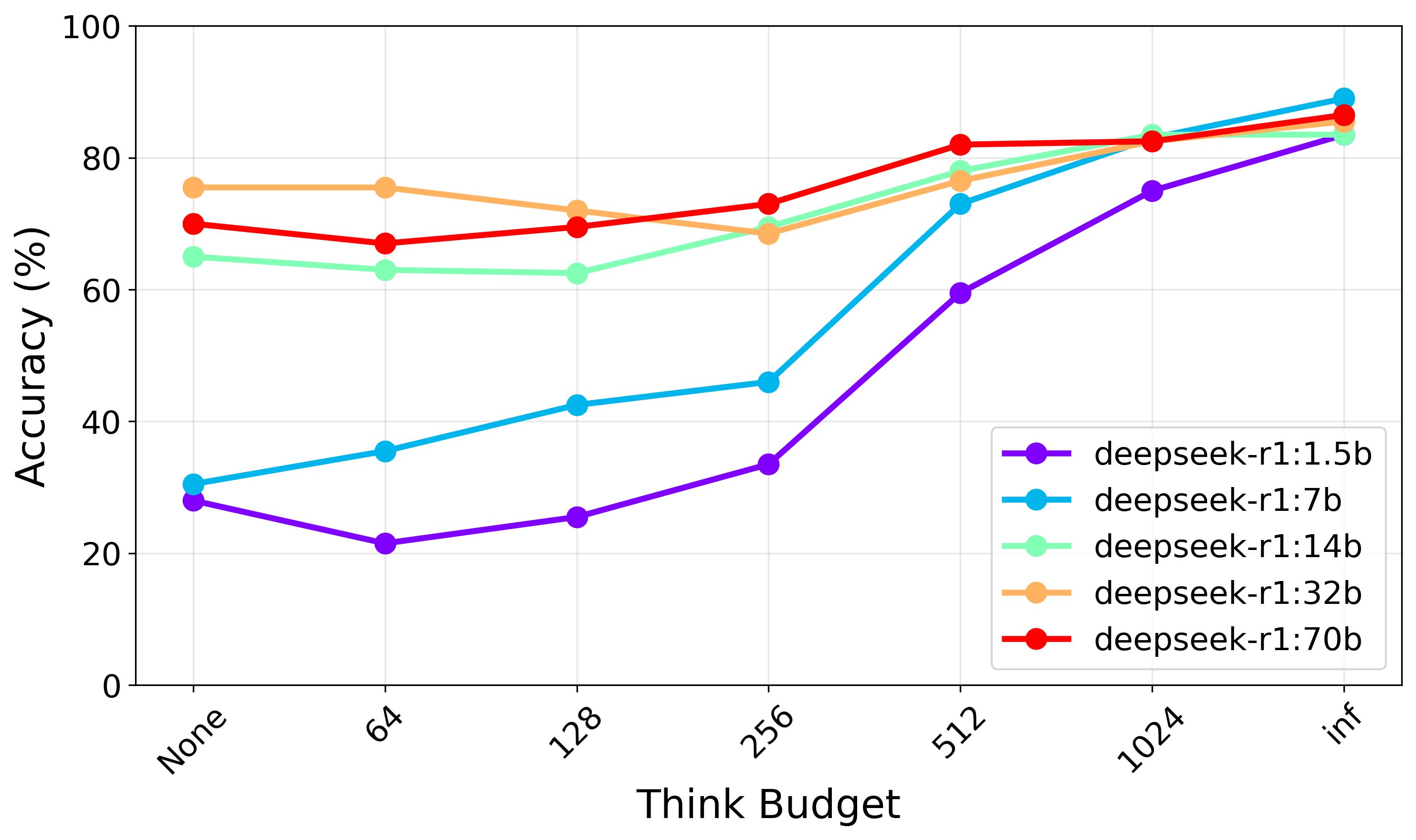}
\caption{Attending (DeepSeek-R1)}
\end{subfigure}
\hfill
\begin{subfigure}[b]{0.24\textwidth}
\centering
\includegraphics[width=\textwidth]{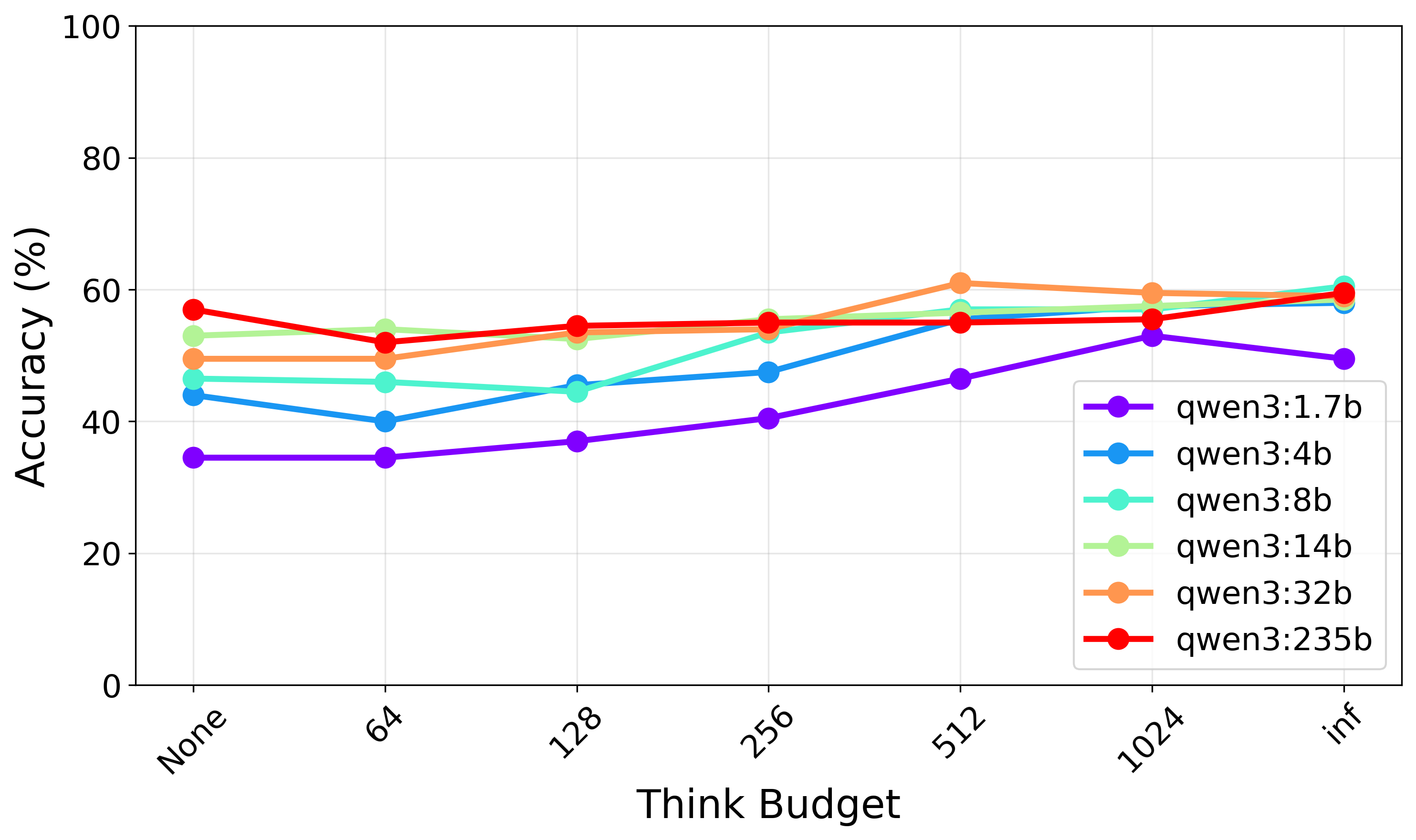}
\caption{Chief (Qwen3)}
\end{subfigure}
\hfill
\begin{subfigure}[b]{0.24\textwidth}
\centering
\includegraphics[width=\textwidth]{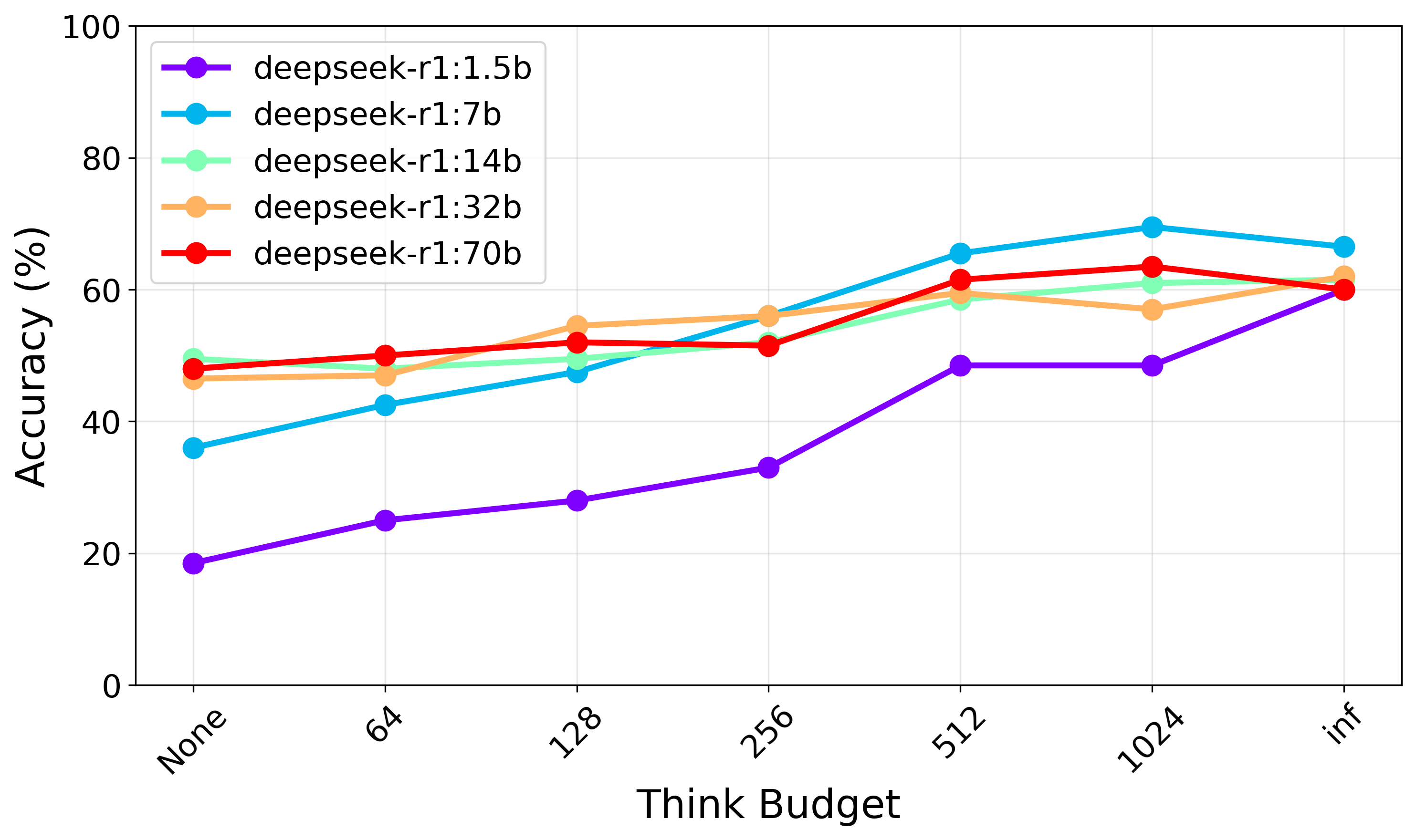}
\caption{Chief (DeepSeek-R1)}
\end{subfigure}

\vspace{0.5em}

\begin{subfigure}[b]{0.24\textwidth}
\centering
\includegraphics[width=\textwidth]{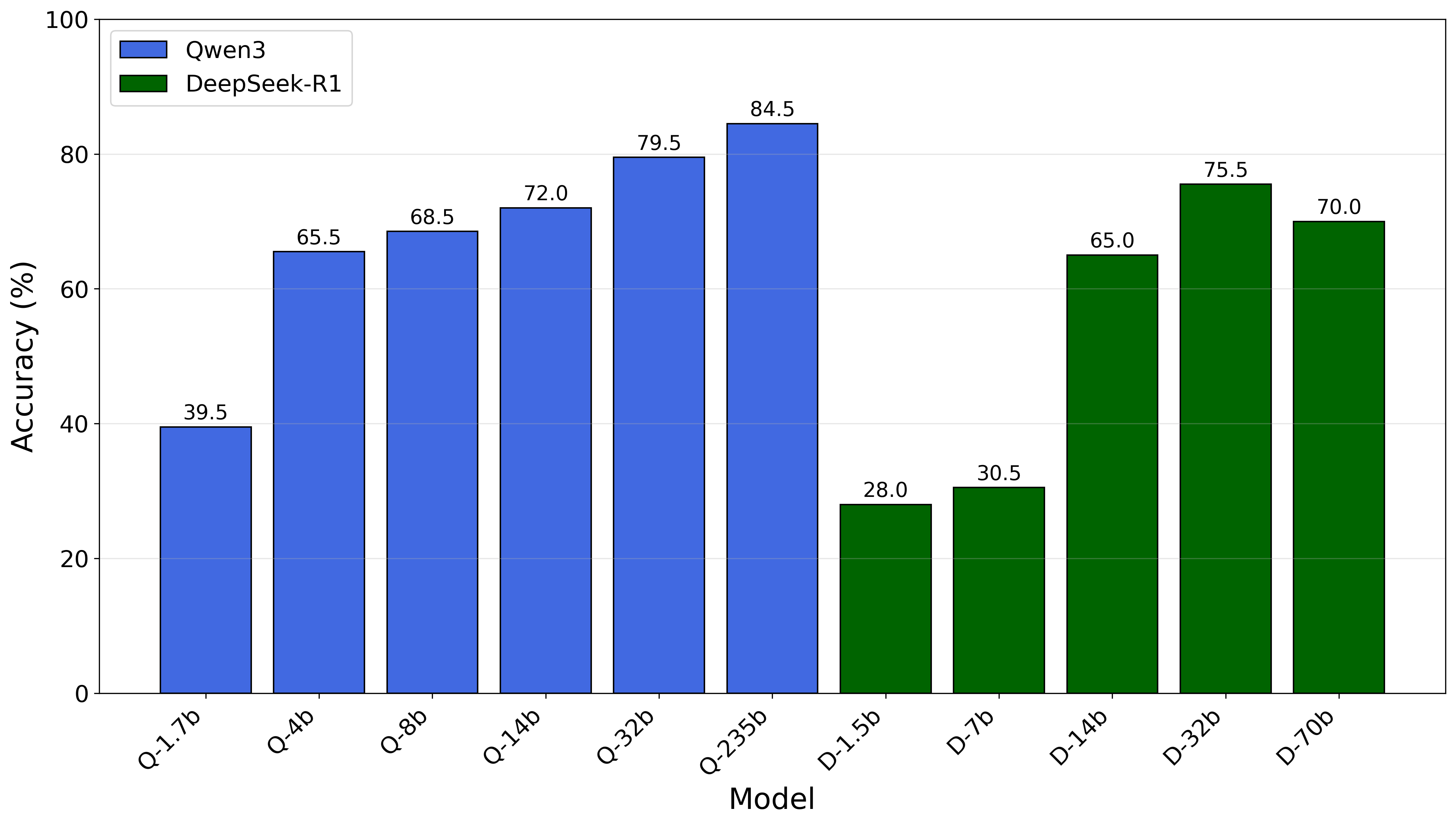}
\caption{Attending (None)}
\end{subfigure}
\hfill
\begin{subfigure}[b]{0.24\textwidth}
\centering
\includegraphics[width=\textwidth]{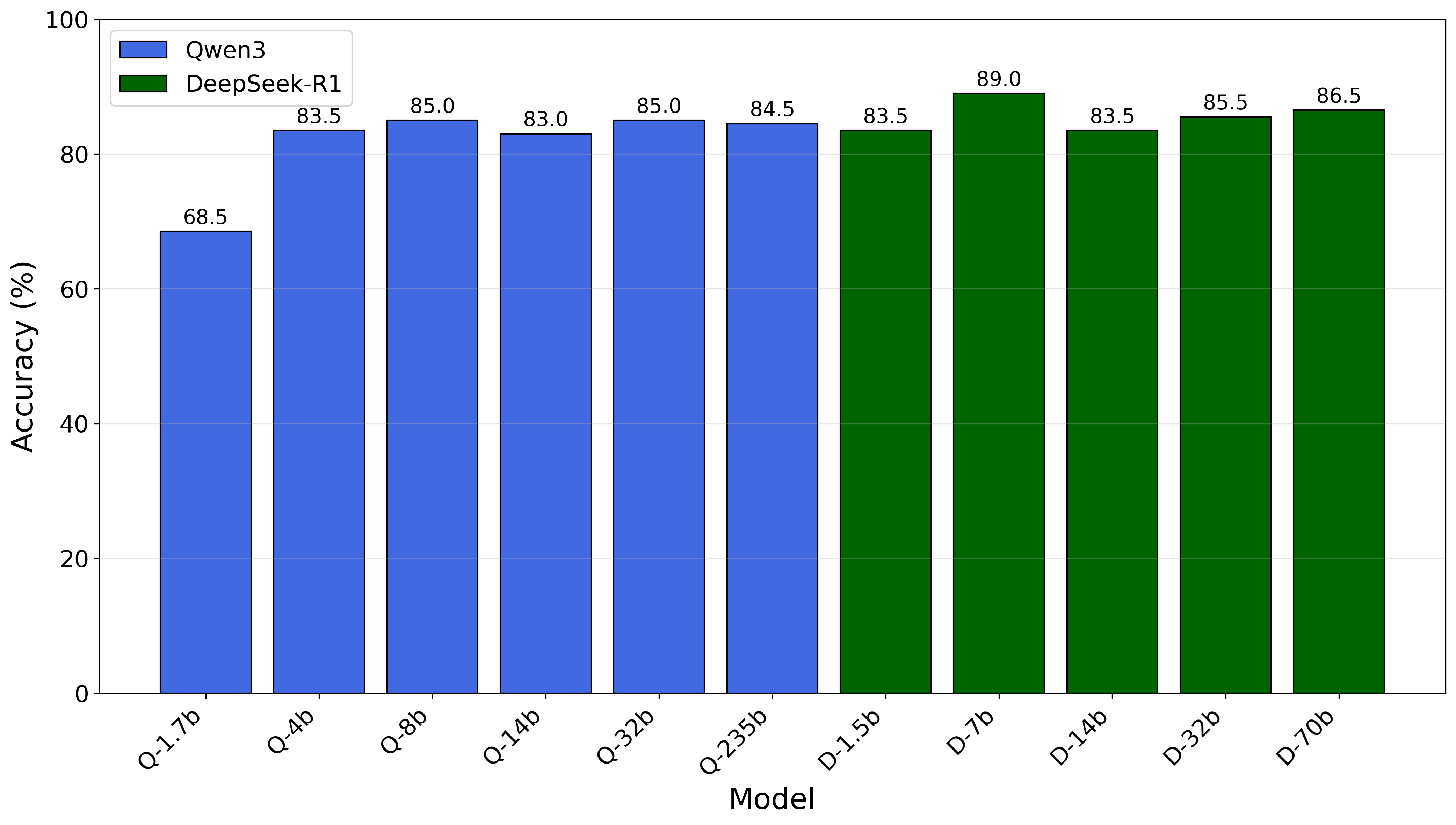}
\caption{Attending (inf)}
\end{subfigure}
\hfill
\begin{subfigure}[b]{0.24\textwidth}
\centering
\includegraphics[width=\textwidth]{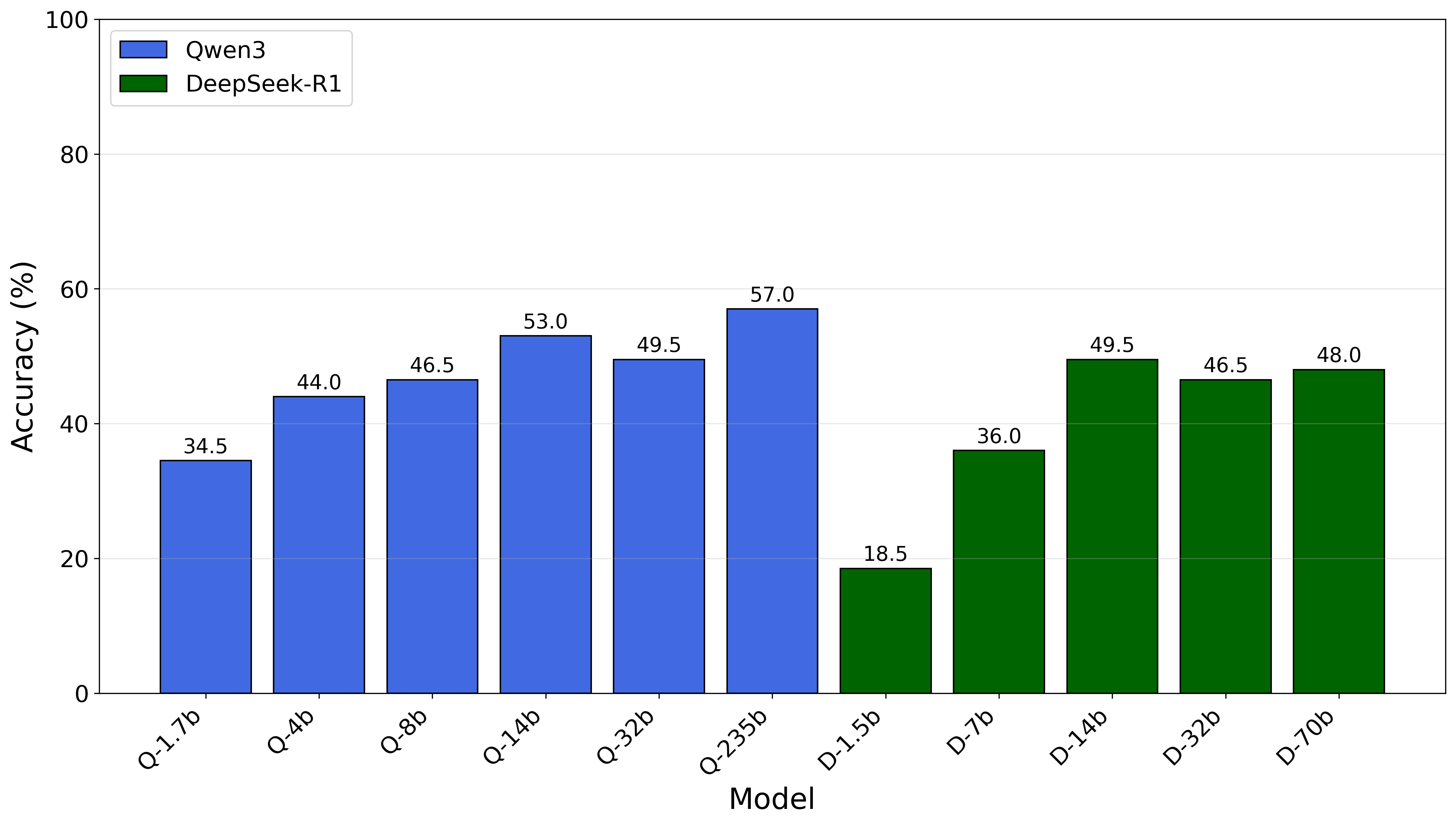}
\caption{Chief (None)}
\end{subfigure}
\hfill
\begin{subfigure}[b]{0.24\textwidth}
\centering
\includegraphics[width=\textwidth]{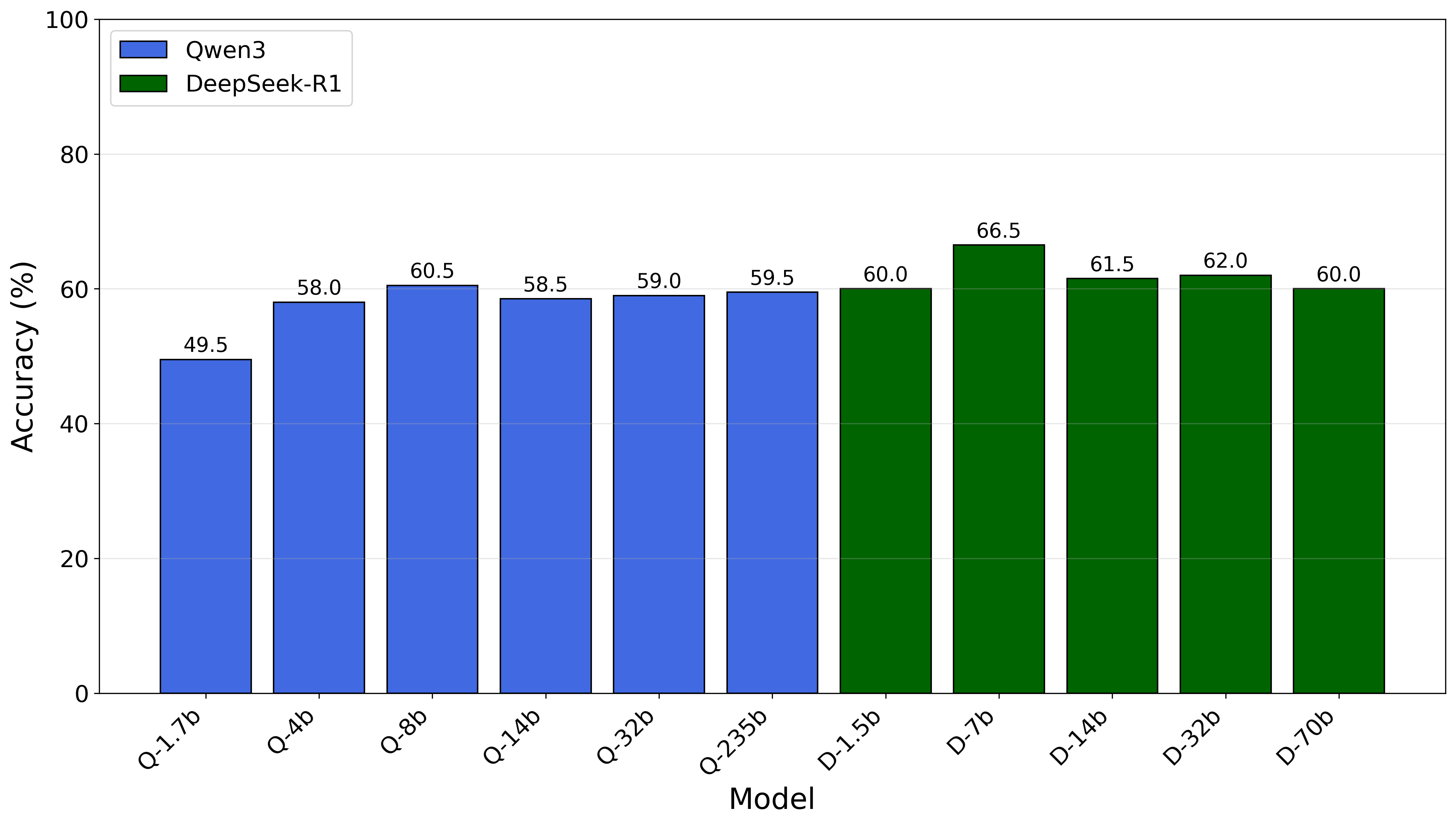}
\caption{Chief (inf)}
\end{subfigure}
\caption{Gastroenterology performance across thinking budgets}
\label{fig:gastroenterology_complete}
\end{figure}

\subsection{Hematology}

\begin{figure}[H]
\centering
\begin{subfigure}[b]{0.24\textwidth}
\centering
\includegraphics[width=\textwidth]{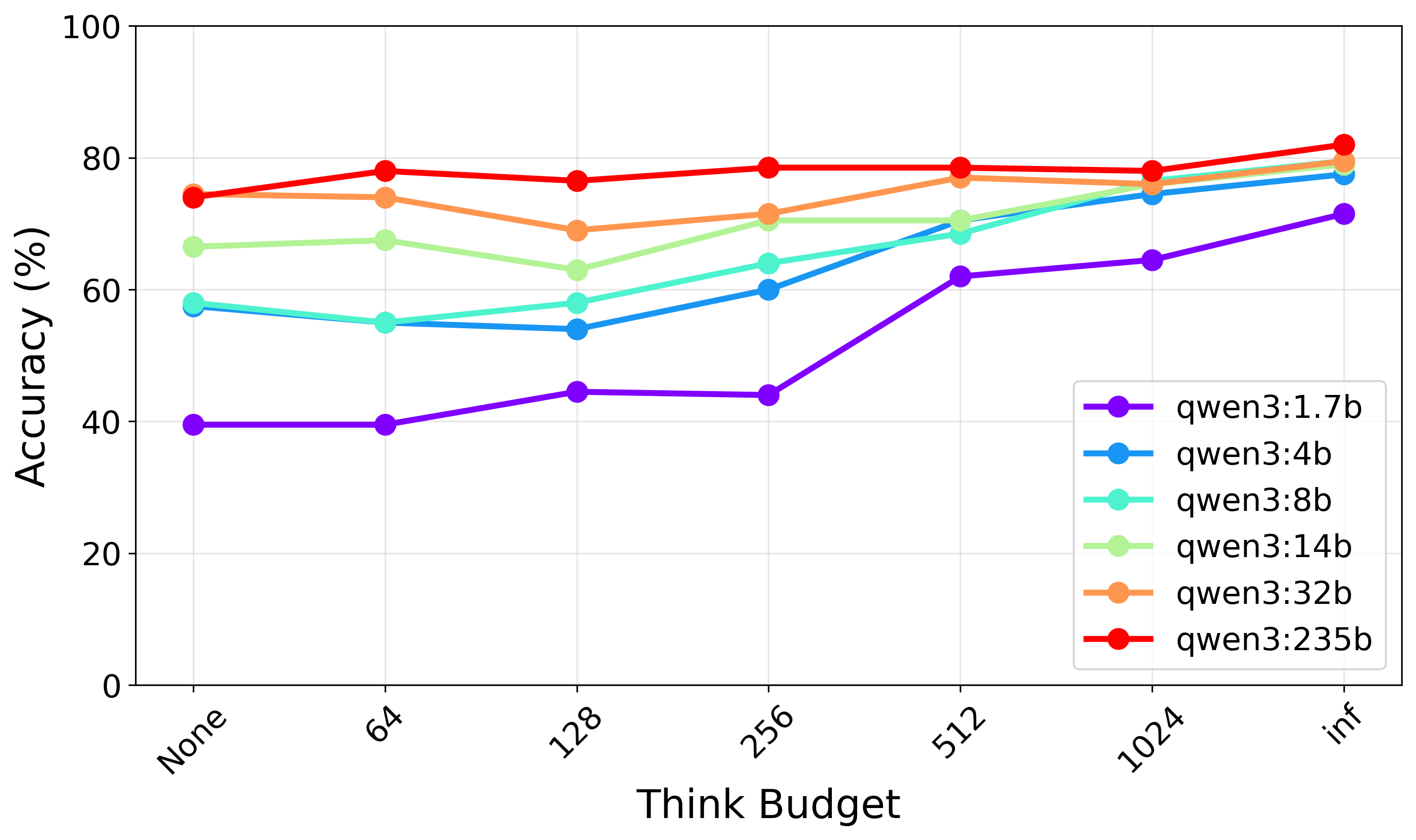}
\caption{Attending (Qwen3)}
\end{subfigure}
\hfill
\begin{subfigure}[b]{0.24\textwidth}
\centering
\includegraphics[width=\textwidth]{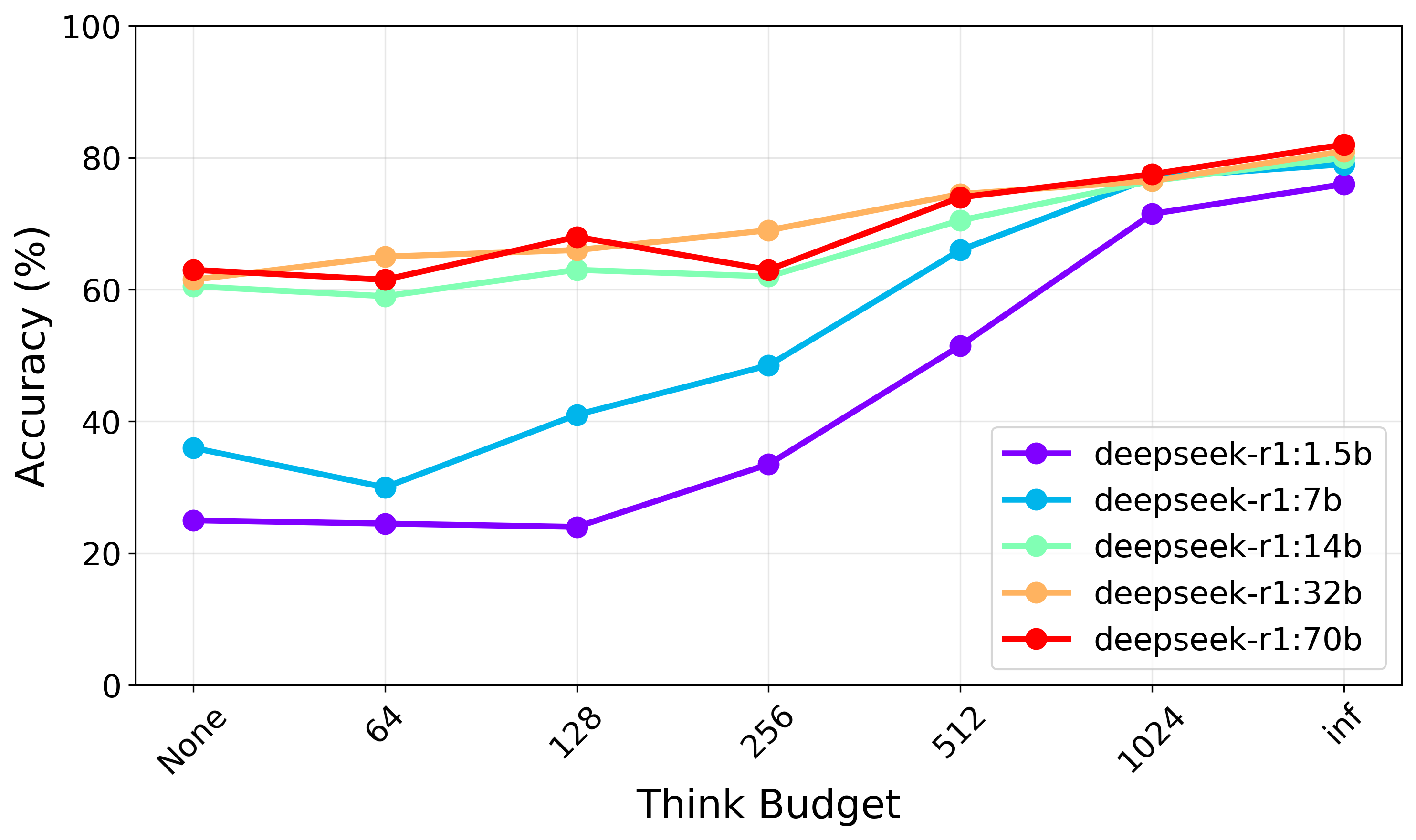}
\caption{Attending (DeepSeek-R1)}
\end{subfigure}
\hfill
\begin{subfigure}[b]{0.24\textwidth}
\centering
\includegraphics[width=\textwidth]{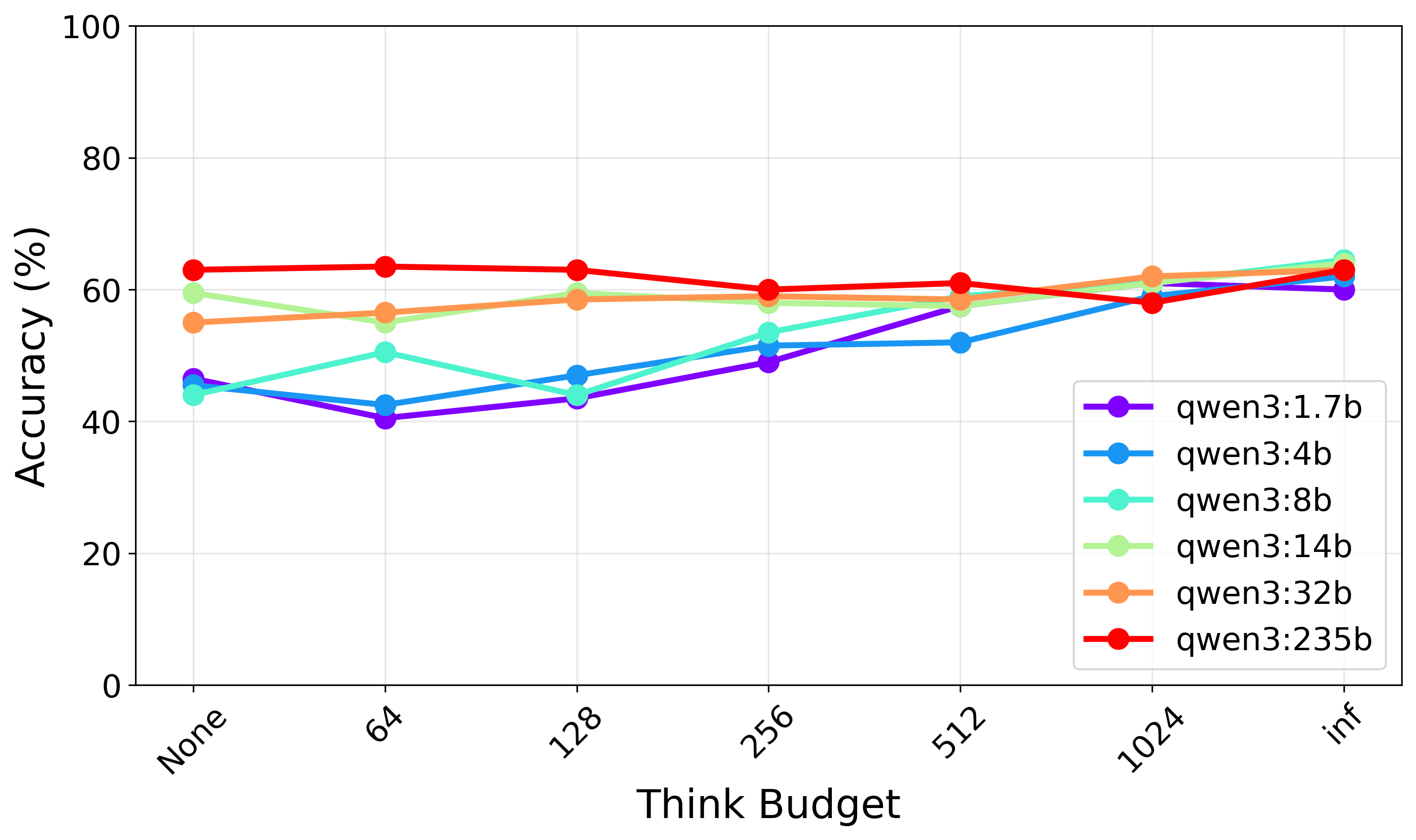}
\caption{Chief (Qwen3)}
\end{subfigure}
\hfill
\begin{subfigure}[b]{0.24\textwidth}
\centering
\includegraphics[width=\textwidth]{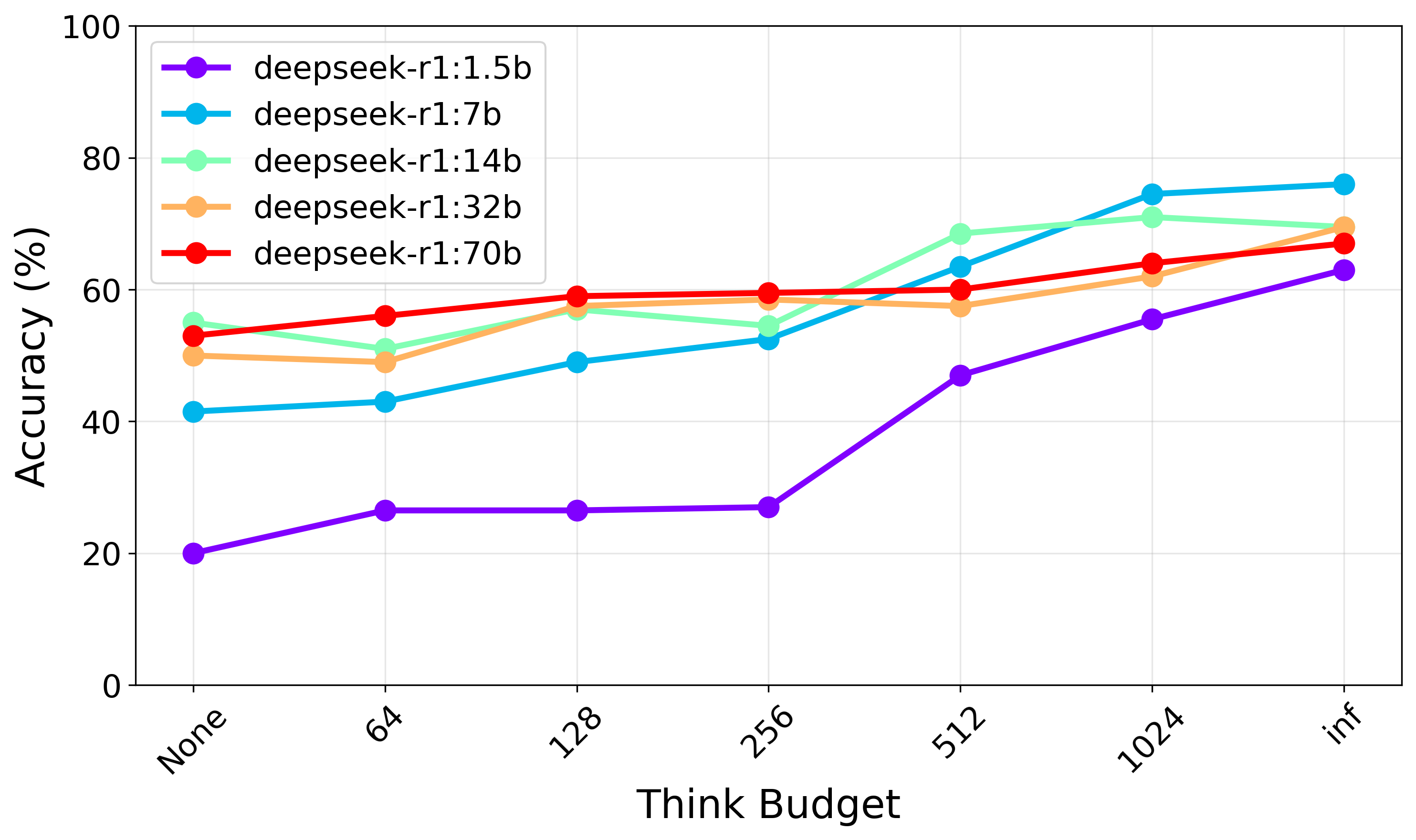}
\caption{Chief (DeepSeek-R1)}
\end{subfigure}

\vspace{0.5em}

\begin{subfigure}[b]{0.24\textwidth}
\centering
\includegraphics[width=\textwidth]{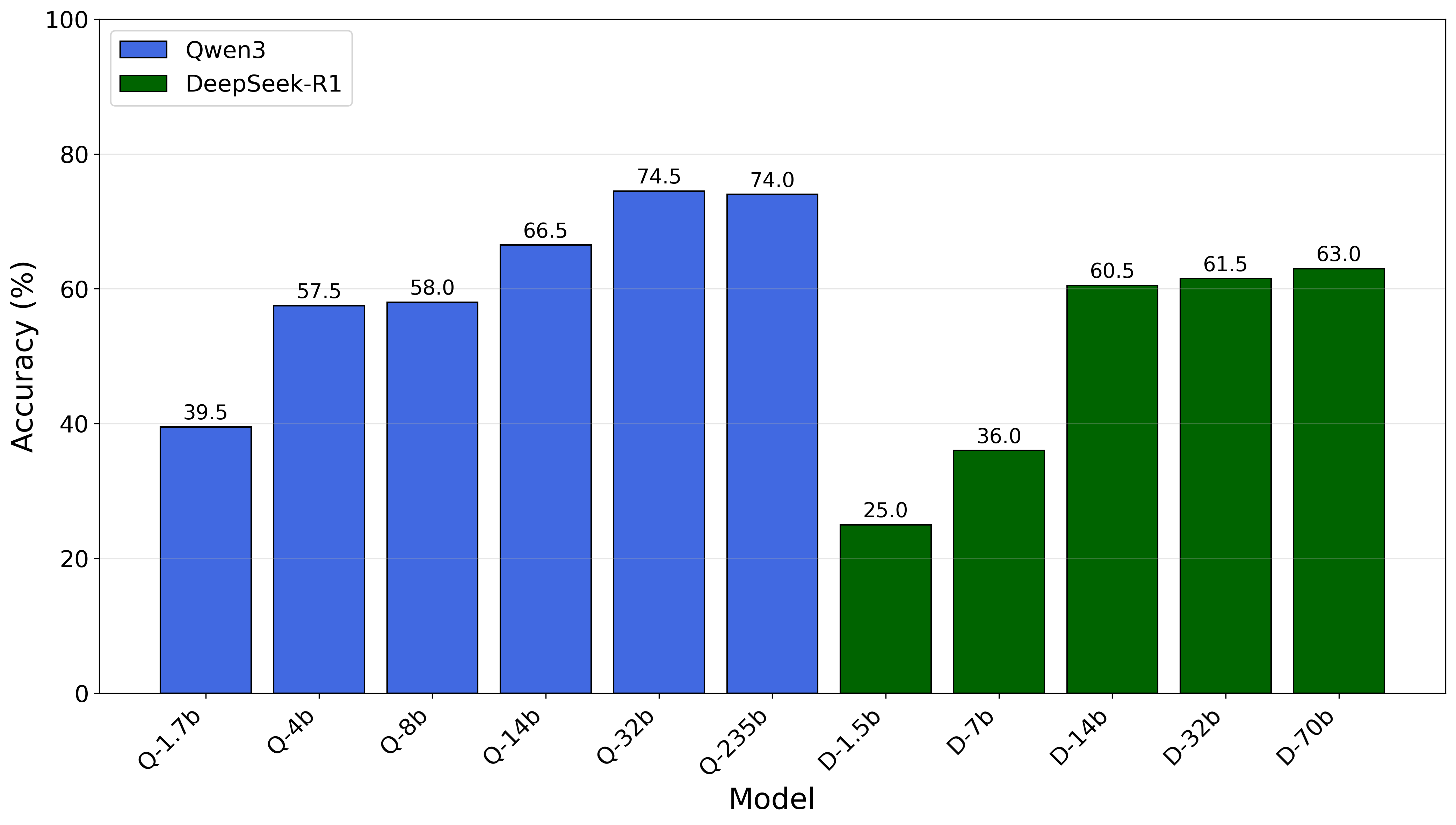}
\caption{Attending (None)}
\end{subfigure}
\hfill
\begin{subfigure}[b]{0.24\textwidth}
\centering
\includegraphics[width=\textwidth]{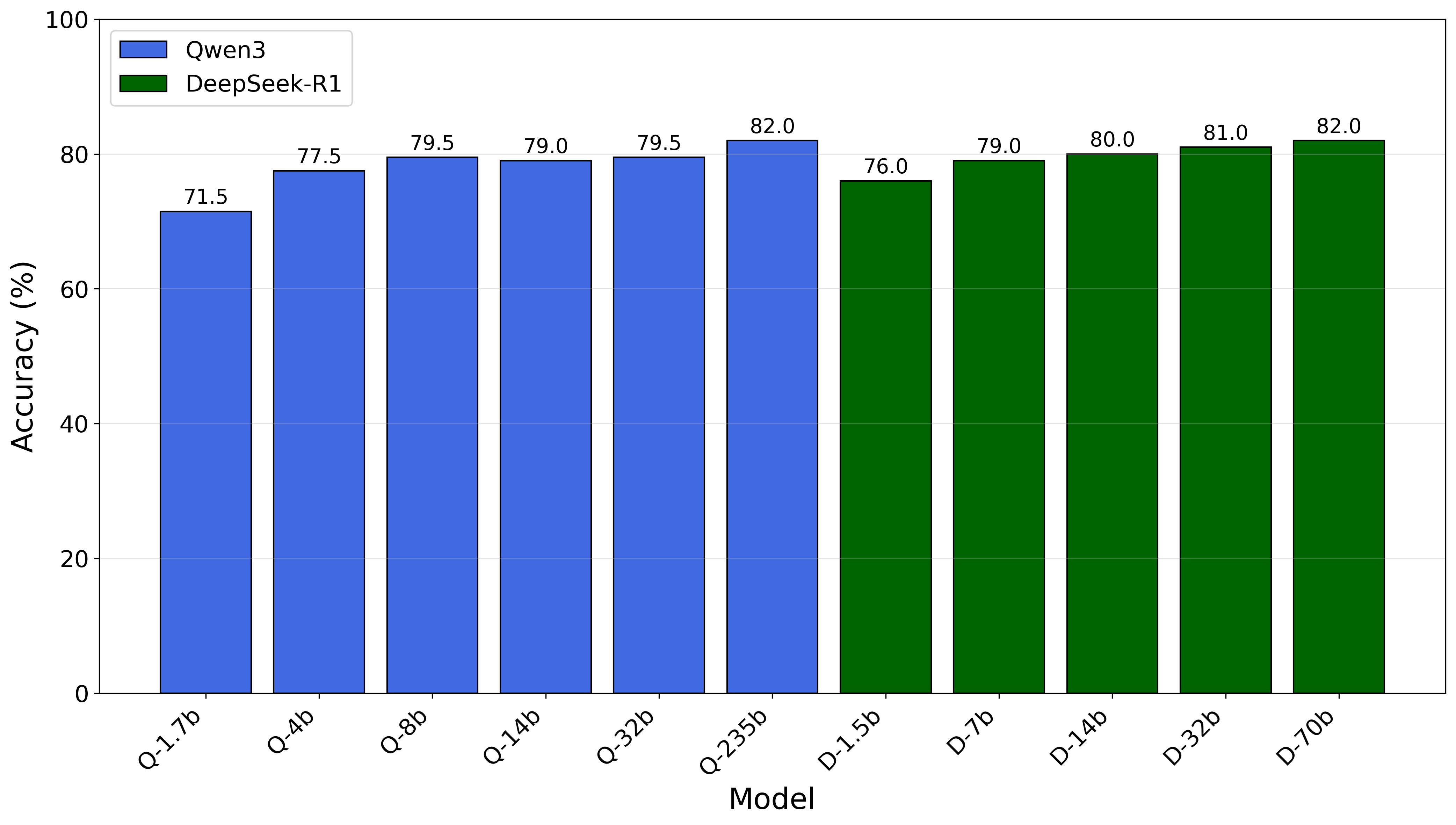}
\caption{Attending (inf)}
\end{subfigure}
\hfill
\begin{subfigure}[b]{0.24\textwidth}
\centering
\includegraphics[width=\textwidth]{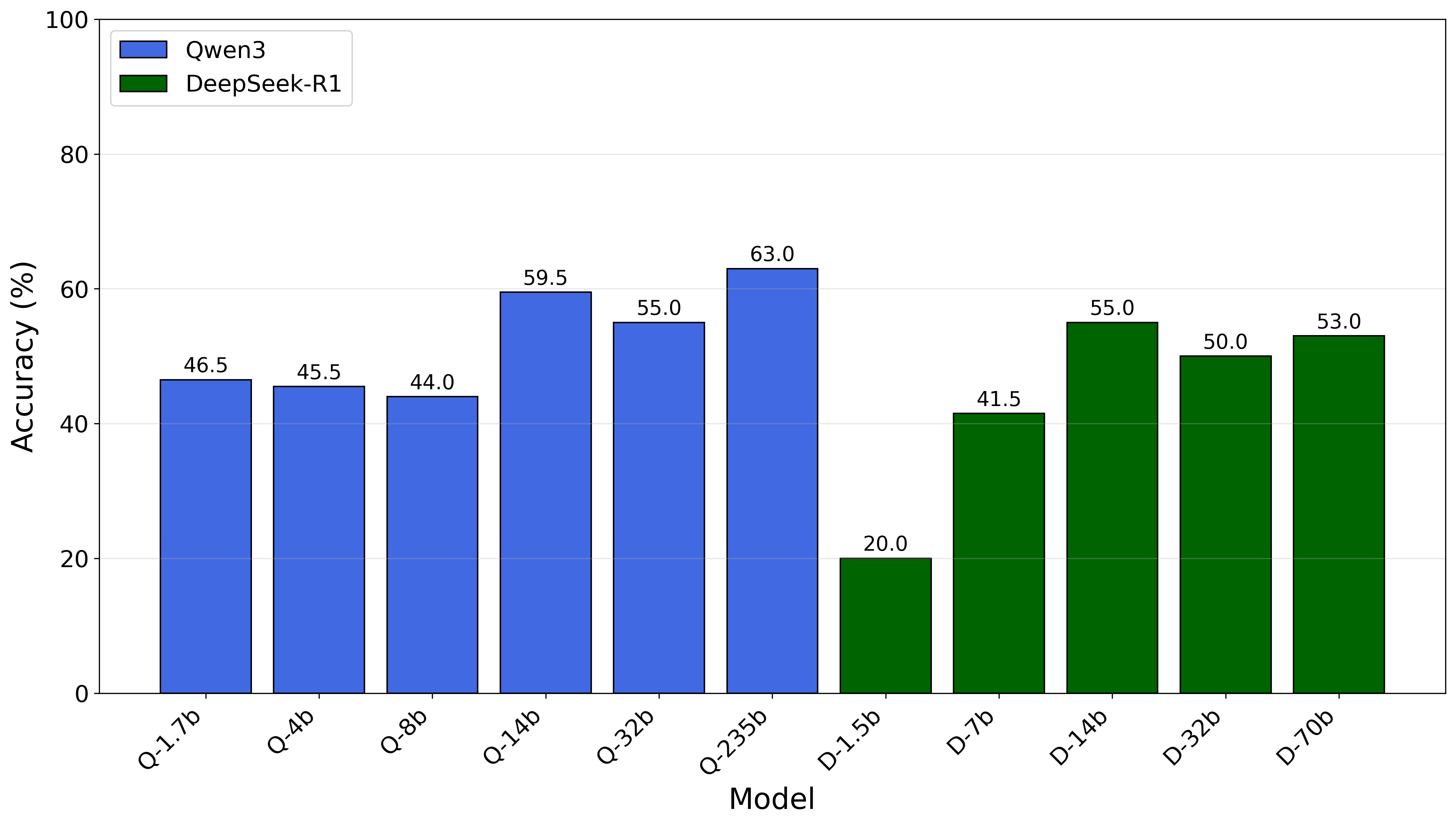}
\caption{Chief (None)}
\end{subfigure}
\hfill
\begin{subfigure}[b]{0.24\textwidth}
\centering
\includegraphics[width=\textwidth]{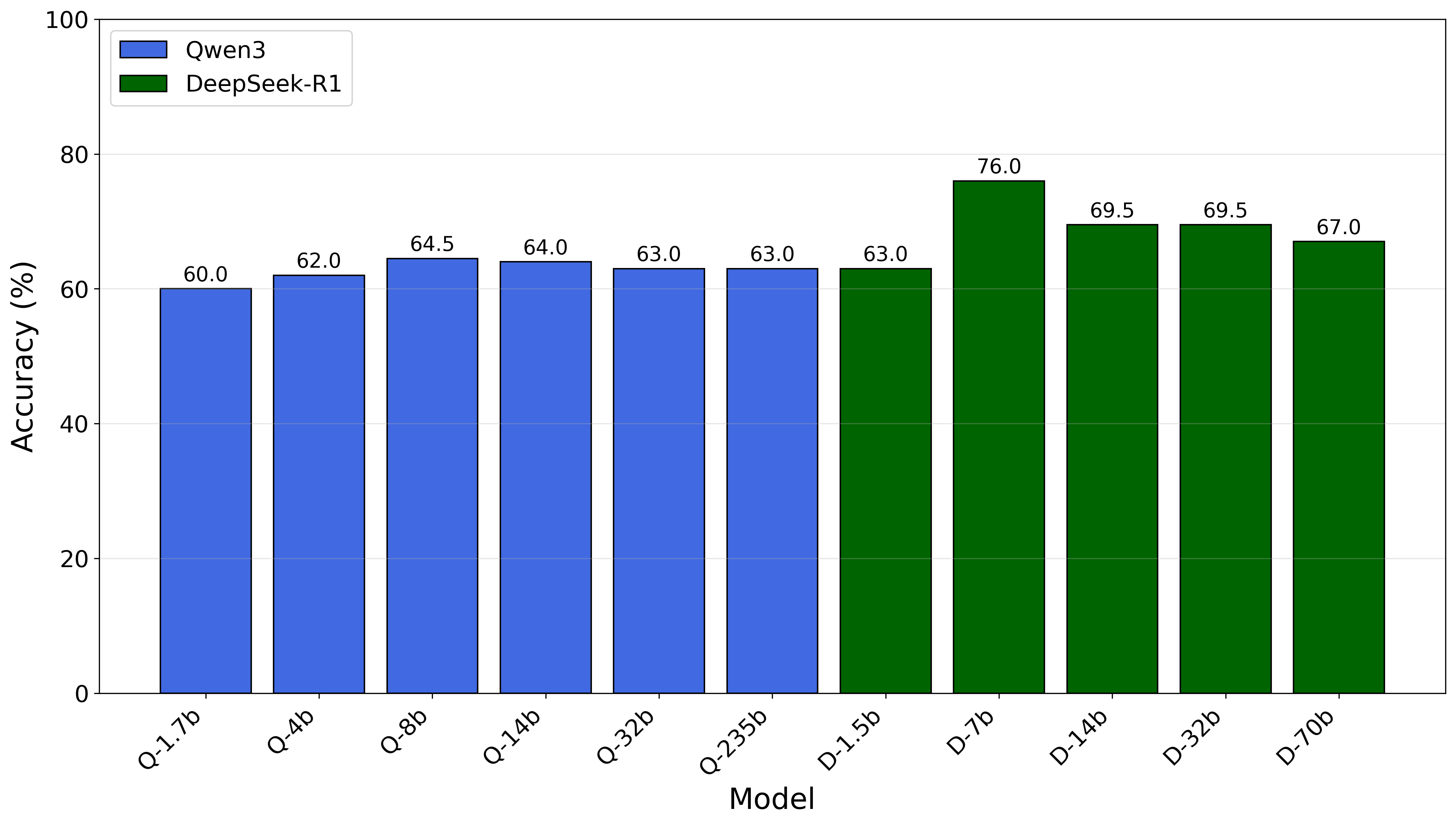}
\caption{Chief (inf)}
\end{subfigure}
\caption{Hematology performance across thinking budgets}
\label{fig:hematology_complete}
\end{figure}

\subsection{Infectious Diseases}

\begin{figure}[H]
\centering
\begin{subfigure}[b]{0.24\textwidth}
\centering
\includegraphics[width=\textwidth]{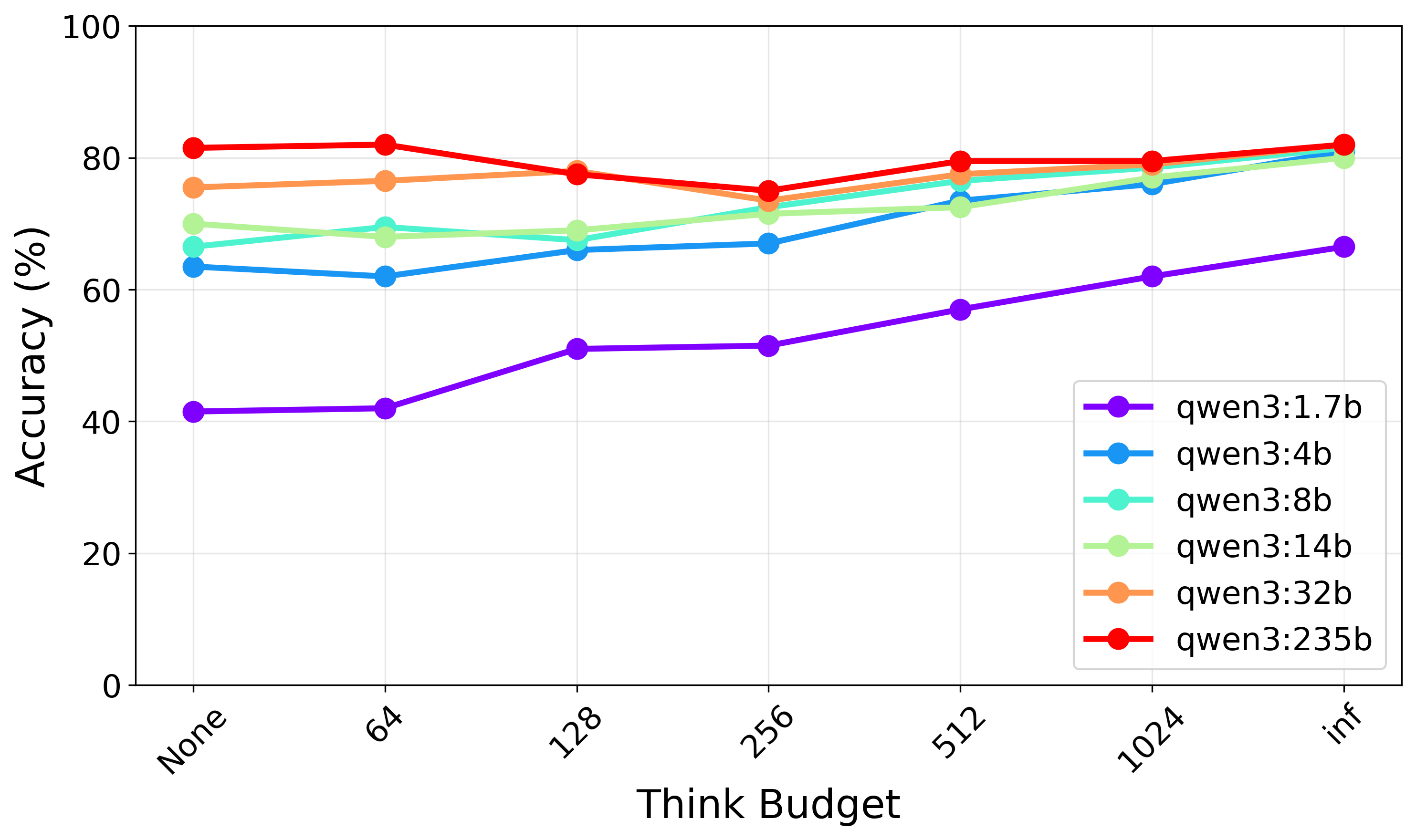}
\caption{Attending (Qwen3)}
\end{subfigure}
\hfill
\begin{subfigure}[b]{0.24\textwidth}
\centering
\includegraphics[width=\textwidth]{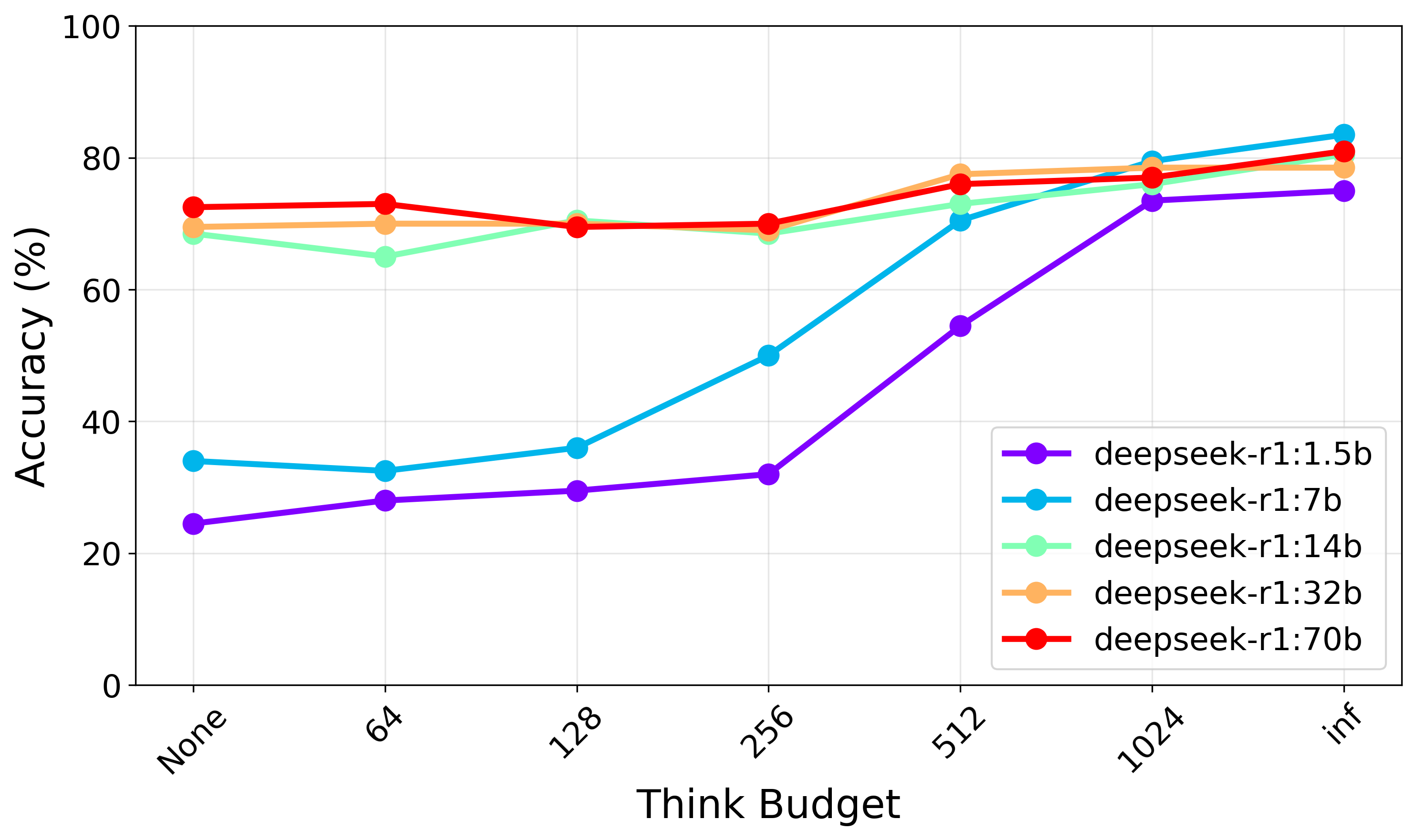}
\caption{Attending (DeepSeek-R1)}
\end{subfigure}
\hfill
\begin{subfigure}[b]{0.24\textwidth}
\centering
\includegraphics[width=\textwidth]{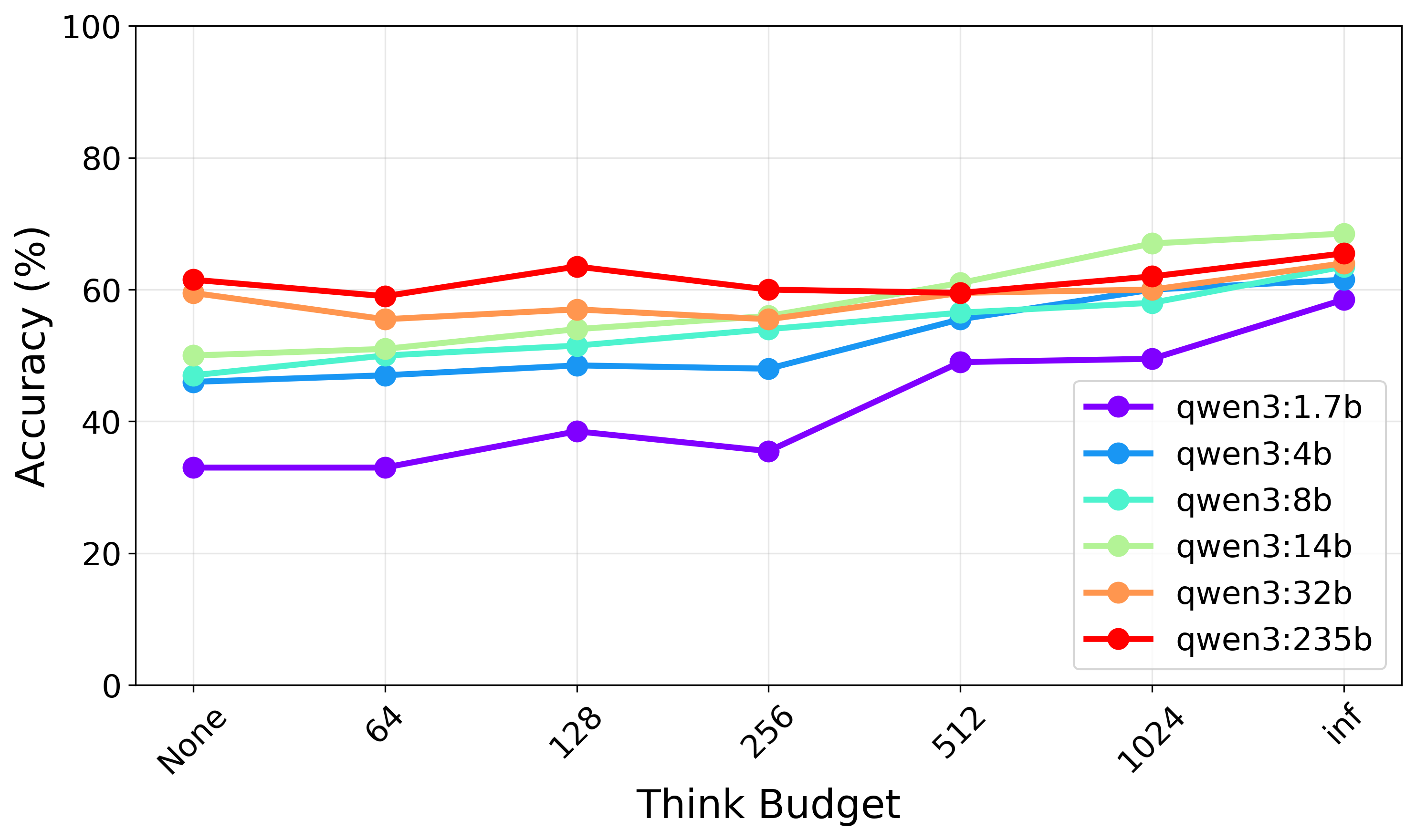}
\caption{Chief (Qwen3)}
\end{subfigure}
\hfill
\begin{subfigure}[b]{0.24\textwidth}
\centering
\includegraphics[width=\textwidth]{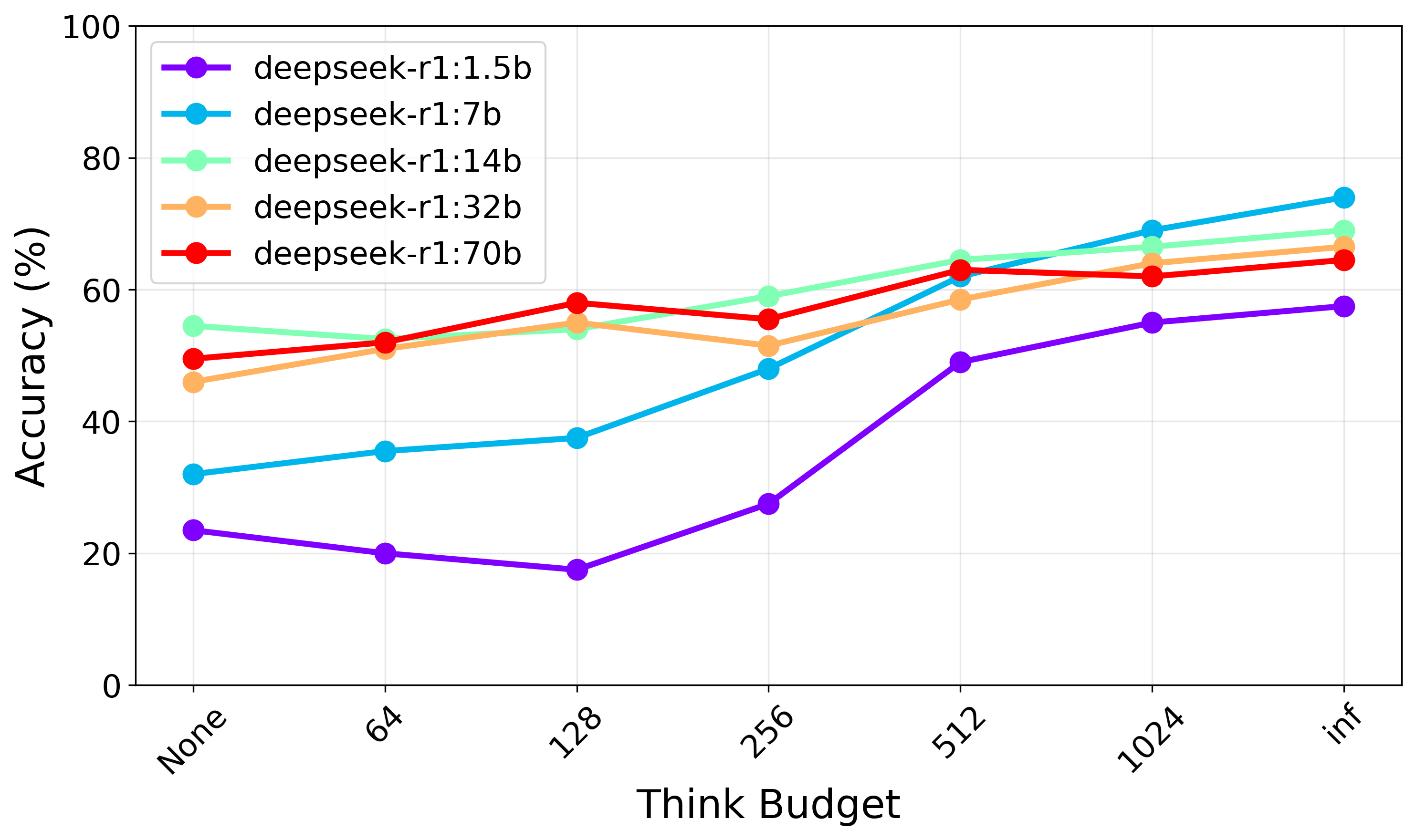}
\caption{Chief (DeepSeek-R1)}
\end{subfigure}

\vspace{0.5em}

\begin{subfigure}[b]{0.24\textwidth}
\centering
\includegraphics[width=\textwidth]{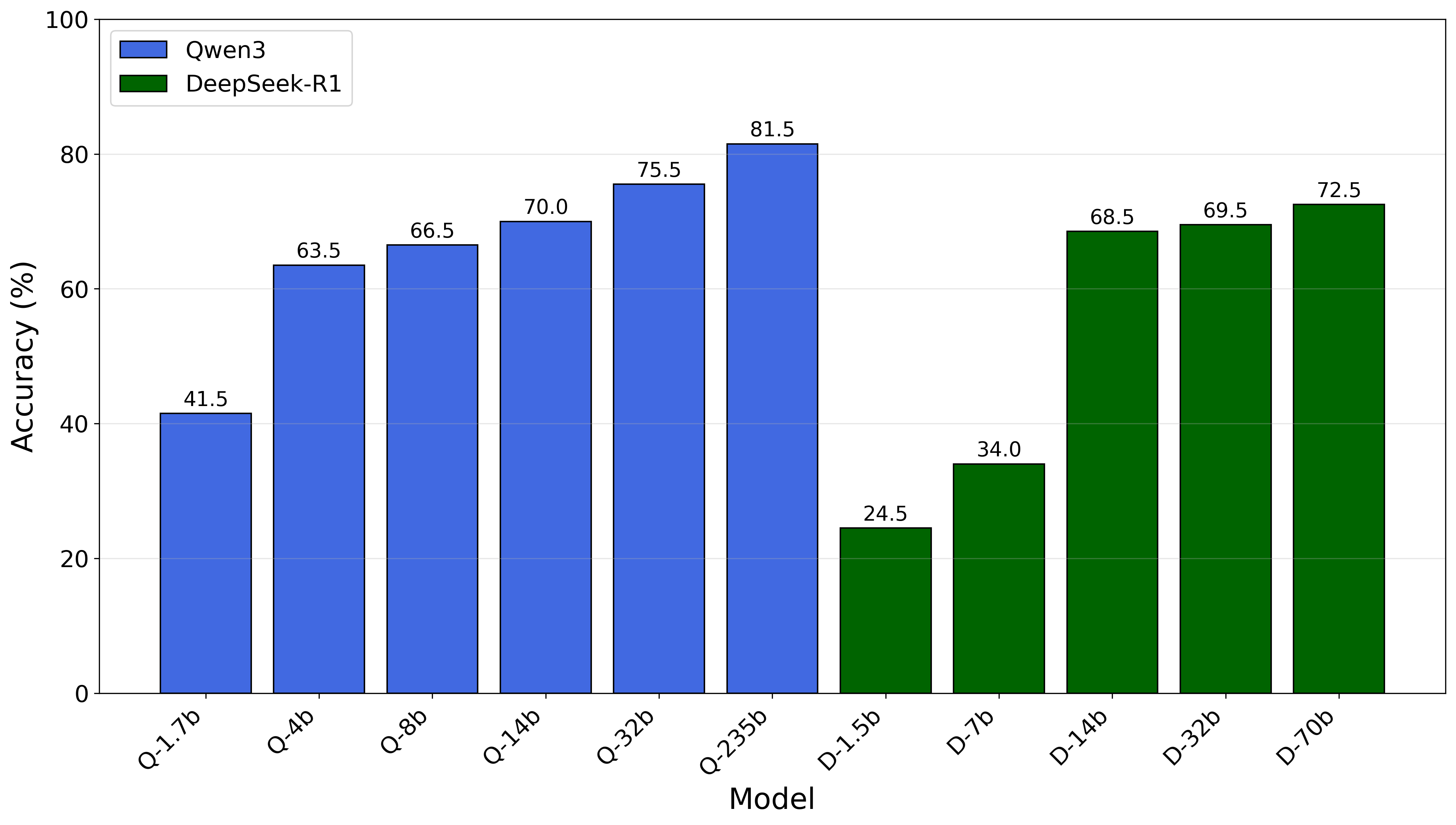}
\caption{Attending (None)}
\end{subfigure}
\hfill
\begin{subfigure}[b]{0.24\textwidth}
\centering
\includegraphics[width=\textwidth]{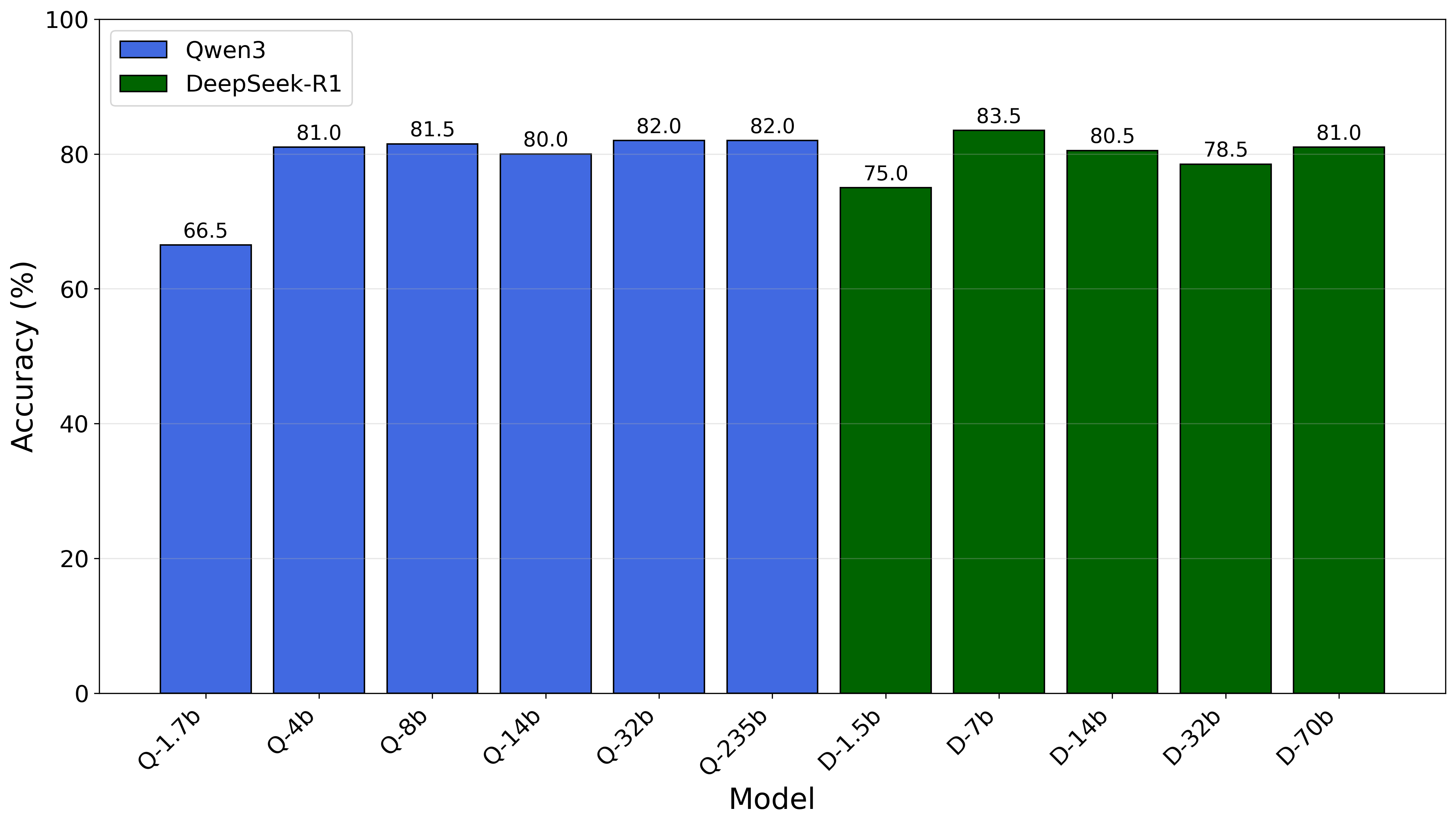}
\caption{Attending (inf)}
\end{subfigure}
\hfill
\begin{subfigure}[b]{0.24\textwidth}
\centering
\includegraphics[width=\textwidth]{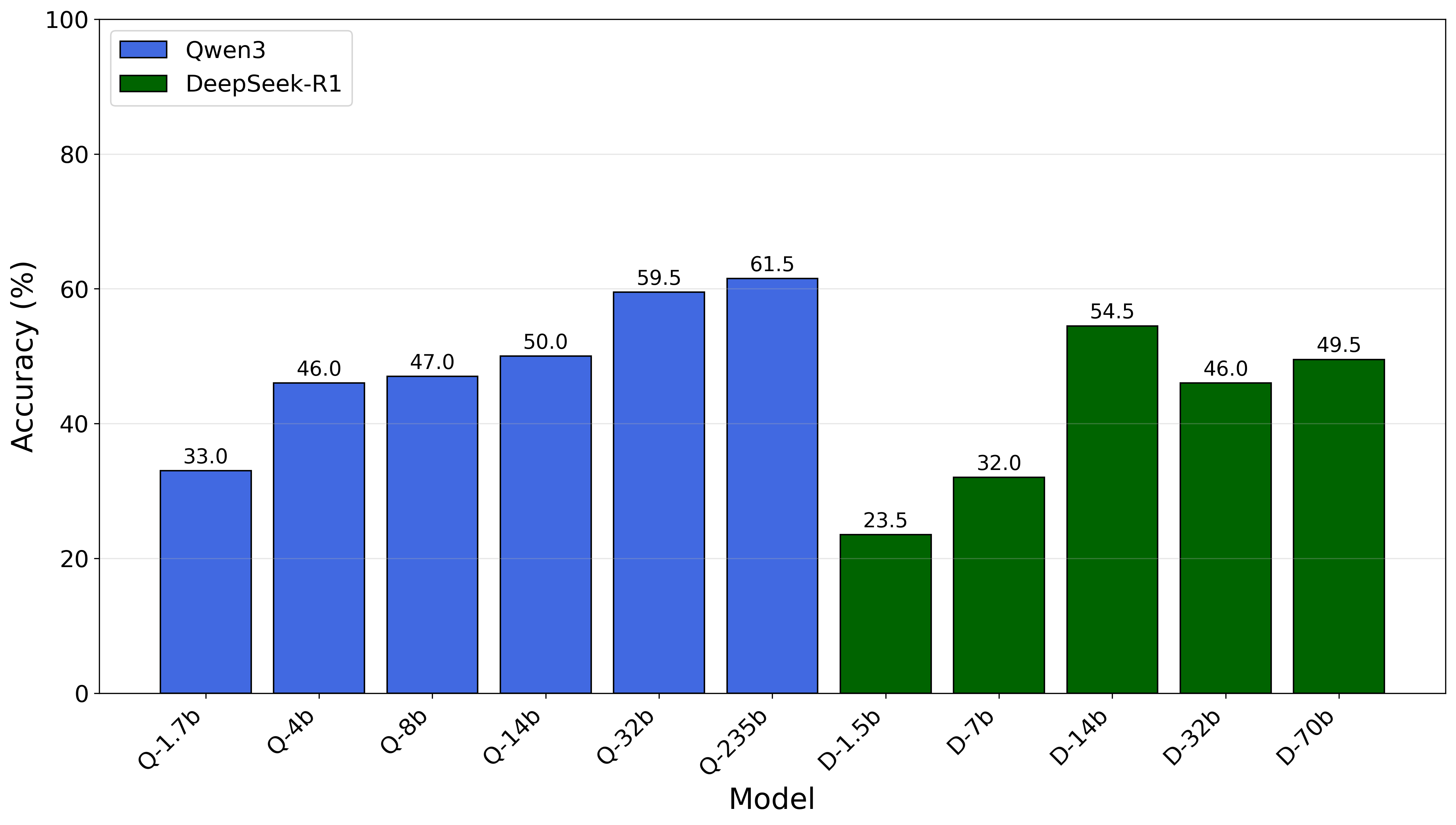}
\caption{Chief (None)}
\end{subfigure}
\hfill
\begin{subfigure}[b]{0.24\textwidth}
\centering
\includegraphics[width=\textwidth]{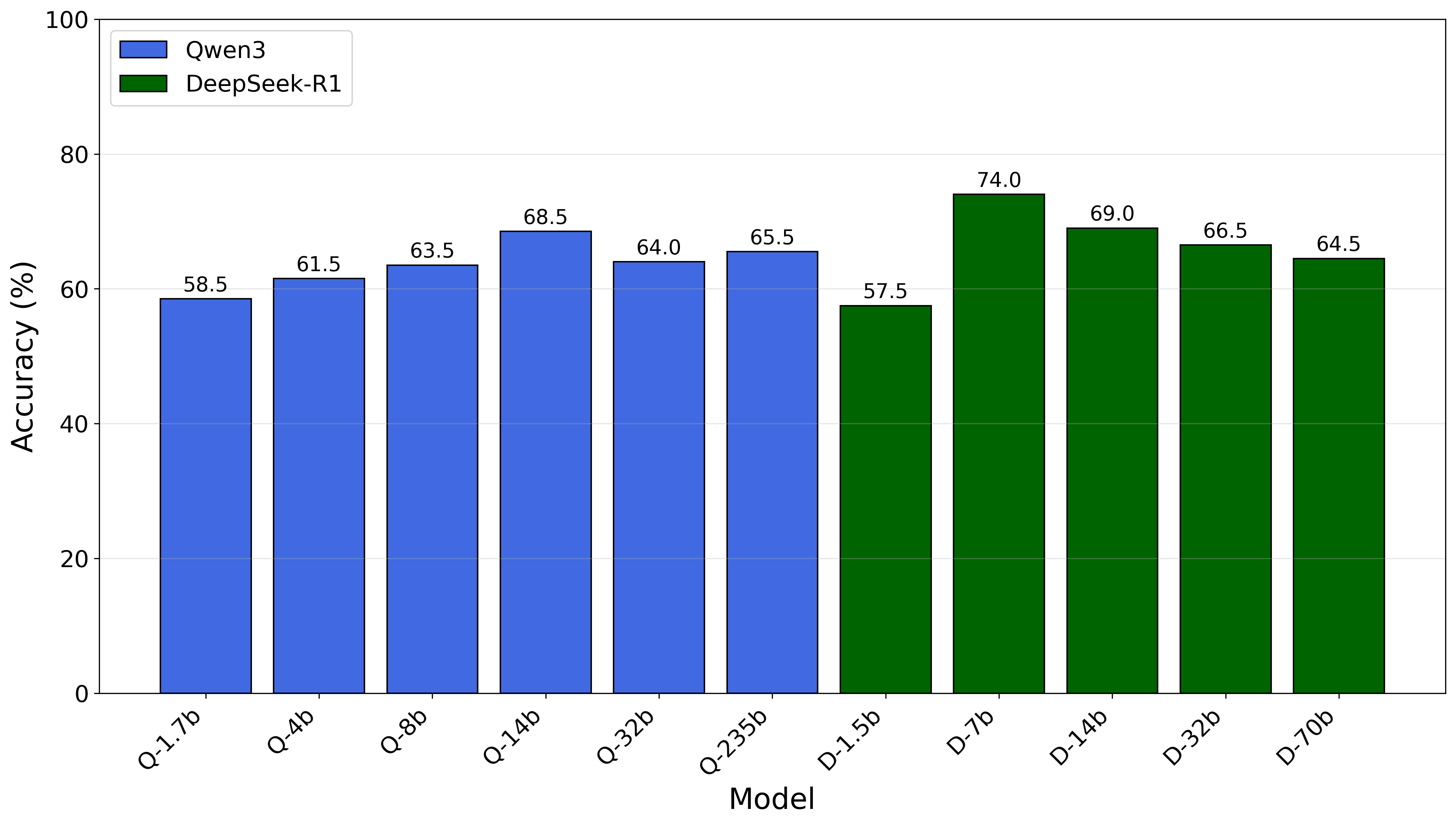}
\caption{Chief (inf)}
\end{subfigure}
\caption{Infectious Diseases performance across thinking budgets}
\label{fig:infectious_diseases_complete}
\end{figure}

\subsection{Nephrology}

\begin{figure}[H]
\centering
\begin{subfigure}[b]{0.24\textwidth}
\centering
\includegraphics[width=\textwidth]{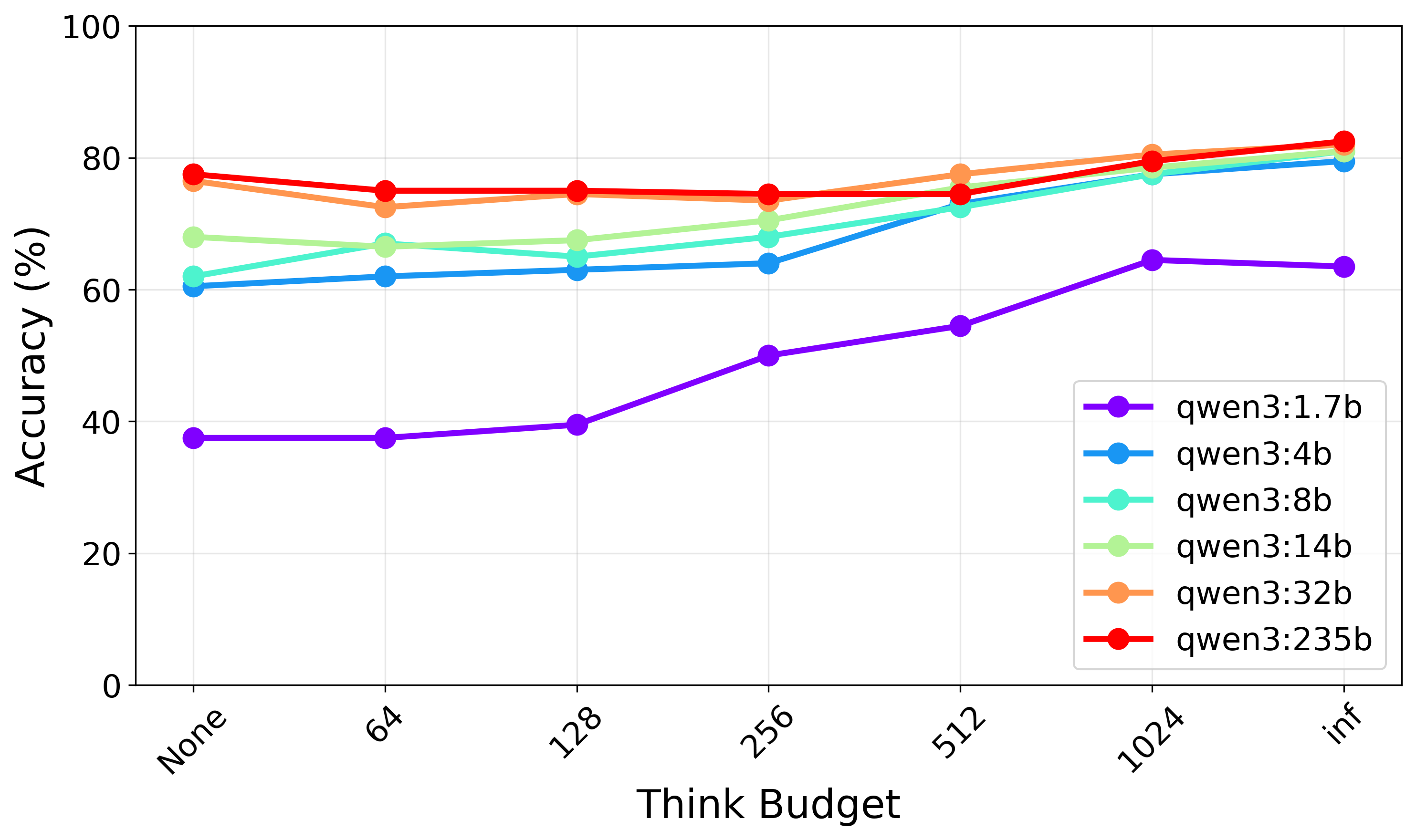}
\caption{Attending (Qwen3)}
\end{subfigure}
\hfill
\begin{subfigure}[b]{0.24\textwidth}
\centering
\includegraphics[width=\textwidth]{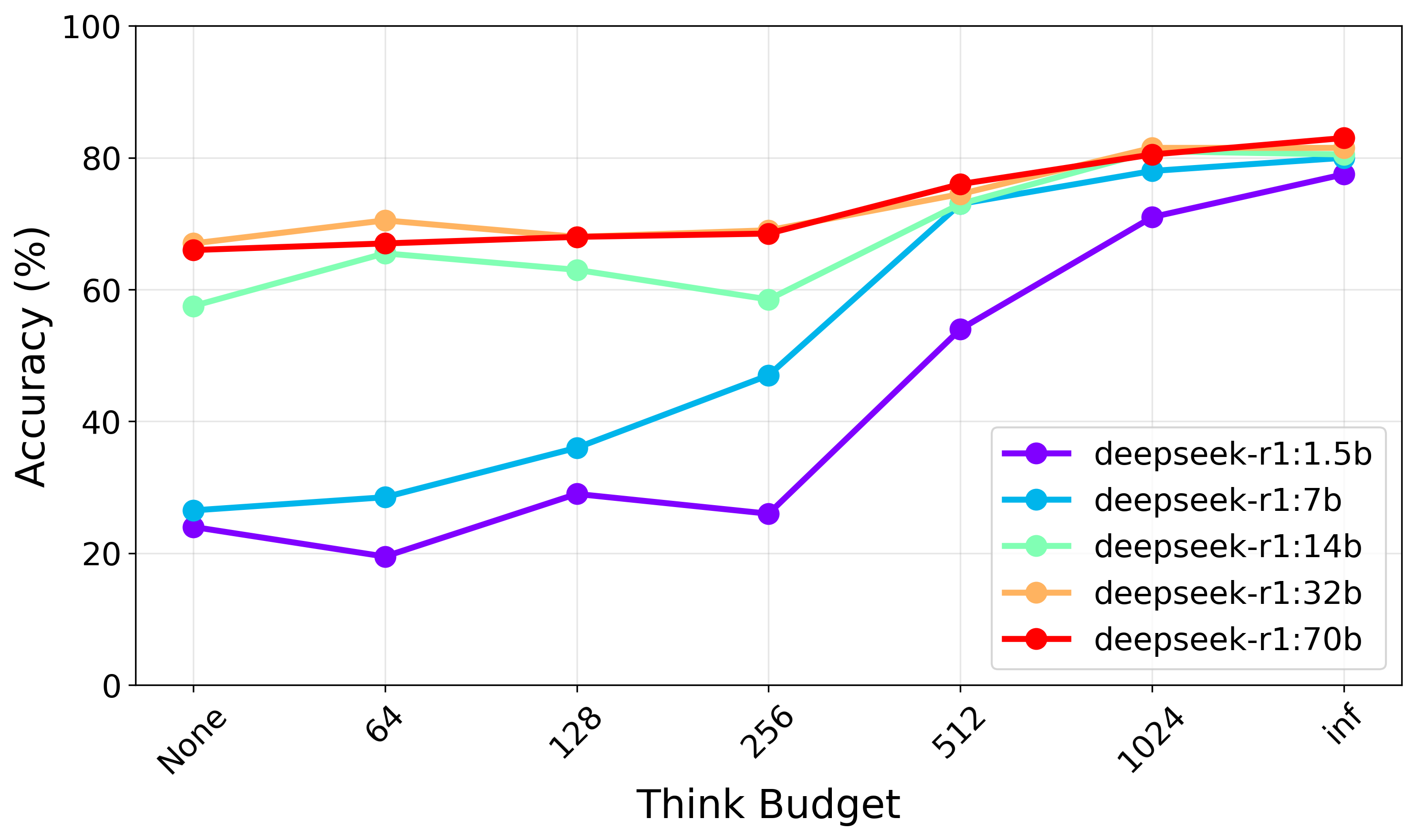}
\caption{Attending (DeepSeek-R1)}
\end{subfigure}
\hfill
\begin{subfigure}[b]{0.24\textwidth}
\centering
\includegraphics[width=\textwidth]{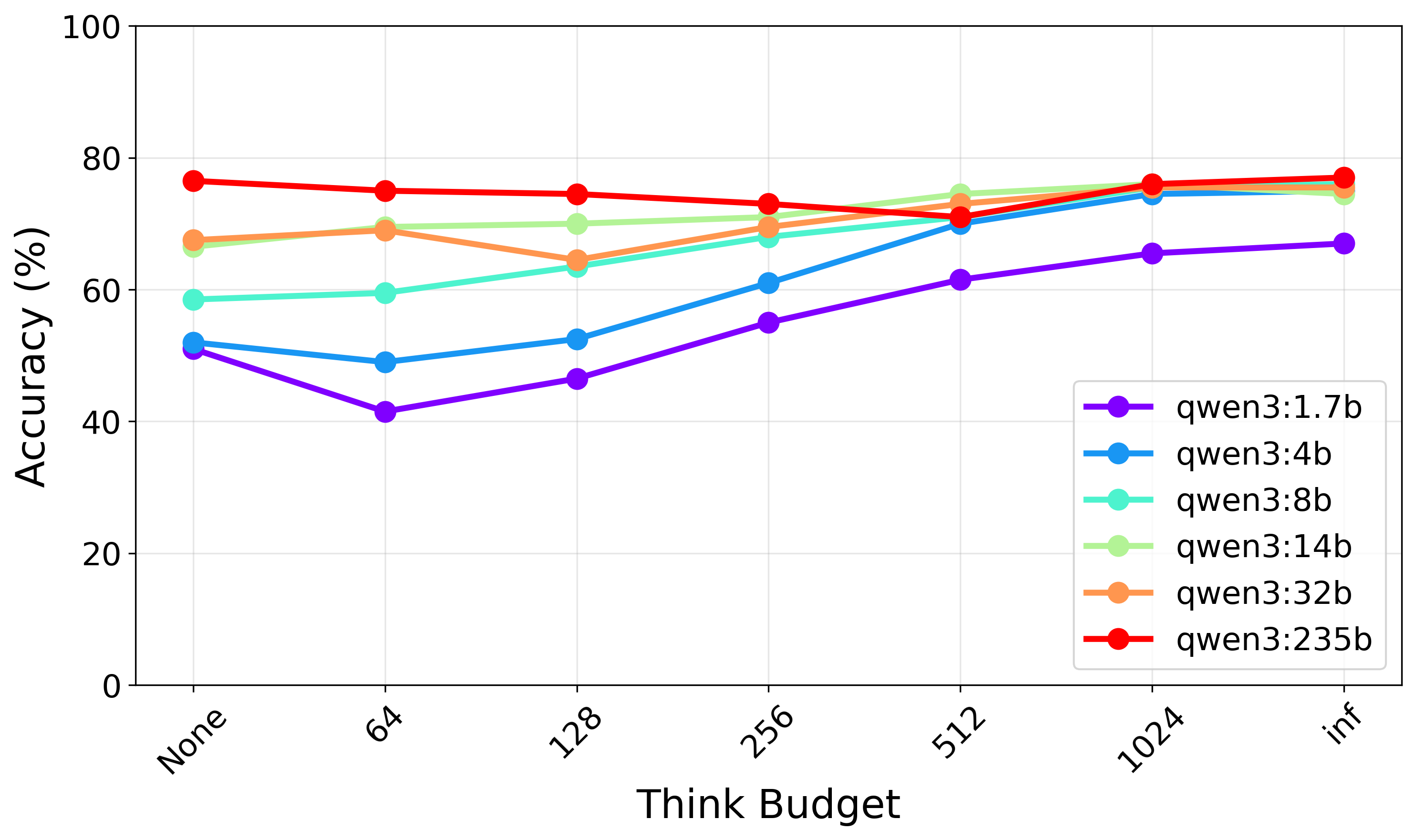}
\caption{Chief (Qwen3)}
\end{subfigure}
\hfill
\begin{subfigure}[b]{0.24\textwidth}
\centering
\includegraphics[width=\textwidth]{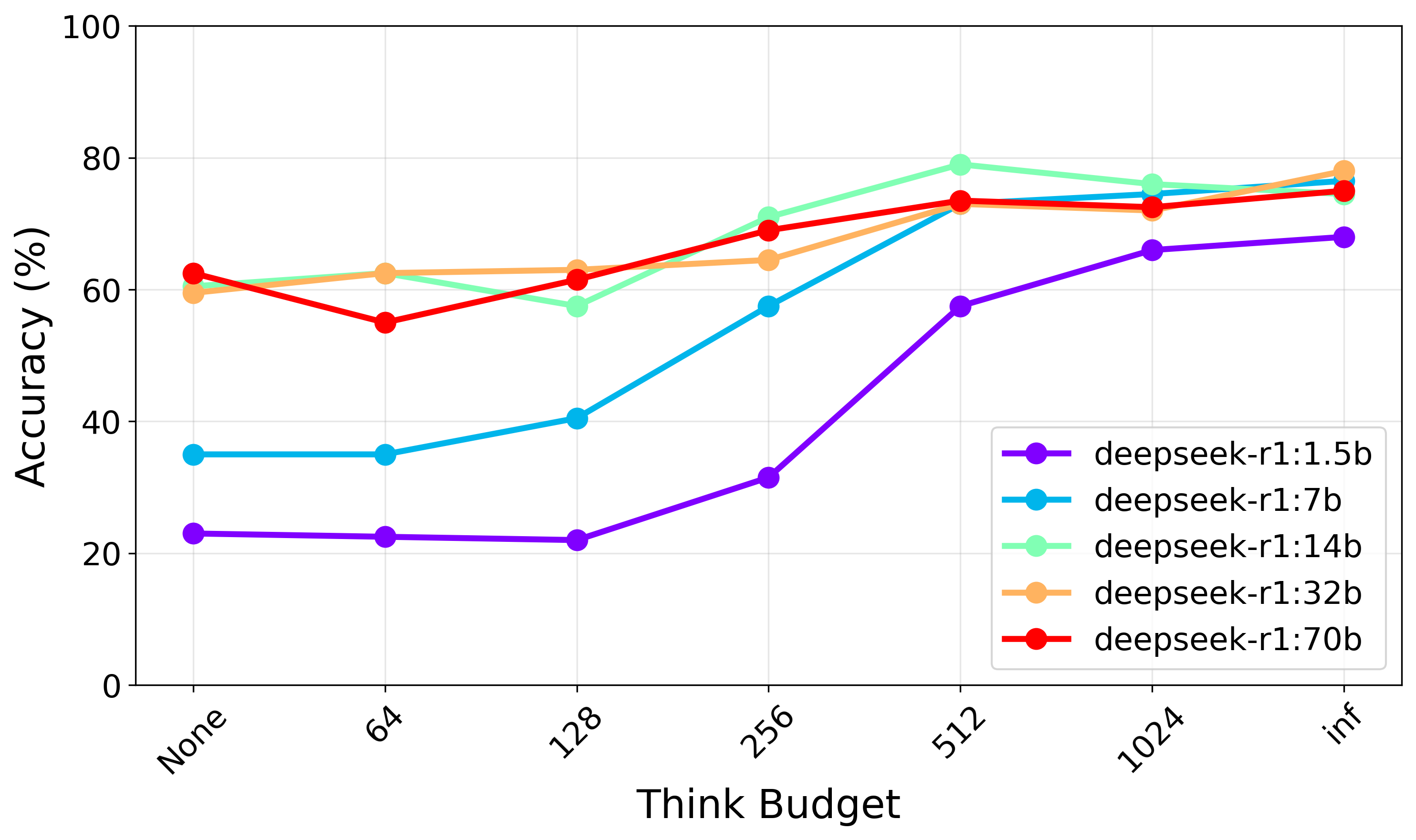}
\caption{Chief (DeepSeek-R1)}
\end{subfigure}

\vspace{0.5em}

\begin{subfigure}[b]{0.24\textwidth}
\centering
\includegraphics[width=\textwidth]{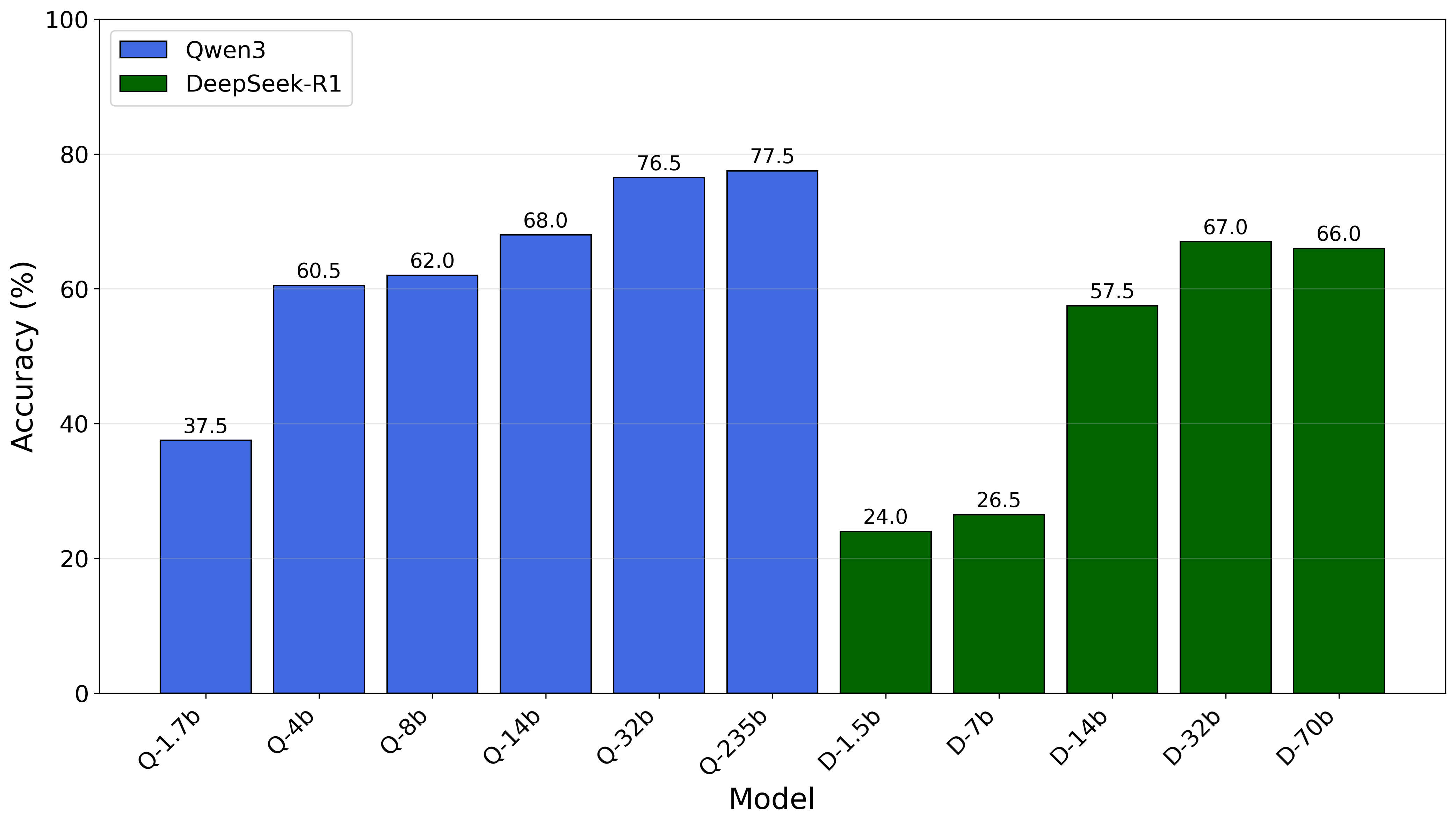}
\caption{Attending (None)}
\end{subfigure}
\hfill
\begin{subfigure}[b]{0.24\textwidth}
\centering
\includegraphics[width=\textwidth]{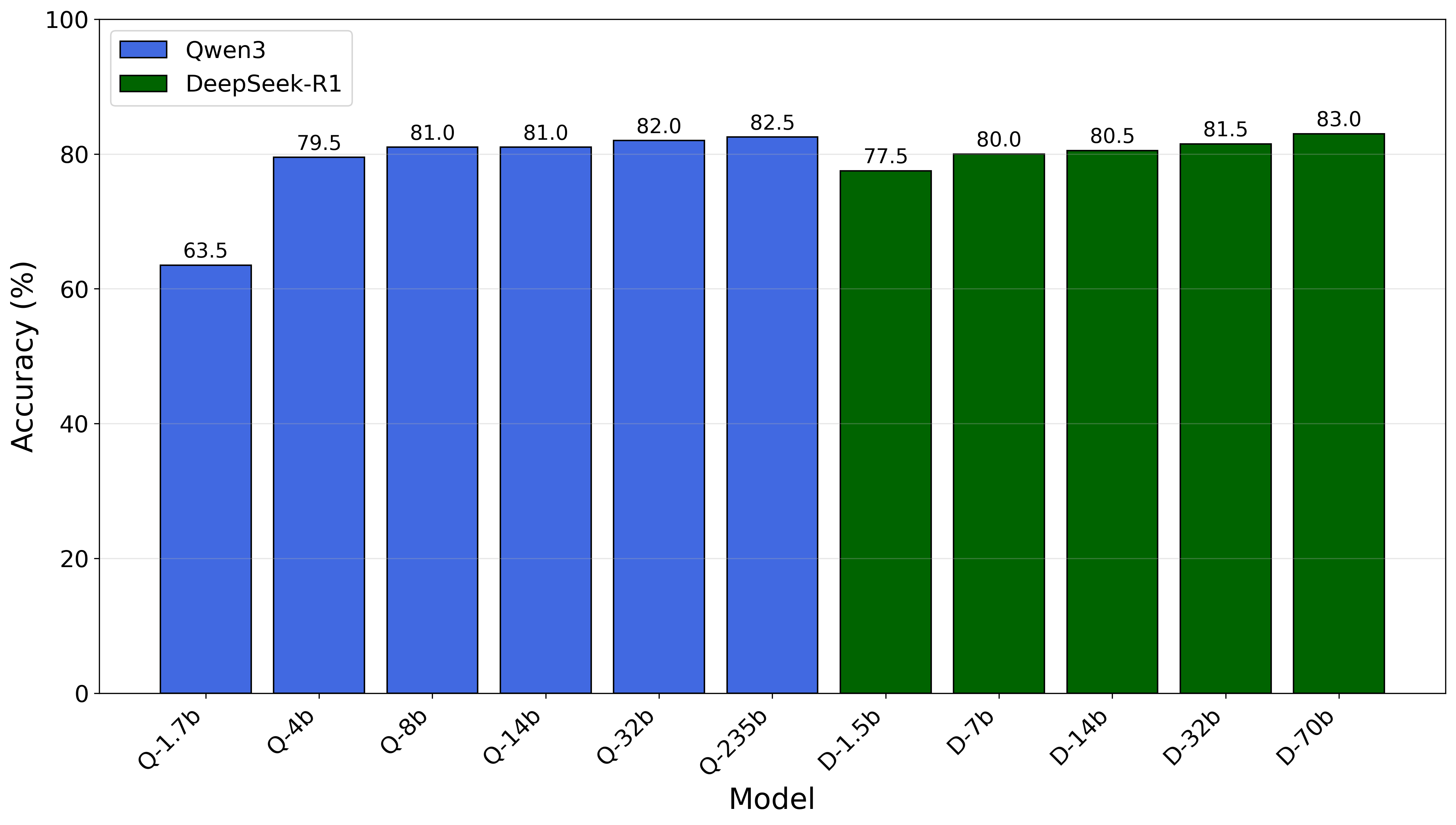}
\caption{Attending (inf)}
\end{subfigure}
\hfill
\begin{subfigure}[b]{0.24\textwidth}
\centering
\includegraphics[width=\textwidth]{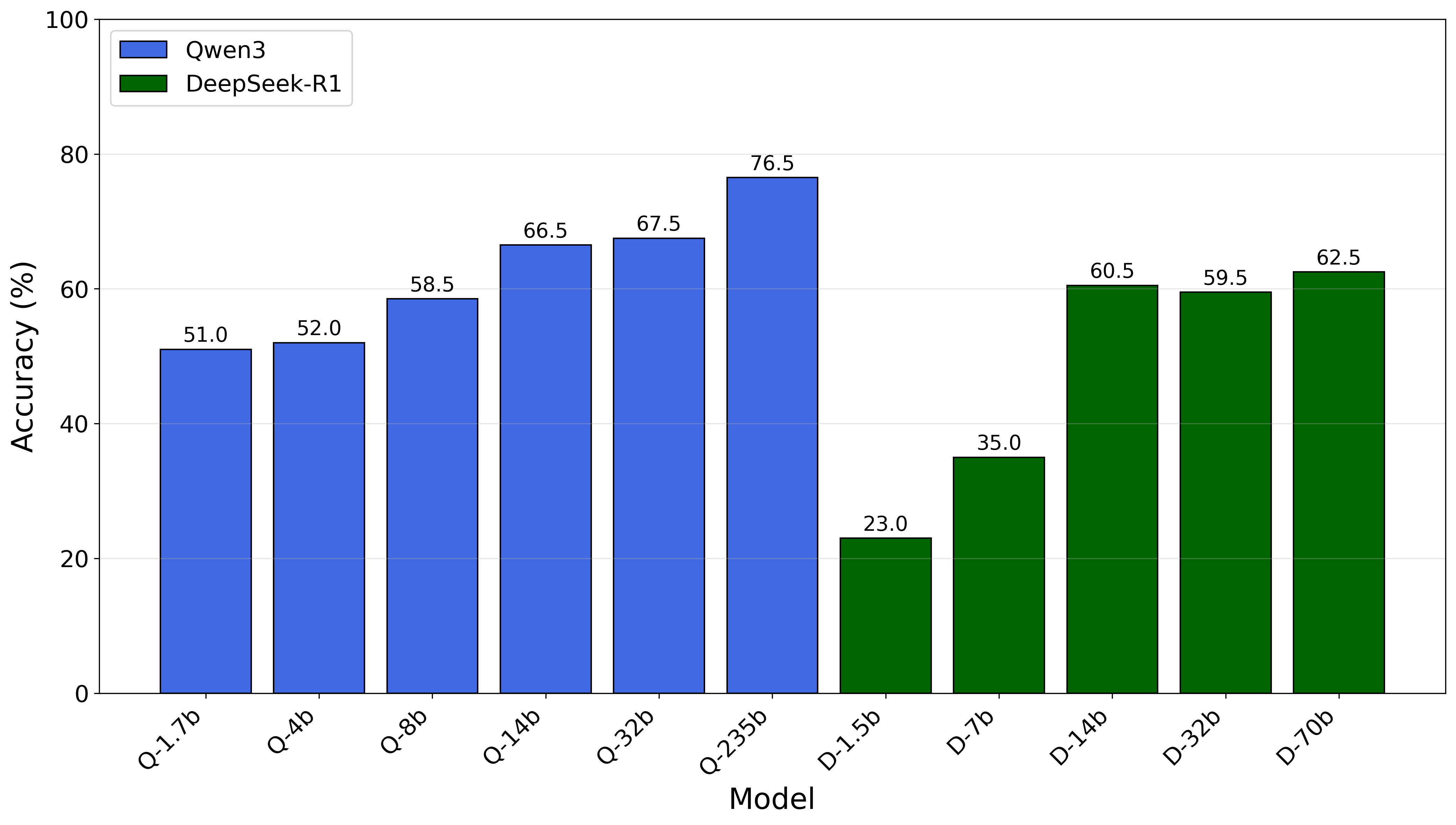}
\caption{Chief (None)}
\end{subfigure}
\hfill
\begin{subfigure}[b]{0.24\textwidth}
\centering
\includegraphics[width=\textwidth]{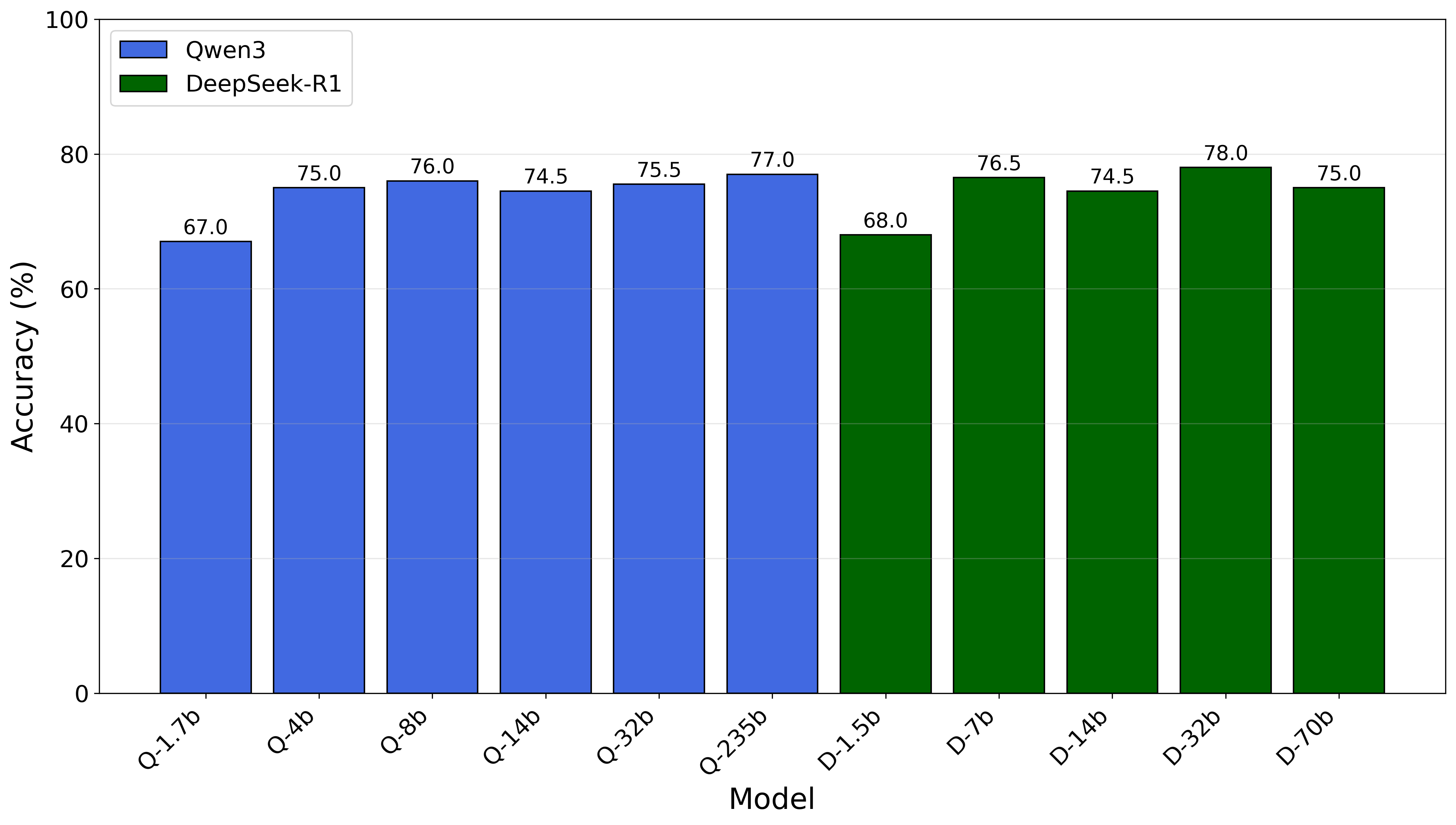}
\caption{Chief (inf)}
\end{subfigure}
\caption{Nephrology performance across thinking budgets}
\label{fig:nephrology_complete}
\end{figure}

\subsection{Neurology}

\begin{figure}[H]
\centering
\begin{subfigure}[b]{0.24\textwidth}
\centering
\includegraphics[width=\textwidth]{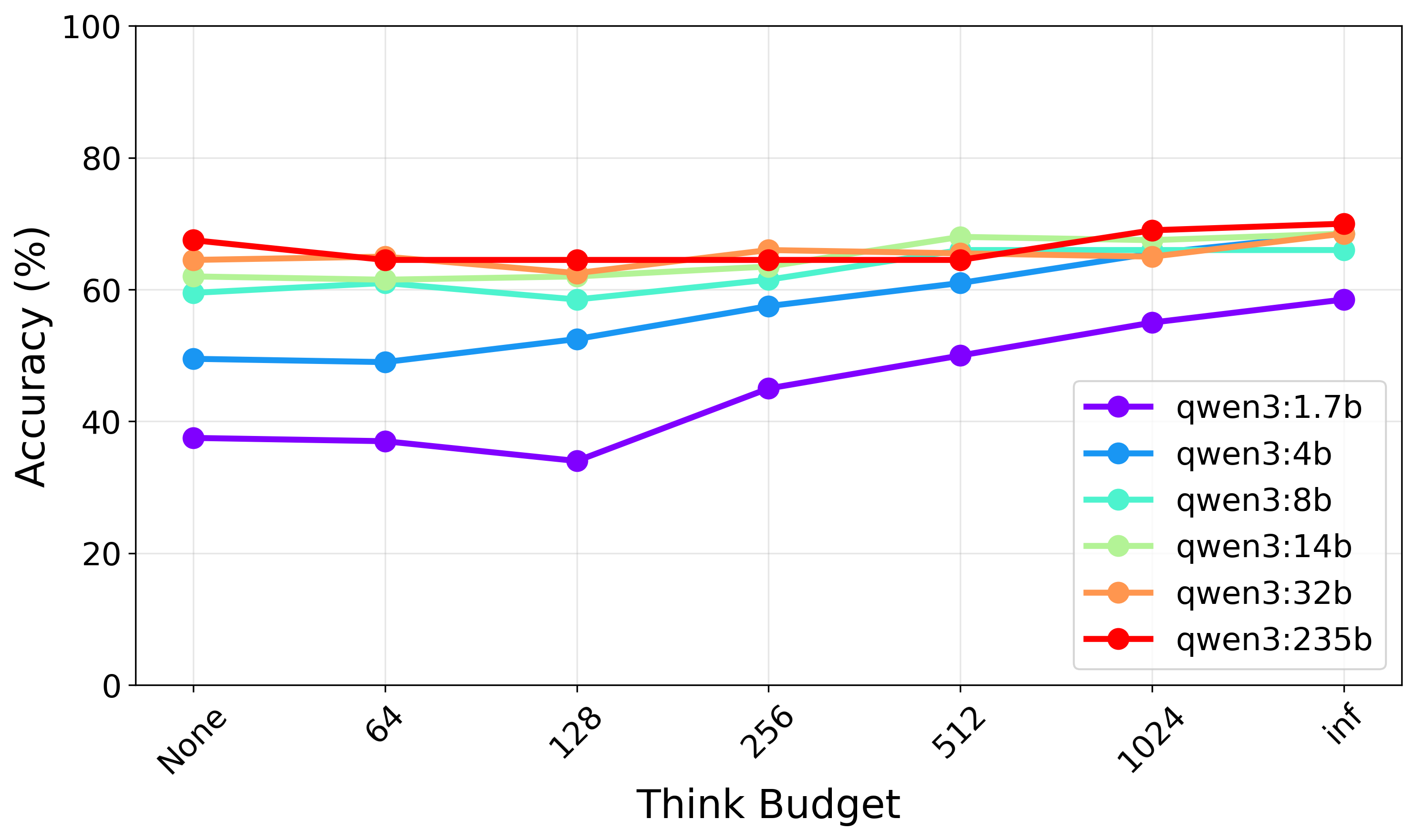}
\caption{Attending (Qwen3)}
\end{subfigure}
\hfill
\begin{subfigure}[b]{0.24\textwidth}
\centering
\includegraphics[width=\textwidth]{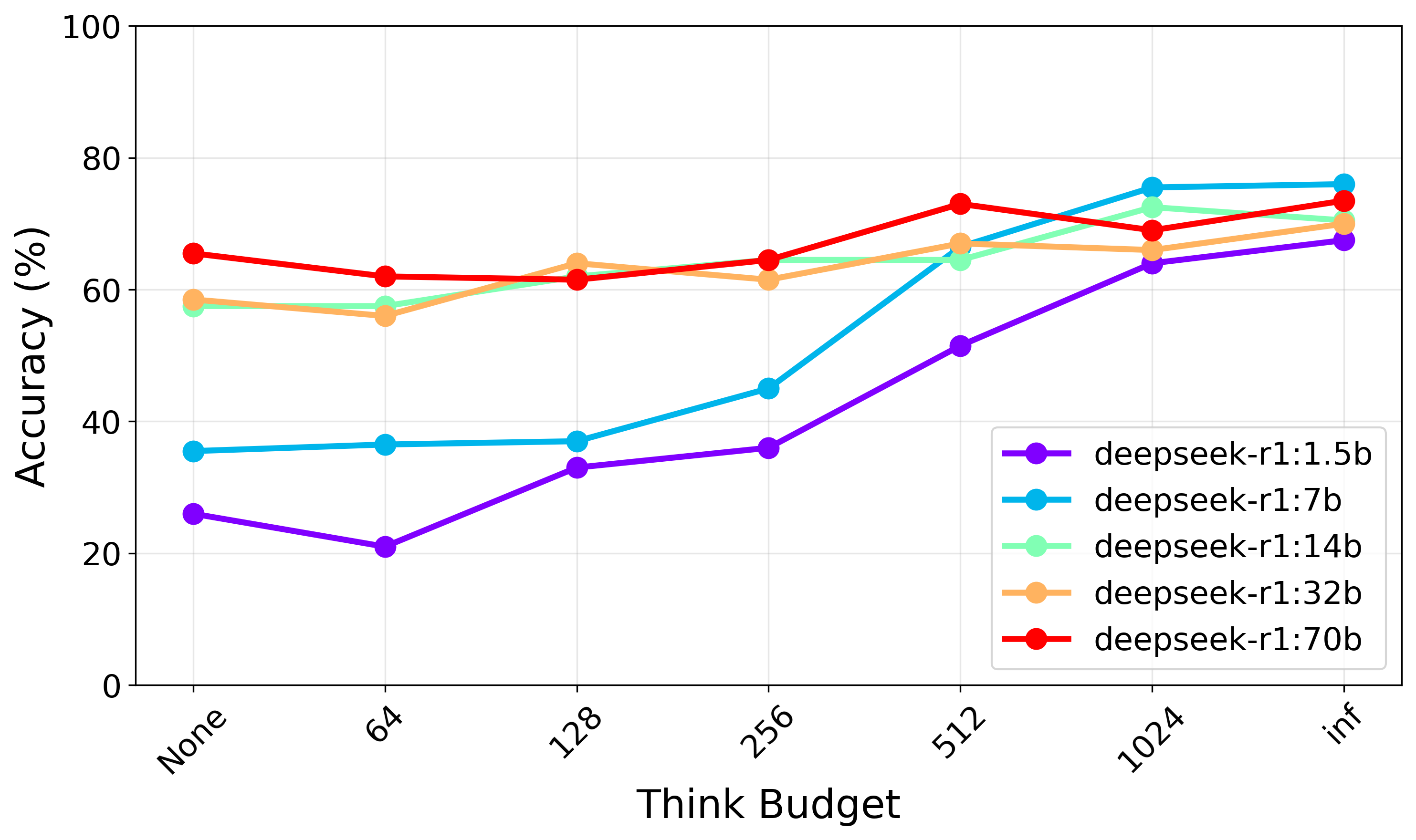}
\caption{Attending (DeepSeek-R1)}
\end{subfigure}
\hfill
\begin{subfigure}[b]{0.24\textwidth}
\centering
\includegraphics[width=\textwidth]{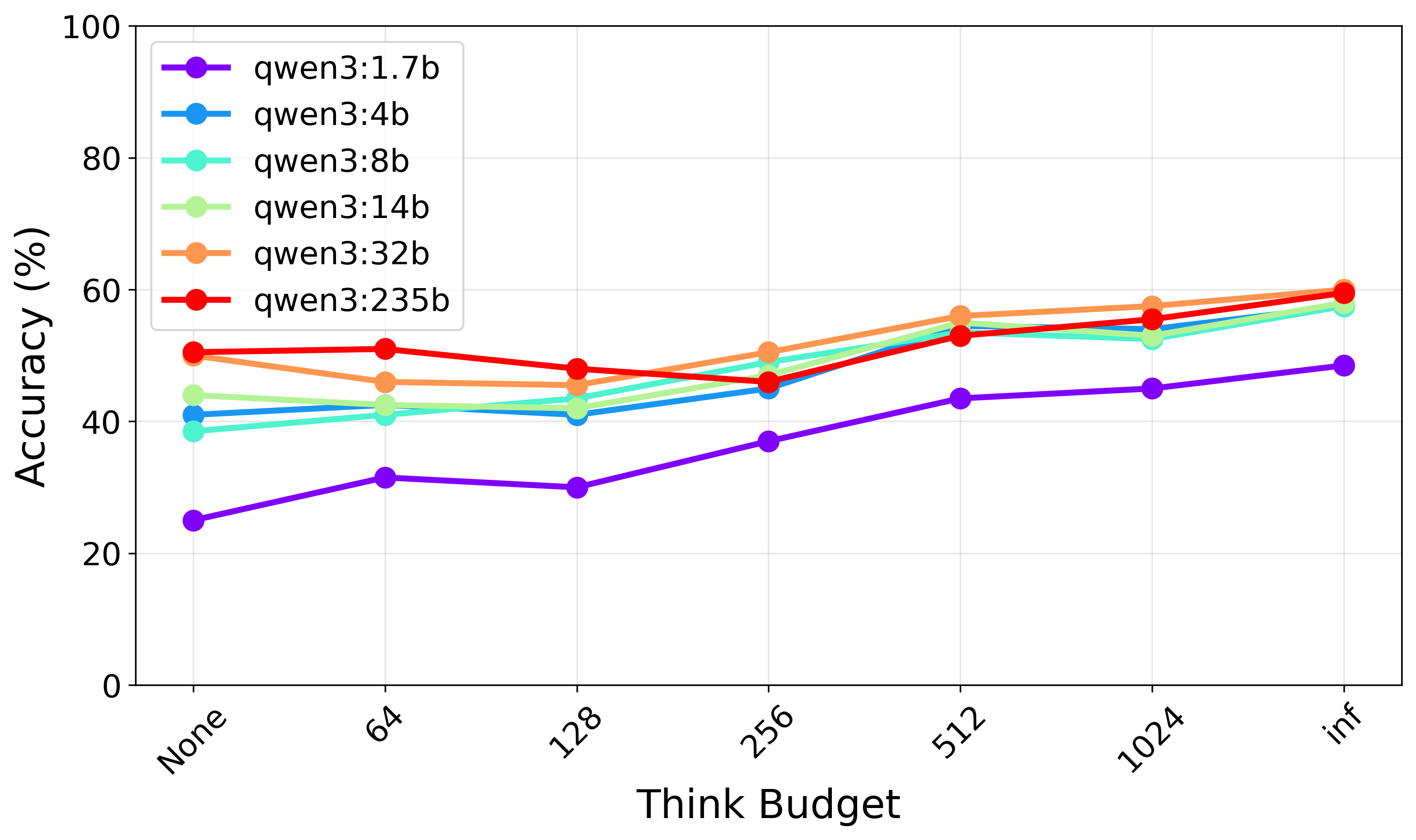}
\caption{Chief (Qwen3)}
\end{subfigure}
\hfill
\begin{subfigure}[b]{0.24\textwidth}
\centering
\includegraphics[width=\textwidth]{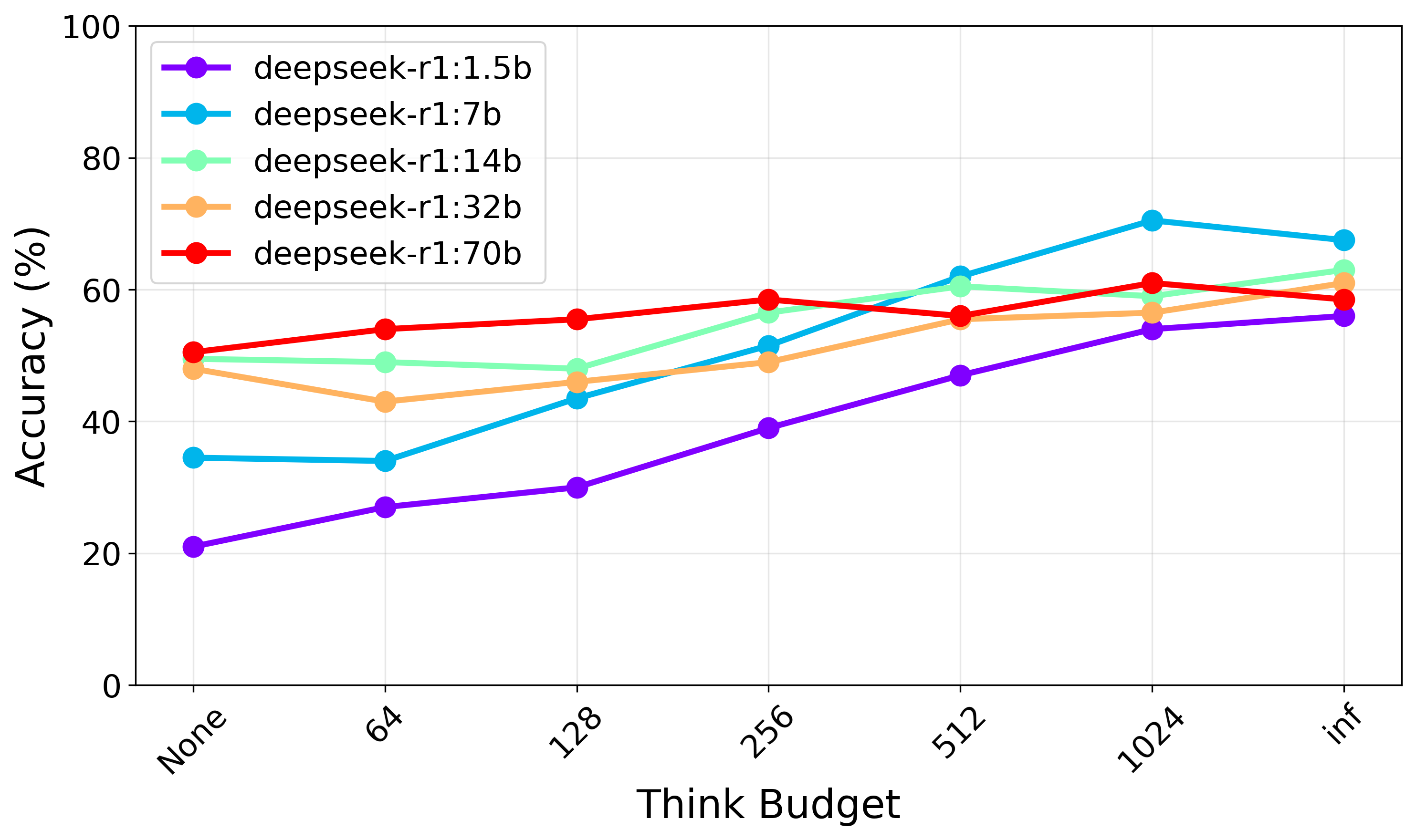}
\caption{Chief (DeepSeek-R1)}
\end{subfigure}

\vspace{0.5em}

\begin{subfigure}[b]{0.24\textwidth}
\centering
\includegraphics[width=\textwidth]{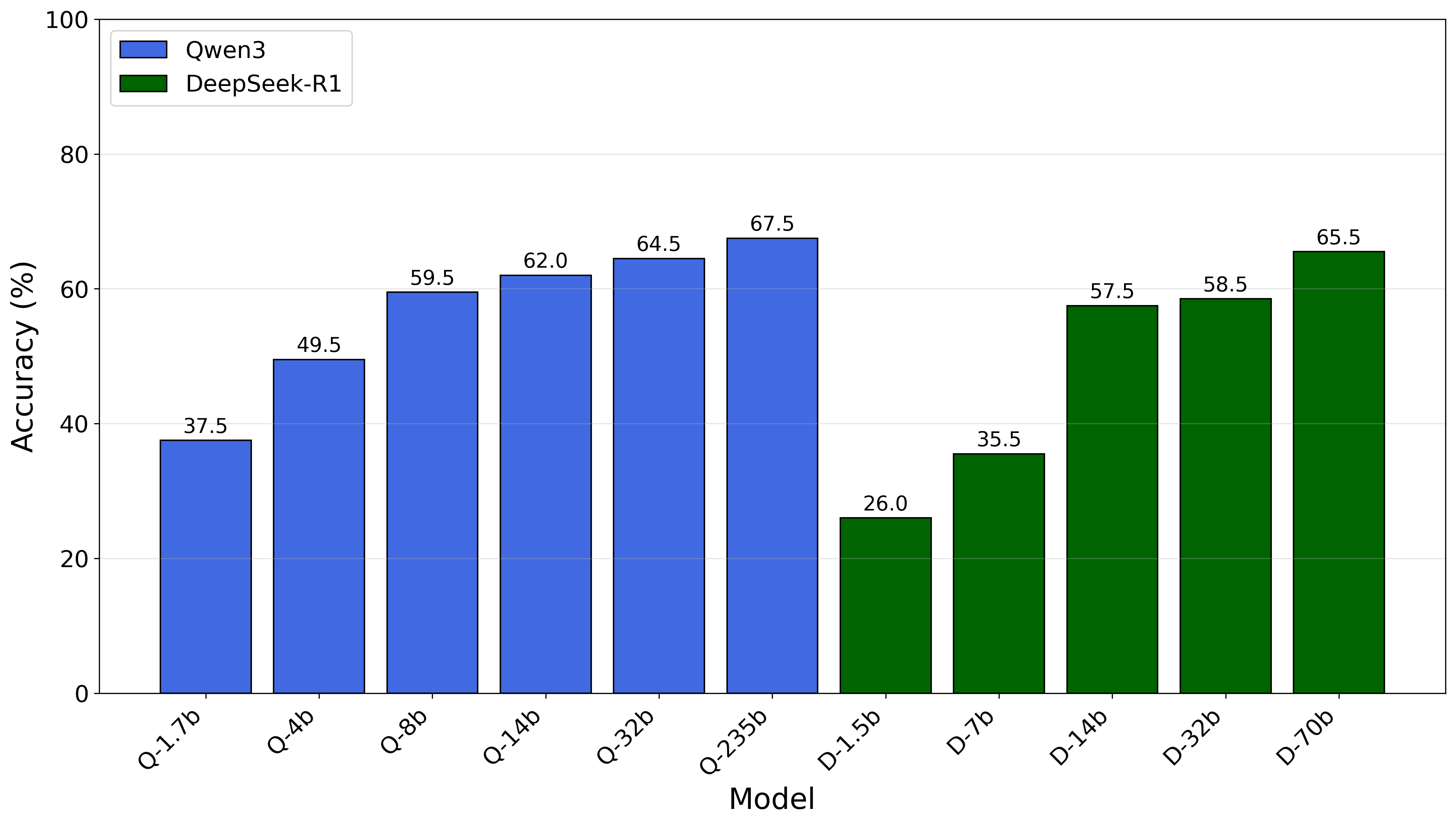}
\caption{Attending (None)}
\end{subfigure}
\hfill
\begin{subfigure}[b]{0.24\textwidth}
\centering
\includegraphics[width=\textwidth]{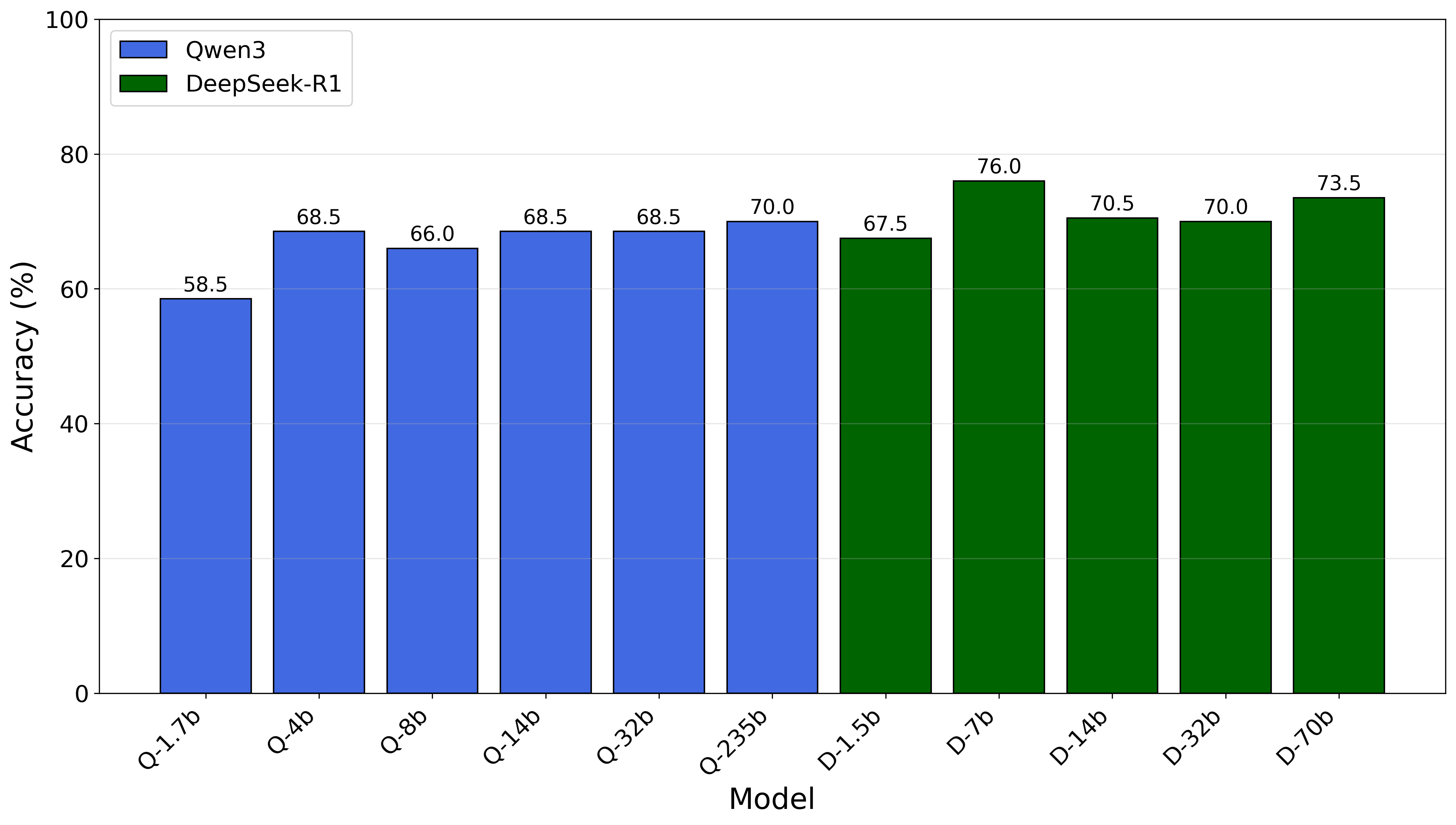}
\caption{Attending (inf)}
\end{subfigure}
\hfill
\begin{subfigure}[b]{0.24\textwidth}
\centering
\includegraphics[width=\textwidth]{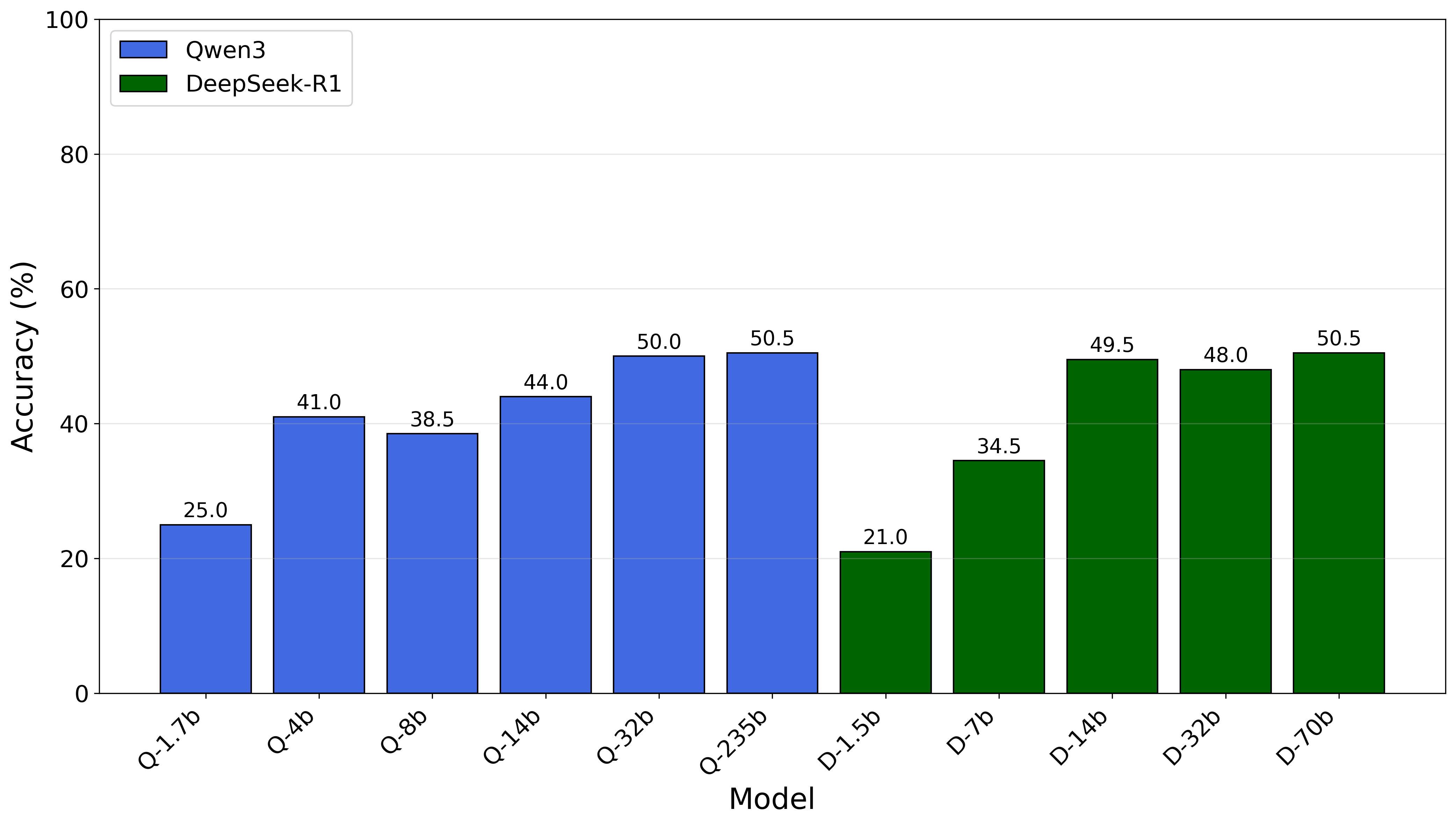}
\caption{Chief (None)}
\end{subfigure}
\hfill
\begin{subfigure}[b]{0.24\textwidth}
\centering
\includegraphics[width=\textwidth]{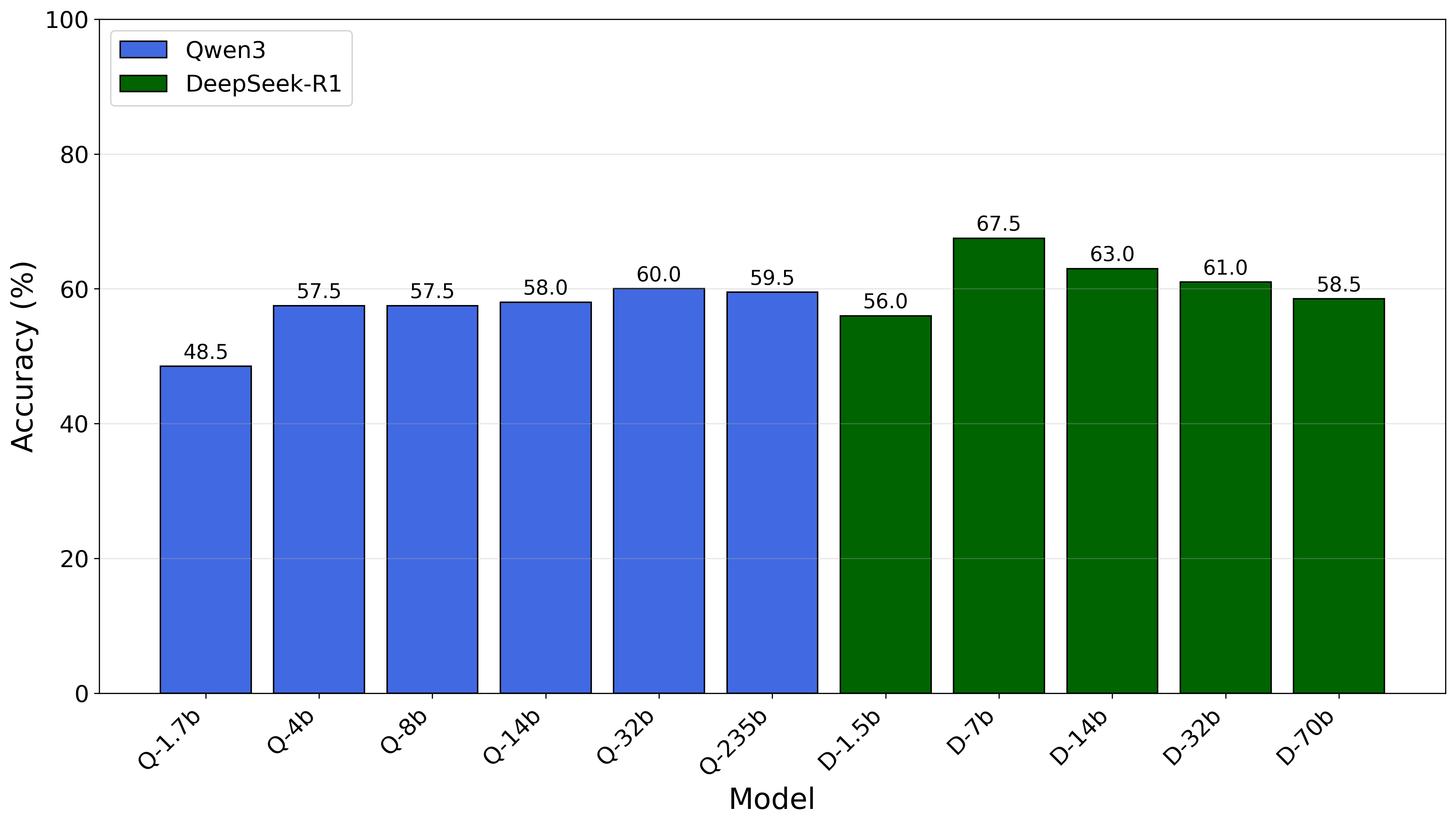}
\caption{Chief (inf)}
\end{subfigure}
\caption{Neurology performance across thinking budgets}
\label{fig:neurology_complete}
\end{figure}

\subsection{Respiratory}

\begin{figure}[H]
\centering
\begin{subfigure}[b]{0.24\textwidth}
\centering
\includegraphics[width=\textwidth]{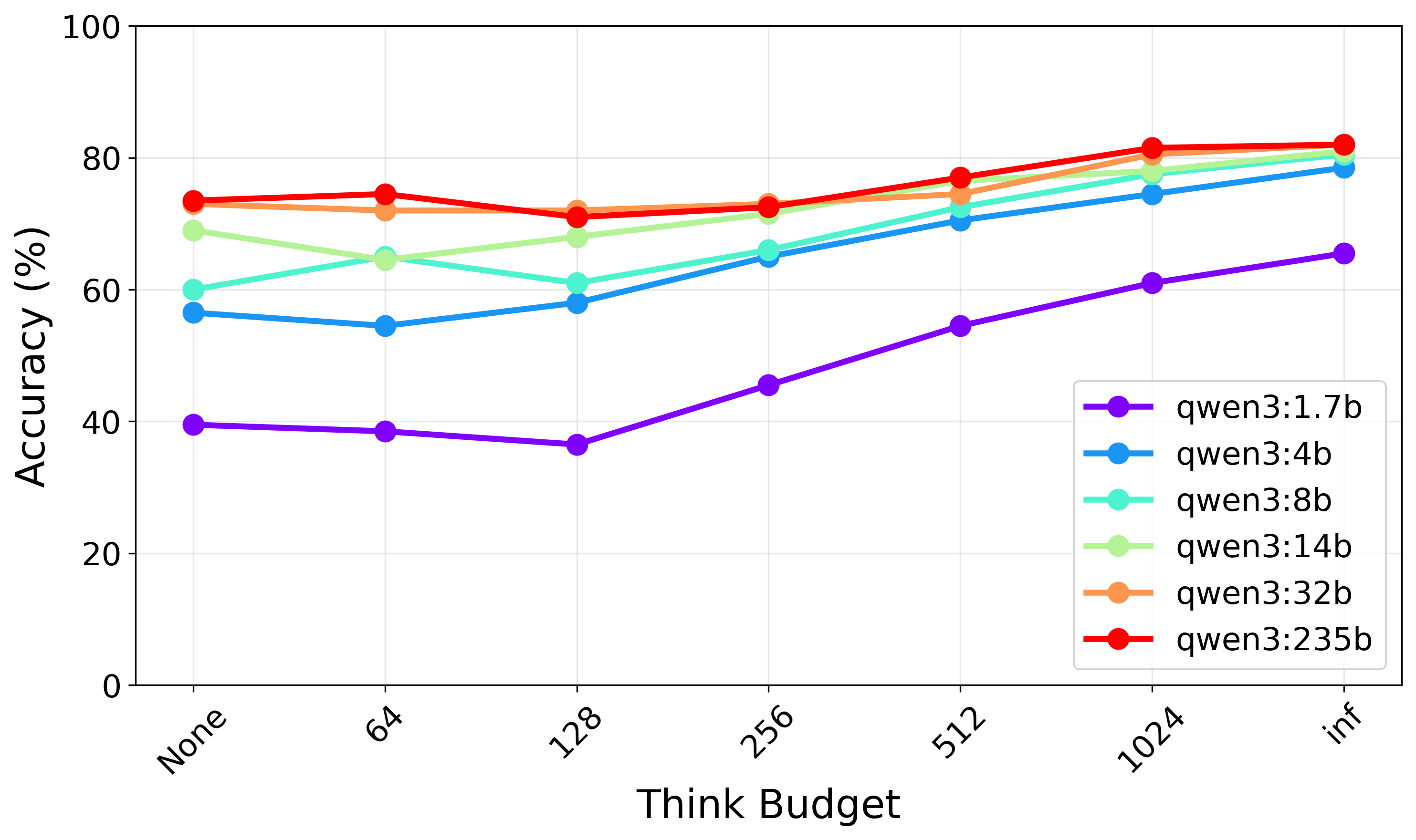}
\caption{Attending (Qwen3)}
\end{subfigure}
\hfill
\begin{subfigure}[b]{0.24\textwidth}
\centering
\includegraphics[width=\textwidth]{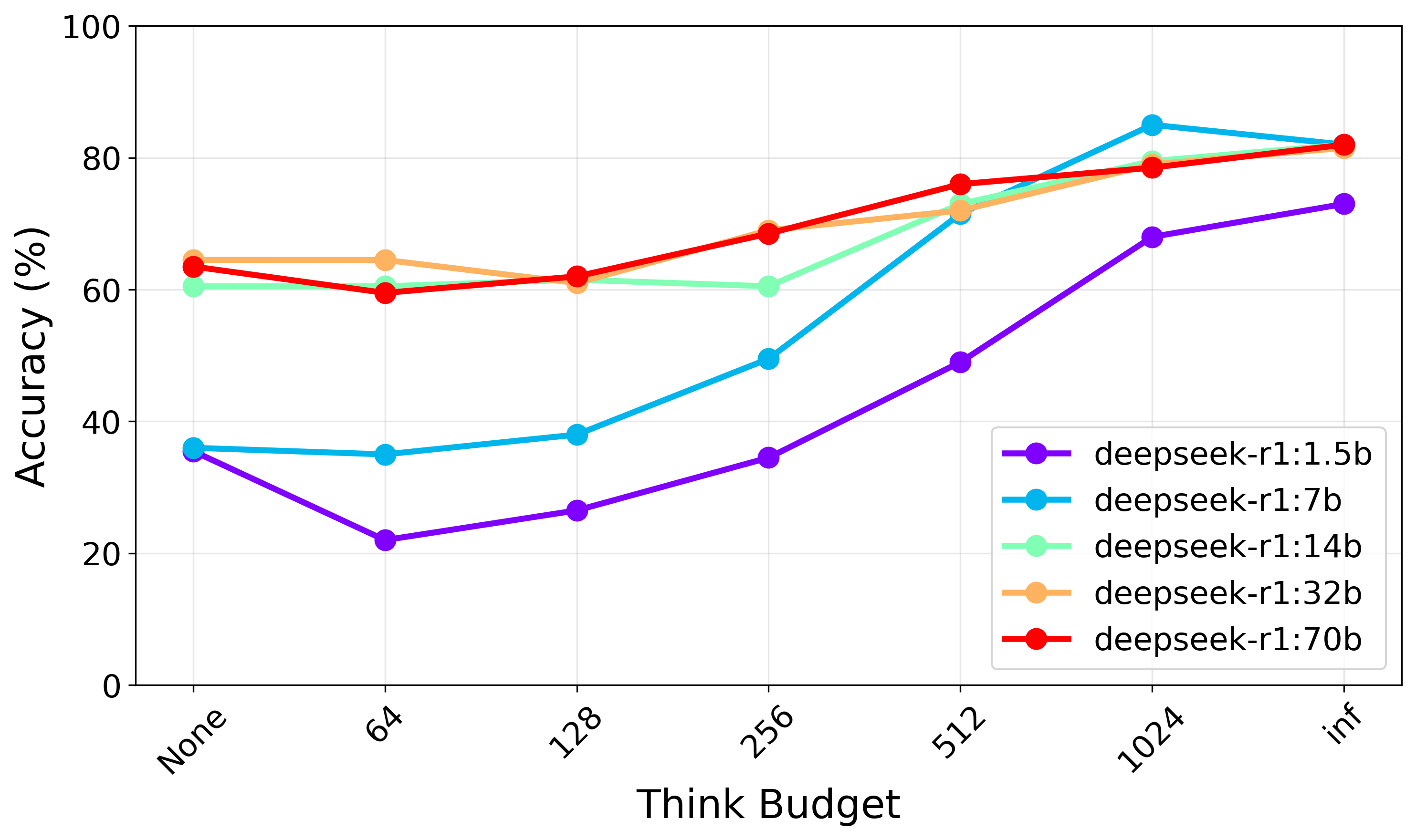}
\caption{Attending (DeepSeek-R1)}
\end{subfigure}
\hfill
\begin{subfigure}[b]{0.24\textwidth}
\centering
\includegraphics[width=\textwidth]{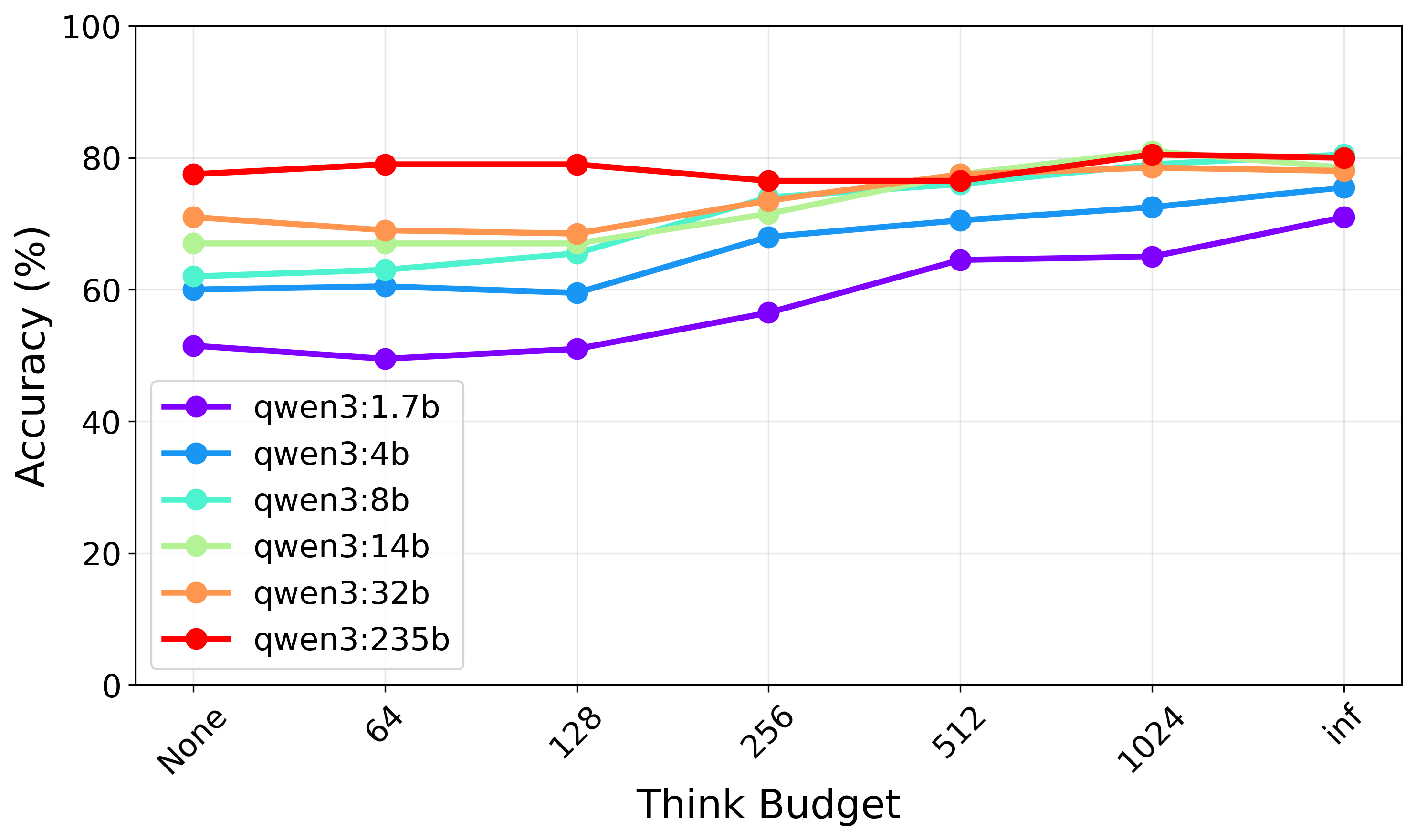}
\caption{Chief (Qwen3)}
\end{subfigure}
\hfill
\begin{subfigure}[b]{0.24\textwidth}
\centering
\includegraphics[width=\textwidth]{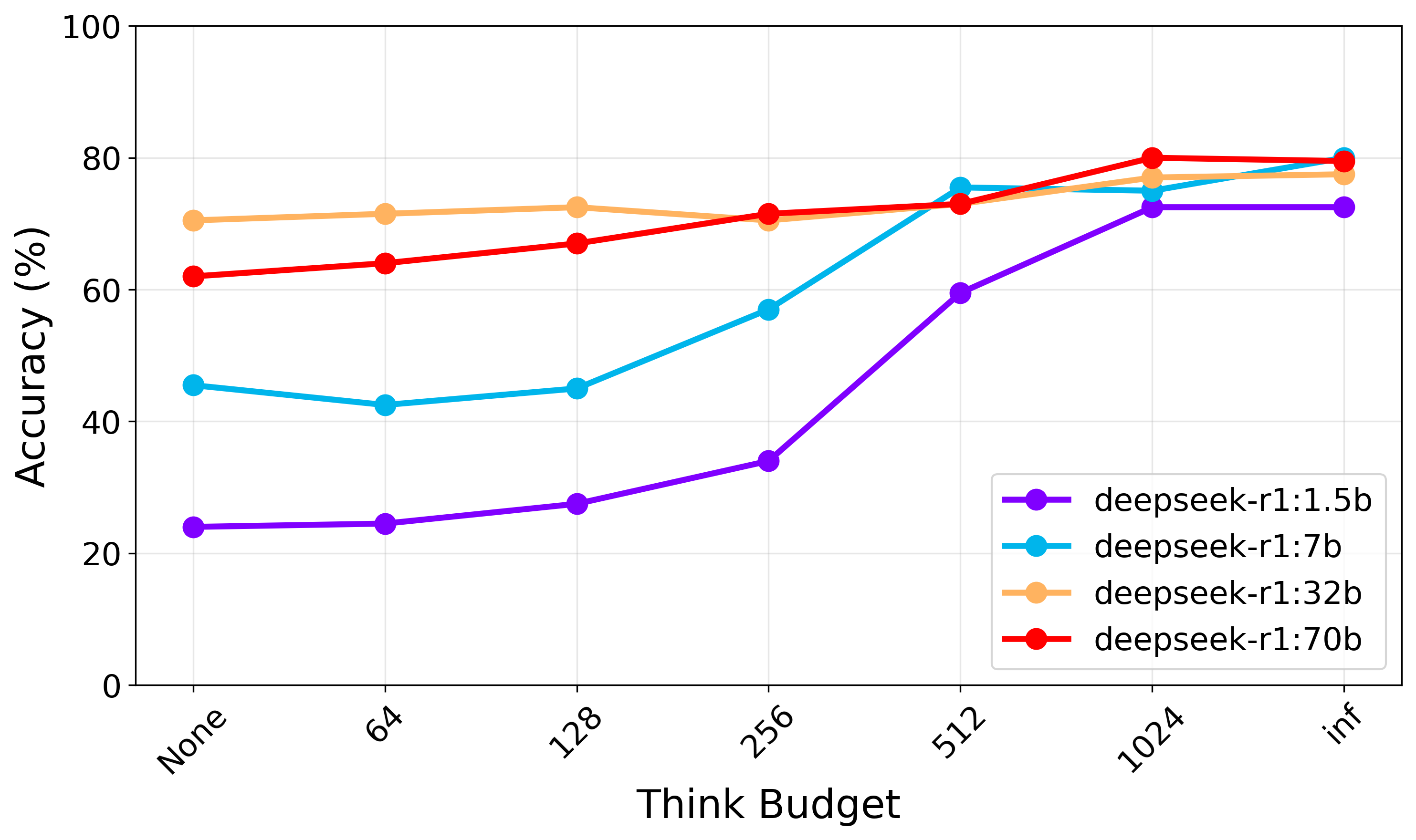}
\caption{Chief (DeepSeek-R1)}
\end{subfigure}

\vspace{0.5em}

\begin{subfigure}[b]{0.24\textwidth}
\centering
\includegraphics[width=\textwidth]{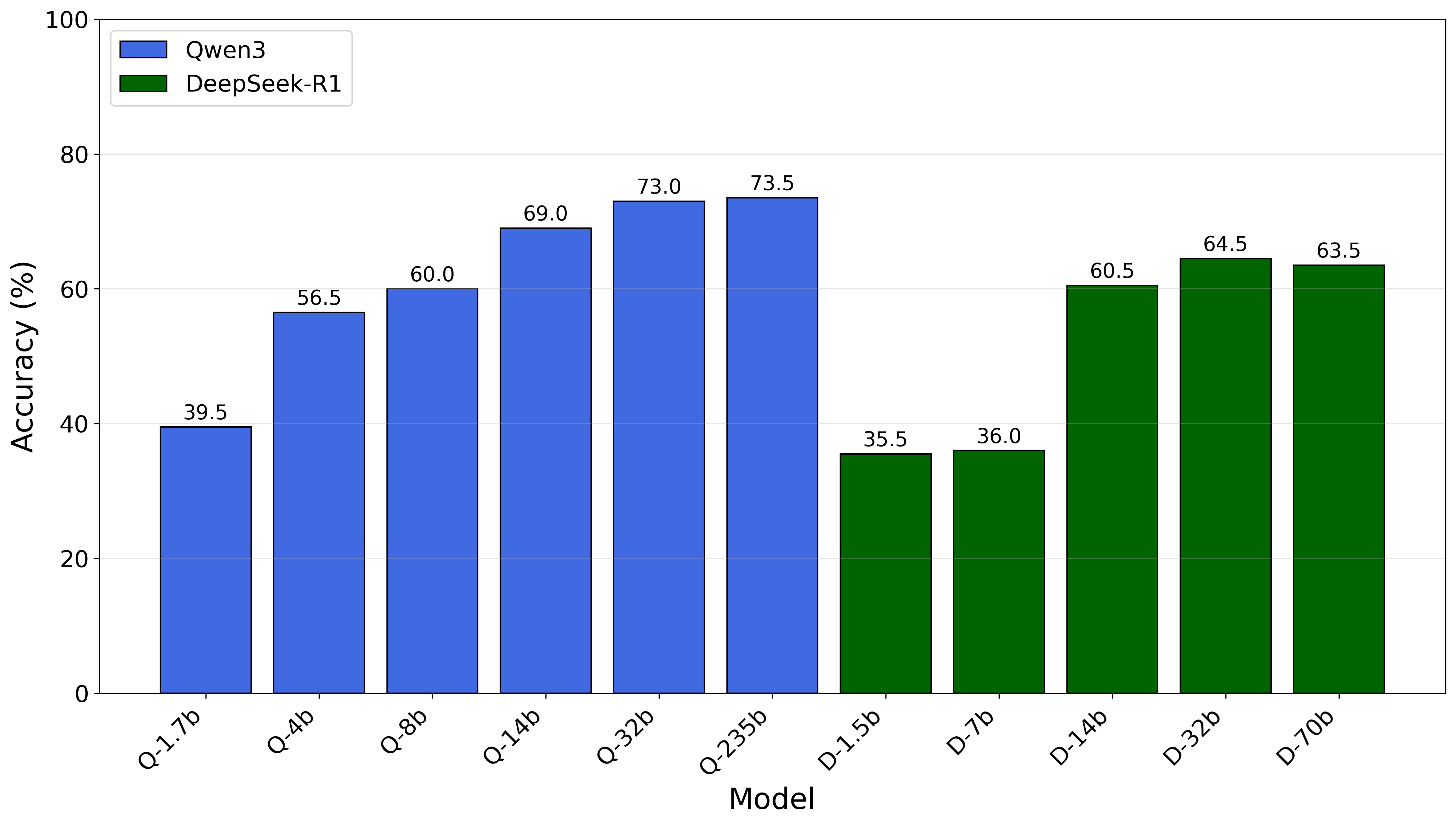}
\caption{Attending (None)}
\end{subfigure}
\hfill
\begin{subfigure}[b]{0.24\textwidth}
\centering
\includegraphics[width=\textwidth]{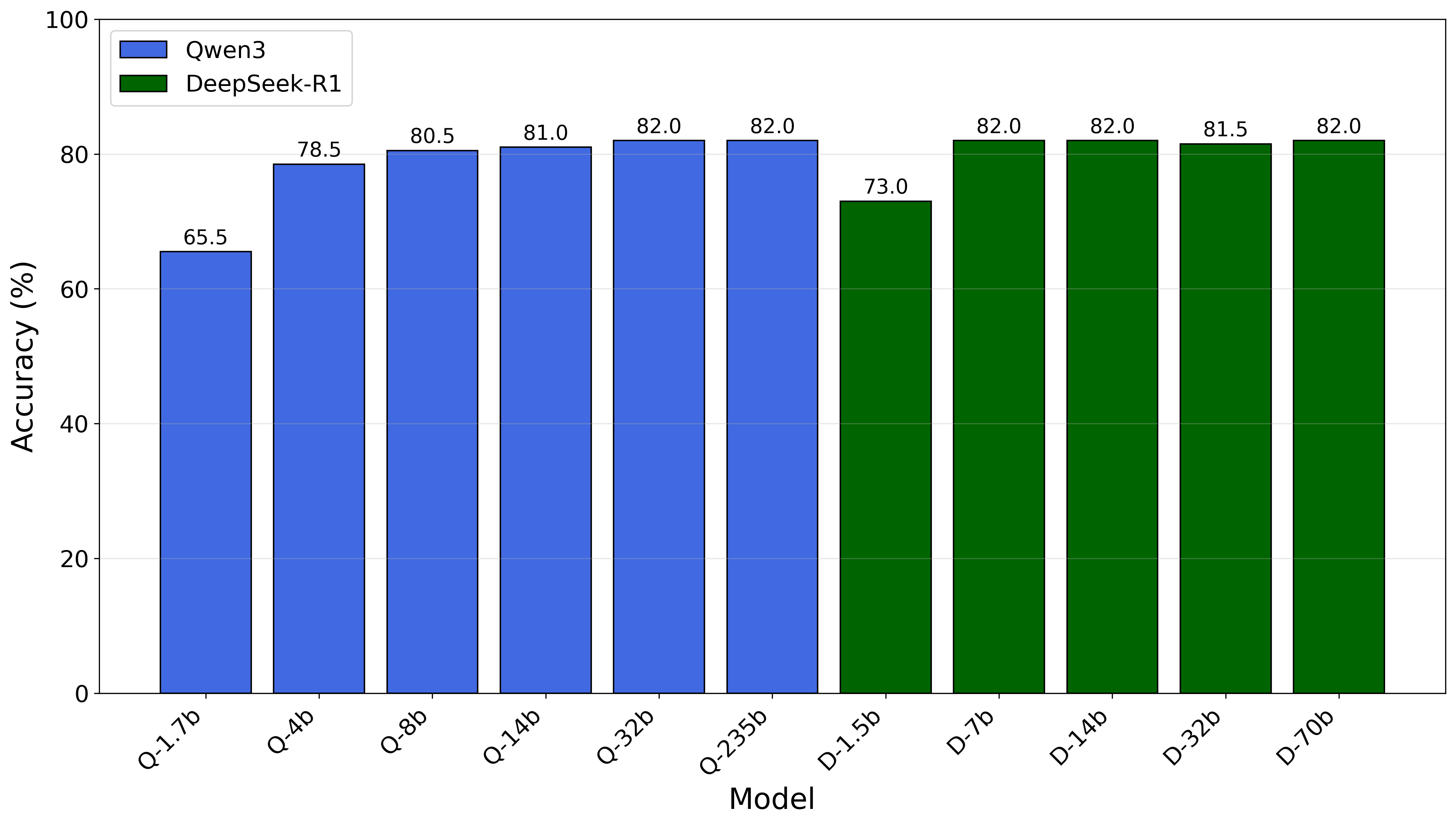}
\caption{Attending (inf)}
\end{subfigure}
\hfill
\begin{subfigure}[b]{0.24\textwidth}
\centering
\includegraphics[width=\textwidth]{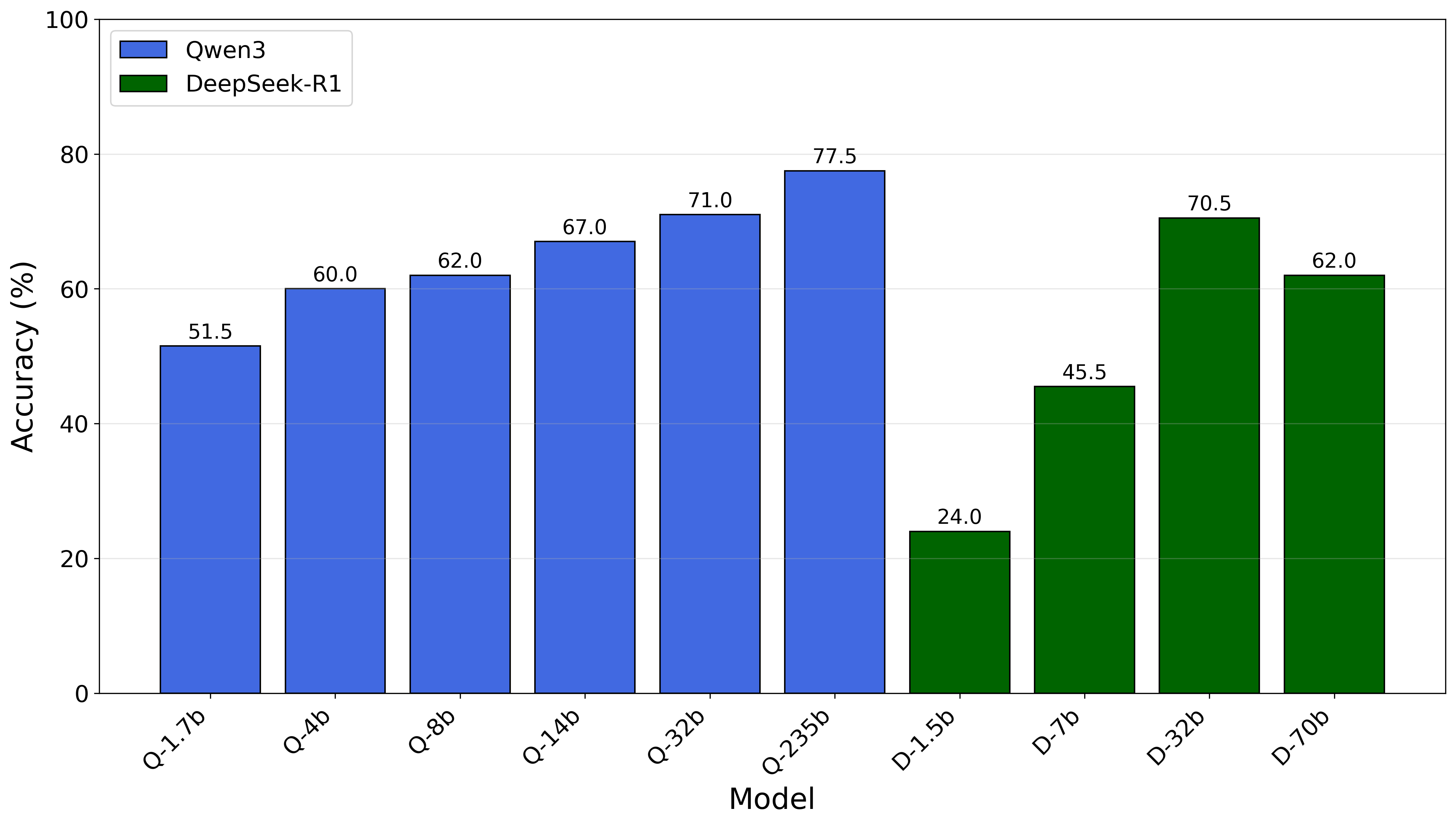}
\caption{Chief (None)}
\end{subfigure}
\hfill
\begin{subfigure}[b]{0.24\textwidth}
\centering
\includegraphics[width=\textwidth]{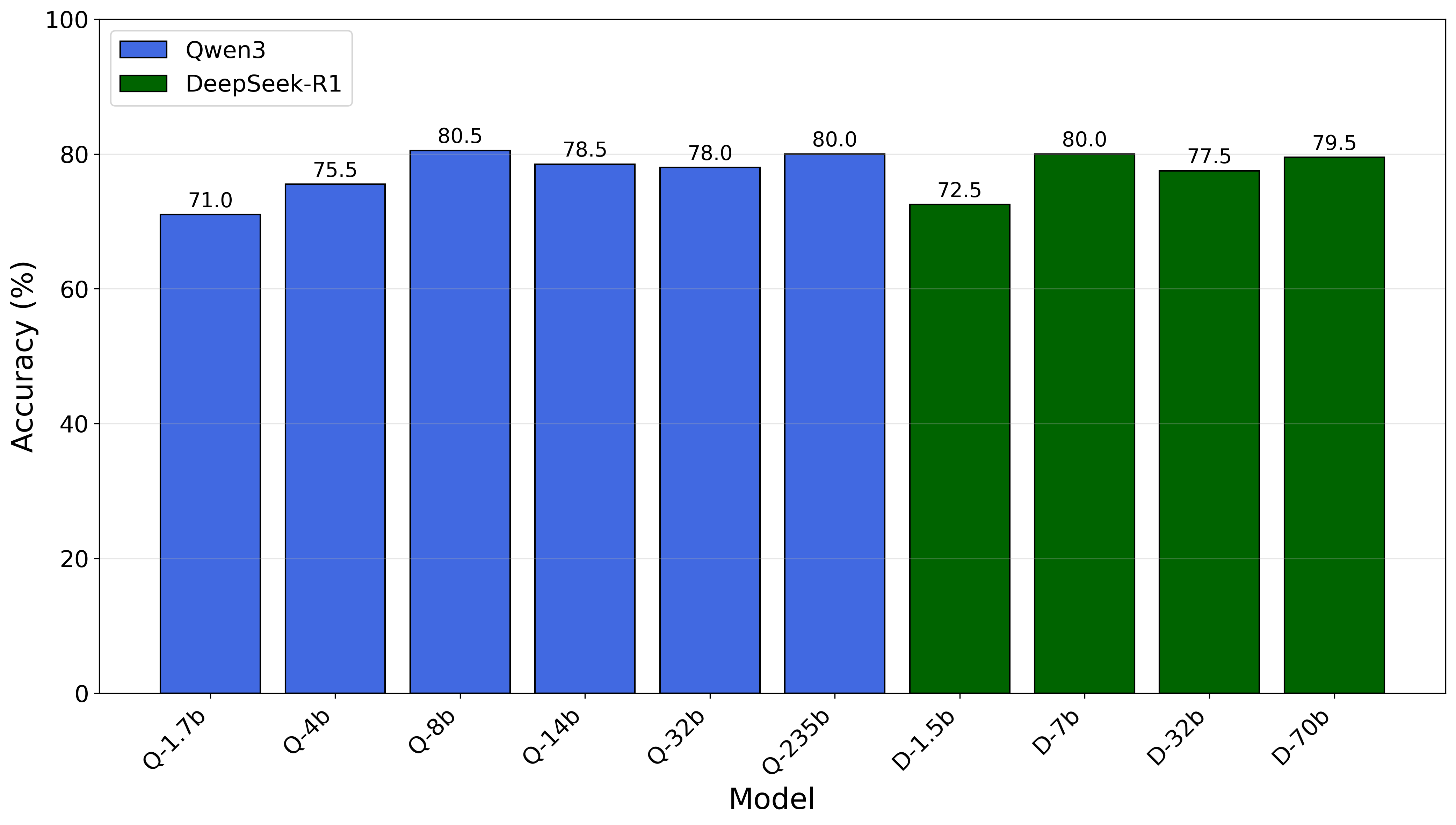}
\caption{Chief (inf)}
\end{subfigure}
\caption{Respiratory medicine performance across thinking budgets}
\label{fig:respiratory_complete}
\end{figure}

\subsection{AJKD}

\begin{figure}[H]
\centering
\begin{subfigure}[b]{0.24\textwidth}
\centering
\includegraphics[width=\textwidth]{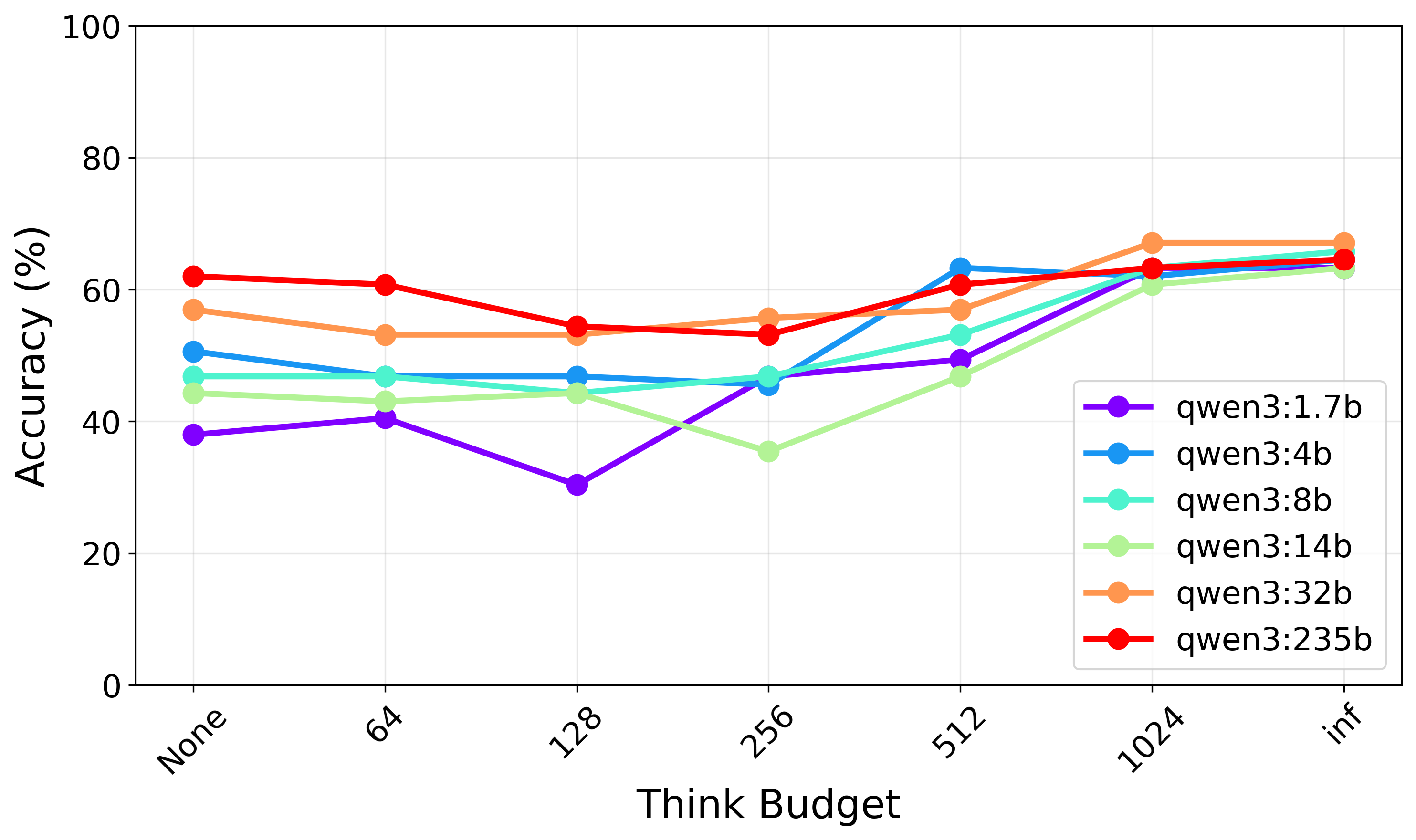}
\caption{AJKD (Qwen3)}
\end{subfigure}
\hfill
\begin{subfigure}[b]{0.24\textwidth}
\centering
\includegraphics[width=\textwidth]{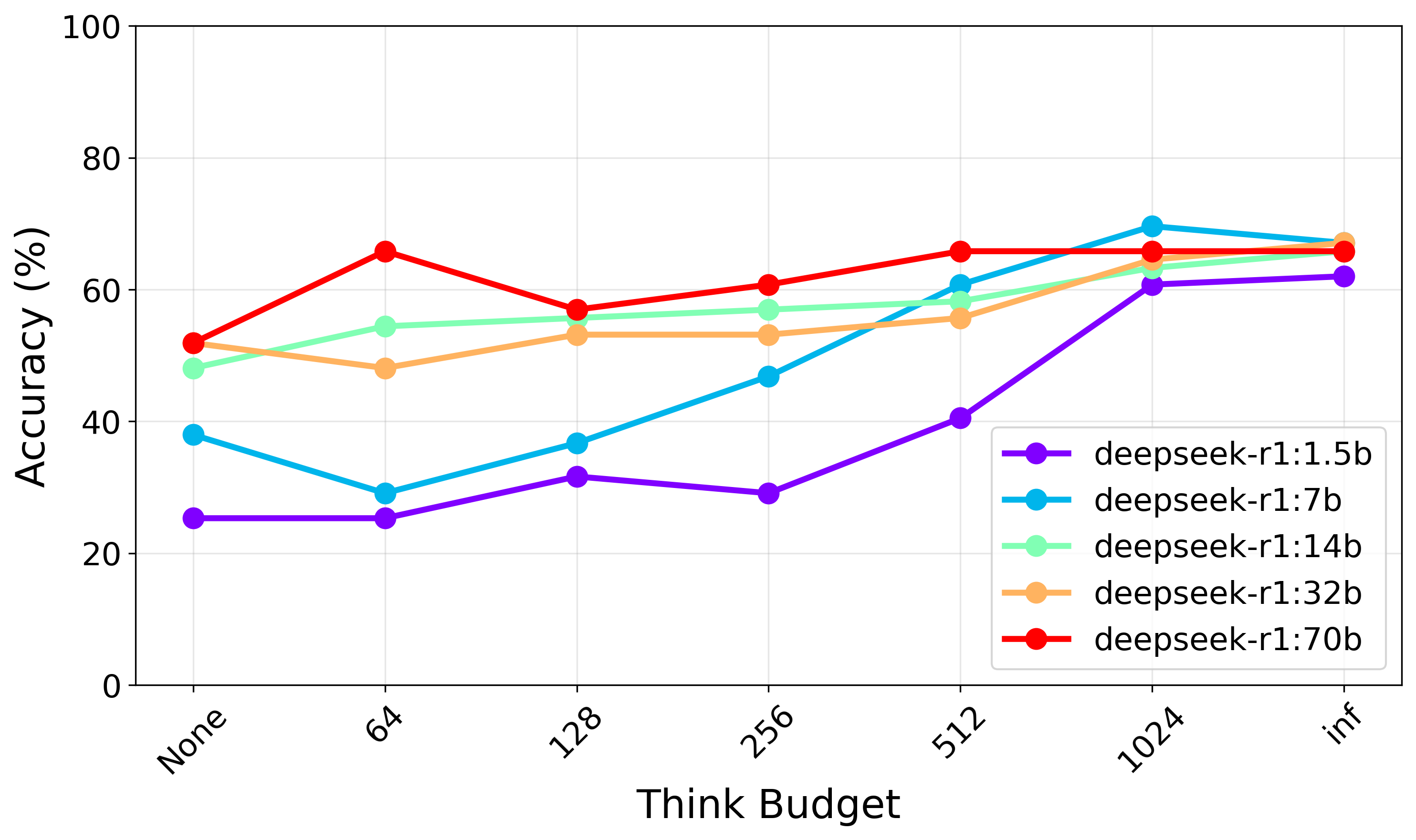}
\caption{AJKD (DeepSeek-R1)}
\end{subfigure}
\hfill
\begin{subfigure}[b]{0.24\textwidth}
\centering
\includegraphics[width=\textwidth]{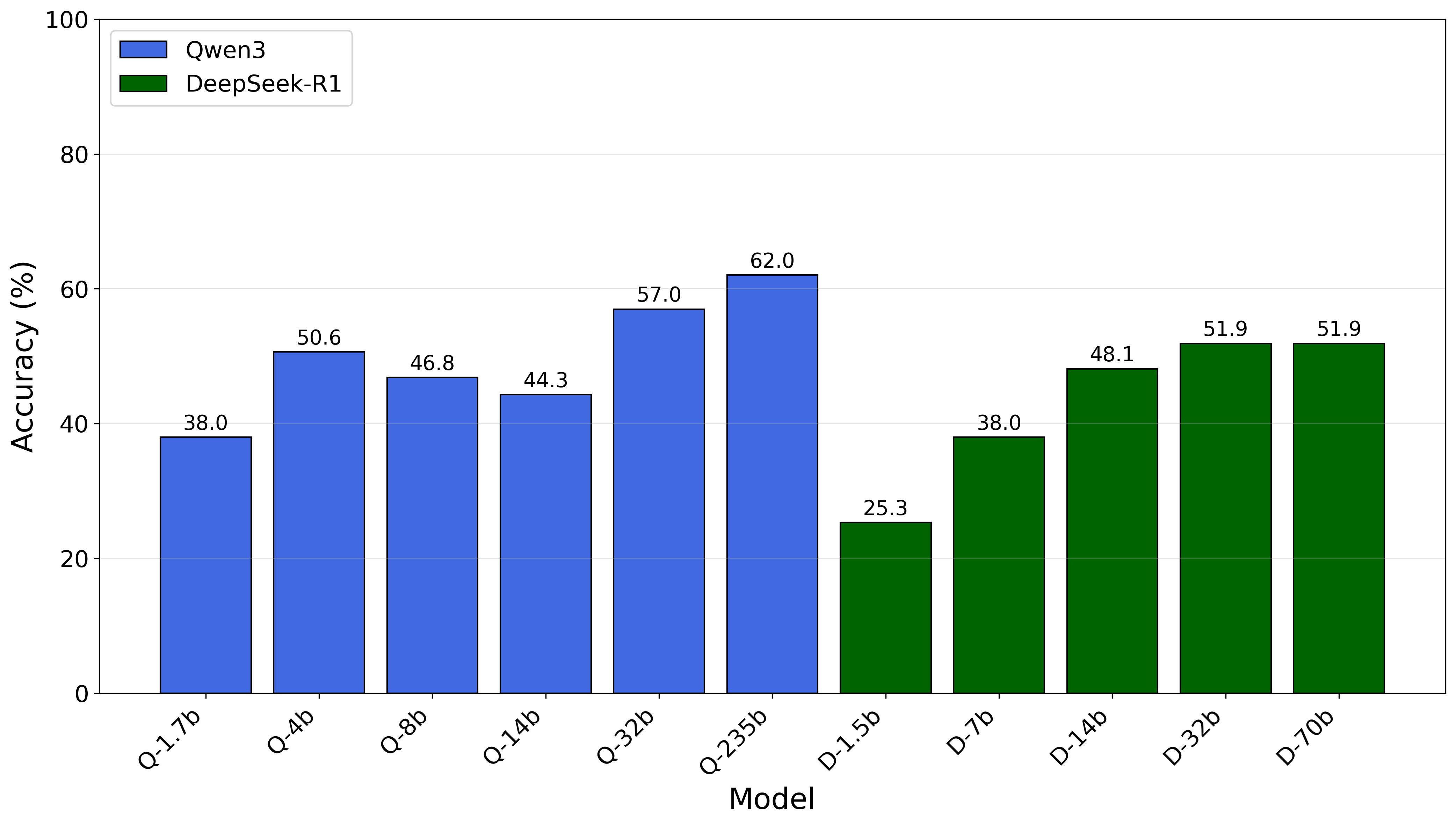}
\caption{AJKD (None)}
\end{subfigure}
\hfill
\begin{subfigure}[b]{0.24\textwidth}
\centering
\includegraphics[width=\textwidth]{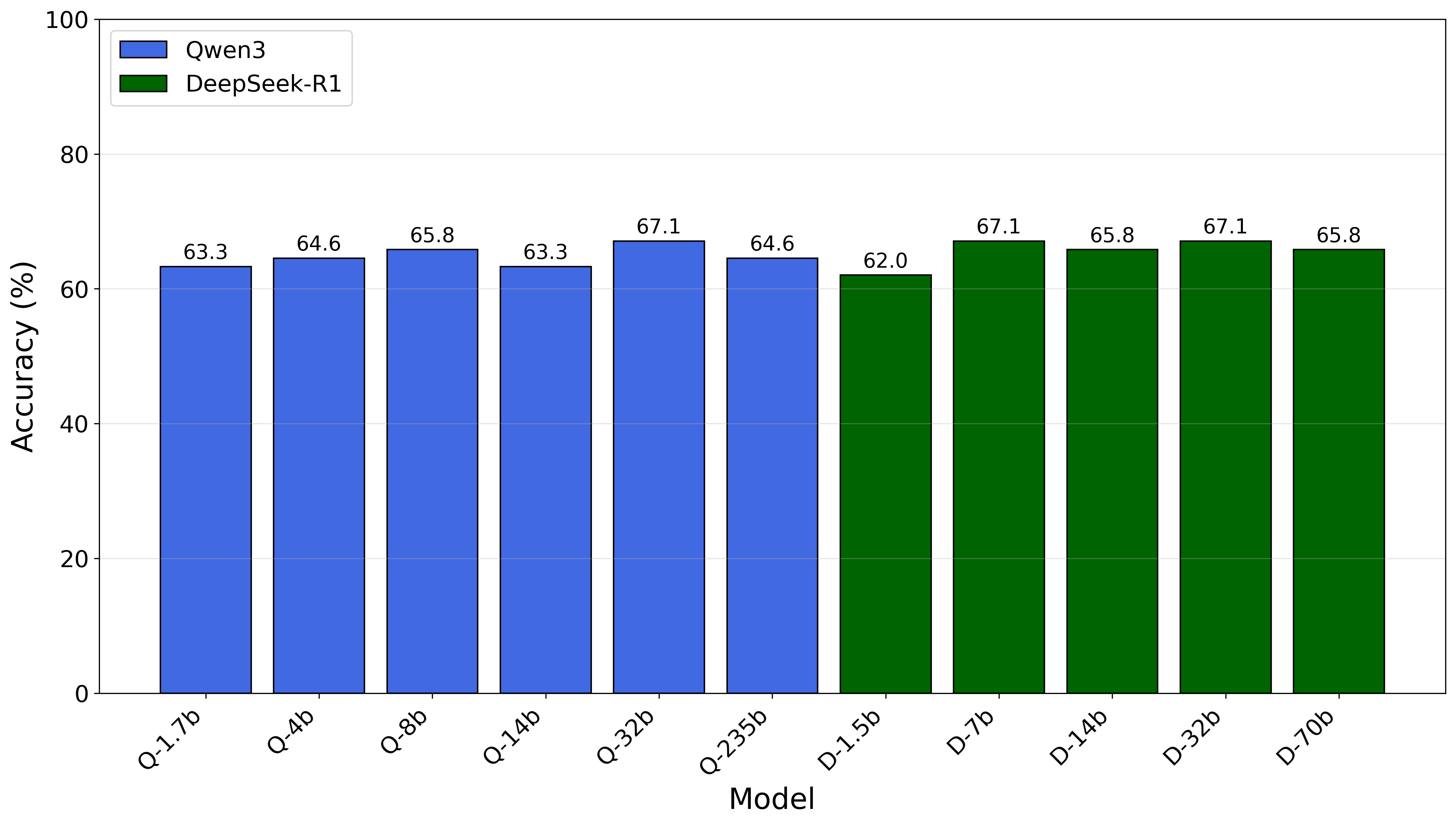}
\caption{AJKD (inf)}
\end{subfigure}
\caption{AJKD (American Journal of Kidney Diseases) performance across thinking budgets}
\label{fig:ajkd_complete}
\end{figure}

%% file: reference.bib
@article{wei2022chain,
    author  = "Jason Wei and Xuezhi Wang and Dale Schuurmans and Maarten Bosma and Ed Chi and Quoc Le and Denny Zhou",
    title   = "Chain-of-Thought Prompting Elicits Reasoning in Large Language Models",
    year    = "2022",
    journal = "Advances in Neural Information Processing Systems",
    volume  = "35",
    pages   = "24824--24837"
}

@article{yang2025qwen3,
  title={Qwen3 technical report},
  author={Yang, An and Li, Anfeng and Yang, Baosong and Zhang, Beichen and Hui, Binyuan and Zheng, Bo and Yu, Bowen and Gao, Chang and Huang, Chengen and Lv, Chenxu and others},
  journal={arXiv preprint arXiv:2505.09388},
  year={2025}
}

@misc{lai2025provable,
  title   = {Provably Robust Adaptation for Language-Empowered Foundation Models},
  author  = {Lai, Yuni and Xue, Xiaoyu and Shen, Linghui and Wu, Yulun and Li, Gaolei and Guo, Song and Zhou, Kai and Xiao, Bin},
  year    = {2025},
  eprint  = {2510.08659},
  archivePrefix = {arXiv},
  primaryClass = {cs.LG}
}

@misc{park2025medalignplus,
  title   = {MedAlign+: Improving Safety and Guideline Adherence of Medical Language Models},
  author  = {Park, Seongjun and Kim, Hyunwoo and Lee, Joon},
  year    = {2025},
  eprint  = {2504.06511},
  archivePrefix = {arXiv},
  primaryClass = {cs.AI}
}

@article{li2025steering,
  title={Steering LLM Thinking with Budget Guidance},
  author={Li, Junyan and Zhao, Wenshuo and Zhang, Yang and Gan, Chuang},
  journal={arXiv preprint arXiv:2506.13752},
  year={2025}
}

@article{salvi2024multi,
  title={Multi-modality approaches for medical support systems: A systematic review of the last decade},
  author={Salvi, Massimo and Loh, Hui Wen and Seoni, Silvia and Barua, Prabal Datta and Garc{\'\i}a, Salvador and Molinari, Filippo and Acharya, U Rajendra},
  journal={Information Fusion},
  volume={103},
  pages={102134},
  year={2024},
  publisher={Elsevier}
}

@article{salybekov2024artificial,
  title={Artificial Intelligence Reporting Guidelines’ adherence in nephrology for improved research and clinical outcomes},
  author={Salybekov, Amankeldi A and Wolfien, Markus and Hahn, Waldemar and Hidaka, Sumi and Kobayashi, Shuzo},
  journal={Biomedicines},
  volume={12},
  number={3},
  pages={606},
  year={2024},
  publisher={MDPI}
}

@book{shortliffe2012computer,
  title={Computer-based medical consultations: MYCIN},
  author={Shortliffe, Edward},
  volume={2},
  year={2012},
  publisher={Elsevier}
}

@article{schumacher2025rare,
  title={Rare Disease Differential Diagnosis with Large Language Models at Scale: From Abdominal Actinomycosis to Wilson's Disease},
  author={Schumacher, Elliot and Naik, Dhruv and Kannan, Anitha},
  journal={arXiv preprint arXiv:2502.15069},
  year={2025}
}

@article{su2025large,
  title={Large Language Models in Medical Diagnostics: Scoping Review With Bibliometric Analysis},
  author={Su, Hankun and Sun, Yuanyuan and Li, Ruiting and Zhang, Aozhe and Yang, Yuemeng and Xiao, Fen and Duan, Zhiying and Chen, Jingjing and Hu, Qin and Yang, Tianli and others},
  journal={Journal of Medical Internet Research},
  volume={27},
  pages={e72062},
  year={2025},
  publisher={JMIR Publications Toronto, Canada}
}

@article{soroush2024large,
  title={Large language models are poor medical coders—benchmarking of medical code querying},
  author={Soroush, Ali and Glicksberg, Benjamin S and Zimlichman, Eyal and Barash, Yiftach and Freeman, Robert and Charney, Alexander W and Nadkarni, Girish N and Klang, Eyal},
  journal={NEJM AI},
  volume={1},
  number={5},
  pages={AIdbp2300040},
  year={2024},
  publisher={Massachusetts Medical Society}
}

@article{kaplan2020scaling,
    author  = "Jared Kaplan and Sam McCandlish and Tom Henighan and Tom B. Brown and Benjamin Chess and Rewon Child and Scott Gray and Alec Radford and Jeffrey Wu and Dario Amodei",
    title   = "Scaling Laws for Neural Language Models",
    year    = "2020",
    journal = "arXiv preprint arXiv:2001.08361"
}

@article{hoffmann2022chinchilla,
    author  = "Jordan Hoffmann and Sebastian Borgeaud and Arthur Mensch and Elena Buchatskaya and Trevor Cai and Eliza Rutherford and Diego de Las Casas and Lisa Anne Hendricks and Johannes Welbl and Aidan Clark and others",
    title   = "Training Compute-Optimal Large Language Models",
    year    = "2022",
    journal = "Advances in Neural Information Processing Systems",
    volume  = "35",
    pages   = "30016--30030"
}

@inproceedings{han2015pruning,
    author  = "Song Han and Jeff Pool and John Tran and William Dally",
    title   = "Learning both Weights and Connections for Efficient Neural Network",
    year    = "2015",
    booktitle = "Advances in Neural Information Processing Systems",
    volume  = "28"
}

@article{jin2021medqa,
    author  = "Di Jin and Eileen Pan and Nassim Oufattole and Wei-Hung Weng and Hanyi Fang and Peter Szolovits",
    title   = "What Disease does this Patient Have? A Large-scale Open Domain Question Answering Dataset from Medical Exams",
    year    = "2021",
    journal = "Applied Sciences",
    volume  = "11",
    number  = "14",
    pages   = "6421"
}

@inproceedings{jin2019pubmedqa,
    author  = "Qiao Jin and Bhuwan Dhingra and Zhengping Liu and William W. Cohen and Xinghua Lu",
    title   = "{PubMedQA}: A Dataset for Biomedical Research Question Answering",
    year    = "2019",
    booktitle = "Proceedings of the 2019 Conference on Empirical Methods in Natural Language Processing",
    pages   = "2567--2577"
}

@article{graves2016adaptive,
    author  = "Alex Graves",
    title   = "Adaptive Computation Time for Recurrent Neural Networks",
    year    = "2016",
    journal = "arXiv preprint arXiv:1603.08983"
}

@inproceedings{vaswani2017attention,
    author  = "Ashish Vaswani and Noam Shazeer and Niki Parmar and Jakob Uszkoreit and Llion Jones and Aidan N. Gomez and Lukasz Kaiser and Illia Polosukhin",
    title   = "Attention is All You Need",
    year    = "2017",
    booktitle = "Advances in Neural Information Processing Systems",
    volume  = "30"
}

@inproceedings{sundararajan2017integrated,
    author  = "Mukund Sundararajan and Ankur Taly and Qiqi Yan",
    title   = "Axiomatic Attribution for Deep Networks",
    year    = "2017",
    booktitle = "Proceedings of the 34th International Conference on Machine Learning",
    pages   = "3319--3328"
}

@inproceedings{kim2018tcav,
    author  = "Been Kim and Martin Wattenberg and Justin Gilmer and Carrie Cai and James Wexler and Fernanda Viegas and Rory Sayres",
    title   = "Interpretability Beyond Feature Attribution: Quantitative Testing with Concept Activation Vectors ({TCAV})",
    year    = "2018",
    booktitle = "Proceedings of the 35th International Conference on Machine Learning",
    pages   = "2668--2677"
}

@article{brown2020gpt3,
    author  = "Tom B. Brown and Benjamin Mann and Nick Ryder and Melanie Subbiah and Jared Kaplan and Prafulla Dhariwal and Arvind Neelakantan and Pranav Shyam and Girish Sastry and Amanda Askell and others",
    title   = "Language Models are Few-Shot Learners",
    year    = "2020",
    journal = "Advances in Neural Information Processing Systems",
    volume  = "33",
    pages   = "1877--1901"
}

@article{openai2023gpt4,
    author  = "{OpenAI}",
    title   = "{GPT-4} Technical Report",
    year    = "2023",
    journal = "arXiv preprint arXiv:2303.08774"
}

@article{alain2025evaluating,
  title={Evaluating User Interactions and Adoption Patterns of Generative AI in Health Care Occupations Using Claude: Cross-Sectional Study},
  author={Alain, Gabriel and Crick, James and Snead, Ella and Quatman-Yates, Catherine C and Quatman, Carmen E},
  journal={Journal of Medical Internet Research},
  volume={27},
  pages={e73918},
  year={2025},
  publisher={JMIR Publications Toronto, Canada}
}

@article{bai2023qwen,
    author  = "Jinze Bai and Shuai Bai and Yunfei Chu and Zeyu Cui and Kai Dang and Xiaodong Deng and Yang Fan and Wenbin Ge and Yu Han and Fei Huang and others",
    title   = "Qwen Technical Report",
    year    = "2023",
    journal = "arXiv preprint arXiv:2309.16609"
}

@article{liu2025deepseek,
    author  = "DeepSeek AI",
    title   = "{DeepSeek-R1}: Incentivizing Reasoning Capability in {LLMs} via Reinforcement Learning",
    year    = "2025",
    journal = "arXiv preprint arXiv:2501.12948"
}

@article{markowitz1952portfolio,
    author  = "Harry Markowitz",
    title   = "Portfolio Selection",
    year    = "1952",
    journal = "The Journal of Finance",
    volume  = "7",
    number  = "1",
    pages   = "77--91"
}

@book{markowitz1959portfolio,
    author  = "Harry Markowitz",
    title   = "Portfolio Selection: Efficient Diversification of Investments",
    year    = "1959",
    publisher = "John Wiley \& Sons",
    address = "New York"
}

@article{kung2023performance,
    author  = "Tiffany H. Kung and Morgan Cheatham and Arielle Medenilla and Czarina Sillos and Lorie De Leon and Camille Elepaño and Maria Madriaga and Rimel Aggabao and Giezel Diaz-Candido and James Maningo and Victor Tseng",
    title   = "Performance of {ChatGPT} on {USMLE}: Potential for {AI}-assisted medical education using large language models",
    year    = "2023",
    journal = "PLOS Digital Health",
    volume  = "2",
    number  = "2",
    pages   = "e0000198"
}

@article{chen2023medprompt,
    author  = "Harsha Nori and Yin Tat Lee and Sheng Zhang and Dean Carignan and Richard Edgar and Nicolo Fusi and Nicholas King and Jonathan Larson and Yuanzhi Li and Weishung Liu and others",
    title   = "Can Generalist Foundation Models Outcompete Special-Purpose Tuning? Case Study in Medicine",
    year    = "2023",
    journal = "arXiv preprint arXiv:2311.16452"
}

@article{zhang2024medqa,
    author  = "Sheng Zhang and Yanbo Xu and Naoto Usuyama and Hanwen Xu and Jaspreet Bagga and Robert Tinn and Sam Preston and Rajesh Rao and Mu Wei and Naveen Valluri and others",
    title   = "{MedQA-CS}: Benchmarking Large Language Models Clinical Skills Using an {AI-SCE} Framework",
    year    = "2024",
    journal = "arXiv preprint arXiv:2410.01553"
}

@article{abacha2024review,
    author  = "Zahra Ranjbar and Faezeh Tabatabaei and Mahdi Goudarzi and Assef Zare",
    title   = "A review of Explainable Artificial Intelligence in healthcare",
    year    = "2024",
    journal = "Computers and Electrical Engineering",
    volume  = "119",
    pages   = "109358"
}

@article{loh2023application,
    author  = "Huey Yi Loh and Boon Pheng Loh and Qi Hong and Raymond Chiong Chio Bai",
    title   = "Application of explainable artificial intelligence in medical health: A systematic review of interpretability methods",
    year    = "2023",
    journal = "Informatics in Medicine Unlocked",
    volume  = "40",
    pages   = "101286"
}

@article{alsaleh2024enhancing,
    author  = "Norah Alsaleh and Nourah Alangari and Abdulaziz Alharbi and Zainab Alshayeb and Heba Kurdi",
    title   = "Enhancing interpretability and accuracy of {AI} models in healthcare: a comprehensive review on challenges and future directions",
    year    = "2024",
    journal = "Frontiers in Robotics and AI",
    volume  = "11",
    pages   = "1444763"
}

@article{chen2024explainable,
    author  = "Haowei Chen and Zhihua Wang and Jieming Ma and Xiaomei Ou and Yan He",
    title   = "Explainable and interpretable artificial intelligence in medicine: a systematic bibliometric review",
    year    = "2024",
    journal = "Discover Artificial Intelligence",
    volume  = "4",
    pages   = "15"
}

@article{farah2024promise,
    author  = "Lara Farah and Jianing Dong and Jacob A. Buckley and Aubrey J. Rossi and Rui Ma and Hongbing Ma",
    title   = "The Promise of Explainable {AI} in Digital Health for Precision Medicine: A Systematic Review",
    year    = "2024",
    journal = "Journal of Personalized Medicine",
    volume  = "14",
    number  = "3",
    pages   = "277"
}

@article{pfeifer2024how,
    author  = "Bernd Pfeifer and Katharina Kumpfmüller and Antonio Parodi and Martin Dietrich and Heimo Müller and Bastian Pfeifer and Karl Roprecht and Andreas Holzinger",
    title   = "How Explainable Artificial Intelligence Can Increase or Decrease Clinicians' Trust in {AI} Applications in Health Care: Systematic Review",
    year    = "2024",
    journal = "JMIR AI",
    volume  = "3",
    pages   = "e53207"
}

@article{kojima2022large,
    author  = "Takeshi Kojima and Shixiang Shane Gu and Machel Reid and Yutaka Matsuo and Yusuke Iwasawa",
    title   = "Large Language Models are Zero-Shot Reasoners",
    year    = "2022",
    journal = "Advances in Neural Information Processing Systems",
    volume  = "35",
    pages   = "22199--22213"
}

@article{fu2023complexitybased,
    author  = "Yao Fu and Hao Peng and Ashish Sabharwal and Peter Clark and Tushar Khot",
    title   = "Complexity-Based Prompting for Multi-Step Reasoning",
    year    = "2023",
    journal = "Proceedings of the International Conference on Learning Representations"
}

@article{lightman2023lets,
    author  = "Hunter Lightman and Vineet Kosaraju and Yura Burda and Harri Edwards and Bowen Baker and Teddy Lee and Jan Leike and John Schulman and Ilya Sutskever and Karl Cobbe",
    title   = "Let's Verify Step by Step",
    year    = "2023",
    journal = "arXiv preprint arXiv:2305.20050"
}

@article{mcduff2023towards,
    author  = "Daniel McDuff and Mike Schaekermann and Tao Tu and Anil Palepu and Amy Wang and Jake Garrison and Karan Singhal and Yash Sharma and Shekoofeh Azizi and Kavita Kulkarni and others",
    title   = "Towards Accurate Differential Diagnosis with Large Language Models",
    year    = "2023",
    journal = "arXiv preprint arXiv:2312.00164"
}

@article{saab2024capabilities,
    author  = "Khaled Saab and Tao Tu and Wei-Hung Weng and Ryutaro Tanno and David Stutz and Ellery Wulczyn and Fan Zhang and Tim Strother and Chunjong Park and Elahe Vedadi and others",
    title   = "Capabilities of {Gemini} Models in Medicine",
    year    = "2024",
    journal = "arXiv preprint arXiv:2404.18416"
}

@article{moor2023medflamingo,
    author  = "Michael Moor and Qian Huang and Shirley Wu and Michihiro Yasunaga and Cyril Zakka and Yash Dalmia and Eduardo Pontes Reis and Pranav Rajpurkar and Jure Leskovec",
    title   = "{Med-Flamingo}: a Multimodal Medical Few-shot Learner",
    year    = "2023",
    journal = "Proceedings of Machine Learning for Healthcare Conference",
    pages   = "353--367"
}

@article{chen2024huatuogpt,
    author  = "Jianquan Zhang and Xidong Wang and Xiangbo Wu and Tianyu Bai and Ying Chen and Xingjian Wang and Ruifei Zhang and Xiaohong Liu and Chunlin Li and Xiang Wan and others",
    title   = "{HuatuoGPT-II}, One-stage Training for Medical Adaption of {LLMs}",
    year    = "2024",
    journal = "arXiv preprint arXiv:2311.09774"
}

@article{lee2023benefits,
    author  = "Peter Lee and Sebastien Bubeck and Joseph Petro",
    title   = "Benefits, Limits, and Risks of {GPT-4} as an {AI} Chatbot for Medicine",
    year    = "2023",
    journal = "New England Journal of Medicine",
    volume  = "388",
    number  = "13",
    pages   = "1233--1239"
}

@article{eriksen2023use,
    author  = "Anne V. Eriksen and Stephan Möller and Jesper Ryg",
    title   = "Use of {GPT-4} to Diagnose Complex Clinical Cases",
    year    = "2023",
    journal = "NEJM AI",
    volume  = "1",
    number  = "1",
    pages   = "AIp2300031"
}

@article{brin2023comparing,
    author  = "Dana Brin and Vera Sorin and Akhil Vaid and Ali Soroush and Benjamin S. Glicksberg and Alexander W. Charney and Girish Nadkarni and Eyal Klang",
    title   = "Comparing {ChatGPT} and {GPT-4} performance in {USMLE} soft skill assessments",
    year    = "2023",
    journal = "Scientific Reports",
    volume  = "13",
    pages   = "16492"
}

@article{ji2023survey,
    author  = "Ziwei Ji and Nayeon Lee and Rita Frieske and Tiezheng Yu and Dan Su and Yan Xu and Etsuko Ishii and Ye Jin Bang and Andrea Madotto and Pascale Fung",
    title   = "Survey of Hallucination in Natural Language Generation",
    year    = "2023",
    journal = "ACM Computing Surveys",
    volume  = "55",
    number  = "12",
    pages   = "1--38"
}

@article{yao2023tree,
    author  = "Shunyu Yao and Dian Yu and Jeffrey Zhao and Izhak Shafran and Thomas L. Griffiths and Yuan Cao and Karthik Narasimhan",
    title   = "Tree of Thoughts: Deliberate Problem Solving with Large Language Models",
    year    = "2023",
    journal = "Advances in Neural Information Processing Systems",
    volume  = "36"
}

@article{madaan2023selfrefine,
    author  = "Aman Madaan and Niket Tandon and Prakhar Gupta and Skyler Hallinan and Luyu Gao and Sarah Wiegreffe and Uri Alon and Nouha Dziri and Shrimai Prabhumoye and Yiming Yang and others",
    title   = "Self-Refine: Iterative Refinement with Self-Feedback",
    year    = "2023",
    journal = "Advances in Neural Information Processing Systems",
    volume  = "36"
}

@article{dhuliawala2023chain,
    author  = "Shehzaad Dhuliawala and Mojtaba Komeili and Jing Xu and Roberta Raileanu and Xian Li and Asli Celikyilmaz and Jason Weston",
    title   = "Chain-of-Verification Reduces Hallucination in Large Language Models",
    year    = "2023",
    journal = "arXiv preprint arXiv:2309.11495"
}

@article{lewis2020retrieval,
    author  = "Patrick Lewis and Ethan Perez and Aleksandra Piktus and Fabio Petroni and Vladimir Karpukhin and Naman Goyal and Heinrich Küttler and Mike Lewis and Wen-tau Yih and Tim Rocktäschel and others",
    title   = "Retrieval-Augmented Generation for Knowledge-Intensive {NLP} Tasks",
    year    = "2020",
    journal = "Advances in Neural Information Processing Systems",
    volume  = "33",
    pages   = "9459--9474"
}

@article{zakka2024almanac,
    author  = "Cyril Zakka and Akash Chaurasia and Rohan Shad and Alex R. Dalal and Jennifer L. Kim and Michael Moor and Kevin Alexander and Euan Ashley and Jack Boyd and Kathleen Boyd and others",
    title   = "{ALMANAC}: Retrieval-Augmented Language Models for Clinical Medicine",
    year    = "2024",
    journal = "NEJM AI",
    volume  = "1",
    number  = "2",
    pages   = "AIoa2300068"
}

@article{xiong2024benchmarking,
    author  = "Guangzhi Xiong and Qiao Jin and Zhiyong Lu and Aidong Zhang",
    title   = "Benchmarking Retrieval-Augmented Generation for Medicine",
    year    = "2024",
    journal = "arXiv preprint arXiv:2402.13178"
}

@article{borgeaud2022improving,
    author  = "Sebastian Borgeaud and Arthur Mensch and Jordan Hoffmann and Trevor Cai and Eliza Rutherford and Katie Millican and George Bm Van Den Driessche and Jean-Baptiste Lespiau and Bogdan Damoc and Aidan Clark and others",
    title   = "Improving language models by retrieving from trillions of tokens",
    year    = "2022",
    journal = "Proceedings of the 39th International Conference on Machine Learning",
    pages   = "2206--2240"
}

@article{wu2024pmc,
    author  = "Chaoyi Wu and Xiaoman Zhang and Ya Zhang and Yanfeng Wang and Weidi Xie",
    title   = "{PMC-LLaMA}: Towards Building Open-source Language Models for Medicine",
    year    = "2024",
    journal = "Journal of the American Medical Informatics Association",
    volume  = "31",
    number  = "9",
    pages   = "1833--1843"
}

@article{chen2023meditron,
    author  = "Zeming Chen and Alejandro Hernández-Cano and Angelika Romanou and Antoine Bonnet and Kyle Matoba and Francesco Salvi and Matteo Pagliardini and Simin Fan and Andreas Köpf and Amirkeivan Mohtashami and others",
    title   = "{MEDITRON-70B}: Scaling Medical Pretraining for Large Language Models",
    year    = "2023",
    journal = "arXiv preprint arXiv:2311.16079"
}

@article{labrak2024biomistral,
    author  = "Yanis Labrak and Adrien Bazoge and Emmanuel Morin and Pierre-Antoine Gourraud and Mickael Rouvier and Richard Dufour",
    title   = "{BioMistral}: A Collection of Open-Source Pretrained Large Language Models for Medical Domains",
    year    = "2024",
    journal = "arXiv preprint arXiv:2402.10373"
}

@article{toma2023clinical,
    author  = "Augustin Toma and Patrick R. Lawler and Jimmy Ba and Rahul G. Krishnan and Barry B. Rubin and Bo Wang",
    title   = "Clinical Camel: An Open Expert-Level Medical Language Model with Dialogue-Based Knowledge Encoding",
    year    = "2023",
    journal = "arXiv preprint arXiv:2305.12031"
}

@article{han2023medalpaca,
    author  = "Tianyu Han and Lisa C. Adams and Jens-Michalis Papaioannou and Paul Grundmann and Tom Oberhauser and Alexander Löser and Daniel Truhn and Keno K. Bressem",
    title   = "{MedAlpaca} -- An Open-Source Collection of Medical Conversational {AI} Models and Training Data",
    year    = "2023",
    journal = "arXiv preprint arXiv:2304.08247"
}

@article{zhang2025igniting,
  title={Igniting language intelligence: The hitchhiker’s guide from chain-of-thought reasoning to language agents},
  author={Zhang, Zhuosheng and Yao, Yao and Zhang, Aston and Tang, Xiangru and Ma, Xinbei and He, Zhiwei and Wang, Yiming and Gerstein, Mark and Wang, Rui and Liu, Gongshen and others},
  journal={ACM Computing Surveys},
  volume={57},
  number={8},
  pages={1--39},
  year={2025},
  publisher={ACM New York, NY}
}

@article{gschwind2025ai,
  title={AI: It's All About Inference Now: Model inference has become the critical driver for model performance.},
  author={Gschwind, Michael},
  journal={Queue},
  volume={23},
  number={2},
  pages={40--78},
  year={2025},
  publisher={ACM New York, NY, USA}
}

@article{singhal2025toward,
  title={Toward expert-level medical question answering with large language models},
  author={Singhal, Karan and Tu, Tao and Gottweis, Juraj and Sayres, Rory and Wulczyn, Ellery and Amin, Mohamed and Hou, Le and Clark, Kevin and Pfohl, Stephen R and Cole-Lewis, Heather and others},
  journal={Nature Medicine},
  pages={1--8},
  year={2025},
  publisher={Nature Publishing Group US New York}
}

@article{yang2022large,
  title={A large language model for electronic health records},
  author={Yang, Xi and Chen, Aokun and PourNejatian, Nima and Shin, Hoo Chang and Smith, Kaleb E and Parisien, Christopher and Compas, Colin and Martin, Cheryl and Costa, Anthony B and Flores, Mona G and others},
  journal={NPJ digital medicine},
  volume={5},
  number={1},
  pages={194},
  year={2022},
  publisher={Nature Publishing Group UK London}
}

@article{tu2025towards,
  title={Towards conversational diagnostic artificial intelligence},
  author={Tu, Tao and Schaekermann, Mike and Palepu, Anil and Saab, Khaled and Freyberg, Jan and Tanno, Ryutaro and Wang, Amy and Li, Brenna and Amin, Mohamed and Cheng, Yong and others},
  journal={Nature},
  pages={1--9},
  year={2025},
  publisher={Nature Publishing Group UK London}
}

@article{li2023llava,
  title={Llava-med: Training a large language-and-vision assistant for biomedicine in one day},
  author={Li, Chunyuan and Wong, Cliff and Zhang, Sheng and Usuyama, Naoto and Liu, Haotian and Yang, Jianwei and Naumann, Tristan and Poon, Hoifung and Gao, Jianfeng},
  journal={Advances in Neural Information Processing Systems},
  volume={36},
  pages={28541--28564},
  year={2023}
}

@article{cao2025development,
  title={The development landscape of large language models for biomedical applications},
  author={Cao, Zhiyuan and Keloth, Vipina K and Xie, Qianqian and Qian, Lingfei and Liu, Yuntian and Wang, Yan and Shi, Rui and Zhou, Weipeng and Yang, Gui and Zhang, Jeffrey and others},
  journal={Annual Review of Biomedical Data Science},
  volume={8},
  year={2025},
  publisher={Annual Reviews}
}

@article{luo2024biomedgpt,
  title={Biomedgpt: An open multimodal large language model for biomedicine},
  author={Luo, Yizhen and Zhang, Jiahuan and Fan, Siqi and Yang, Kai and Hong, Massimo and Wu, Yushuai and Qiao, Mu and Nie, Zaiqing},
  journal={IEEE Journal of Biomedical and Health Informatics},
  year={2024},
  publisher={IEEE}
}

@inproceedings{mathur2024mind,
  title={MIND over Body: Adaptive Thinking using Dynamic Computation},
  author={Mathur, Mrinal and Pearlmutter, Barak A and Plis, Sergey M},
  booktitle={The Thirteenth International Conference on Learning Representations},
  year={2024}
}

@article{khan2025optimizing,
  title={Optimizing Deep Learning Models for Resource-Constrained Environments With Cluster-Quantized Knowledge Distillation},
  author={Khan, Niaz Ashraf and Rafat, AM Saadman},
  journal={Engineering Reports},
  volume={7},
  number={5},
  pages={e70187},
  year={2025},
  publisher={Wiley Online Library}
}

@article{singhal2023large,
  title={Large language models encode clinical knowledge},
  author={Singhal, Karan and Azizi, Shekoofeh and Tu, Tao and Mahdavi, S Sara and Wei, Jason and Chung, Hyung Won and Scales, Nathan and Tanwani, Ajay and Cole-Lewis, Heather and Pfohl, Stephen and others},
  journal={Nature},
  volume={620},
  number={7972},
  pages={172--180},
  year={2023},
  publisher={Nature Publishing Group}
}

@article{zhang2024mpmoe,
  title={Mpmoe: Memory efficient moe for pre-trained models with adaptive pipeline parallelism},
  author={Zhang, Zheng and Xia, Yaqi and Wang, Hulin and Yang, Donglin and Hu, Chuang and Zhou, Xiaobo and Cheng, Dazhao},
  journal={IEEE Transactions on Parallel and Distributed Systems},
  volume={35},
  number={6},
  pages={998--1011},
  year={2024},
  publisher={IEEE}
}

@article{muhammad2024unveiling,
  title={Unveiling the black box: A systematic review of Explainable Artificial Intelligence in medical image analysis},
  author={Muhammad, Dost and Bendechache, Malika},
  journal={Computational and structural biotechnology journal},
  volume={24},
  pages={542--560},
  year={2024},
  publisher={Elsevier}
}

@article{rosenbacke2024explainable,
  title={How explainable artificial intelligence can increase or decrease clinicians’ trust in AI applications in health care: systematic review},
  author={Rosenbacke, Rikard and Melhus, {\AA}sa and McKee, Martin and Stuckler, David},
  journal={Jmir Ai},
  volume={3},
  pages={e53207},
  year={2024},
  publisher={JMIR Publications Toronto, Canada}
}
